%% file: main.tex
\documentclass[]{fairmeta}

\title{
    \resizebox{\linewidth}{!}{Watermarking Autoregressive Image Generation}
} 

\author[1,2,\dagger]{Nikola Jovanović}
\author[1,3]{Ismail Labiad}
\author[1]{Tomáš Souček} 
\author[2]{Martin Vechev}
\author[1]{Pierre Fernandez}
   
\affiliation[1]{Meta FAIR}
\affiliation[2]{ETH Zurich}
\affiliation[3]{Université Paris-Saclay} 

\contribution[\dagger]{Work done during a Meta internship.}

\abstract{
\input{sections/0-abstract}
} 
 
\correspondence{\email{nikola.jovanovic@inf.ethz.ch}, \email{pfz@meta.com}}

\metadata[Code]{\url{https://github.com/facebookresearch/wmar}}
\newcommand{\firstpagefooter}{39th Conference on Neural Information Processing Systems (NeurIPS 2025).}

\PassOptionsToPackage{numbers, sort&compress}{natbib}

\usepackage[utf8]{inputenc} %
\usepackage[T1]{fontenc}    %
\usepackage{url}            %

\usepackage{amsfonts}       %
\usepackage{nicefrac}       %
\usepackage{microtype}      %
\usepackage{xcolor}         %
\usepackage{enumitem}
\usepackage{nicefrac}       %

\usepackage{amsmath}        %
\usepackage{dsfont}        %

\usepackage{graphicx}       %
\usepackage{caption} 
\usepackage{siunitx} 
\usepackage{booktabs}
\usepackage{multirow} 
\usepackage{colortbl}  %
\usepackage{tabularx}
\usepackage{subcaption}
\usepackage{wrapfig}
\usepackage{tabulary} 
\usepackage{bbm}
\usepackage{lipsum}

\usepackage[most]{tcolorbox} 
\usepackage{listings}

\creflabelformat{equation}{#2#1#3}

\usepackage{titlesec}
\titlespacing*{\paragraph}{0pt}{1em}{1em}

\usepackage{fancyhdr}
\fancypagestyle{firstpage}{%
    \fancyhf{}%
    \fancyfoot[C]{\vspace{-1em} \hrule \vspace{1em} \small\firstpagefooter}%
}

\input{math_commands.tex}

\input{macros.tex}

\begin{document}
 \sisetup{
text-series-to-math = true,
}

\maketitle

\thispagestyle{firstpage}

\input{sections/1-introduction}

\input{sections/2-3-bgr-relwork}
\input{sections/4a-method}

\input{sections/4b-finetuning}

\input{sections/4c-sync}

\input{sections/5-experiments}

\input{sections/6-audio}

\input{sections/7-conclusion}

\clearpage
\section*{Acknowledgements}

We thank Sylvestre-Alvise Rebuffi, Tom Sander, Hervé Jégou, Alex Mourachko, Hady Elsahar, Robin San Roman, Ram Pasunuru, and Emily Dinan for insightful discussions throughout the project. We are grateful to anonymous reviewers for their valuable feedback.

\bibliographystyle{ieeenat_fullname}
\bibliography{references}

\appendix
\section*{{\Huge Appendix}}
\input{sections/app-other}

\input{sections/app-watermarks}
\input{sections/app-consistency}
\input{sections/app-syncdetails}

\input{sections/app-expdetails}
\input{sections/app-moreresults}

\input{sections/app-qualitative}

\input{sections/app-interleaved}

\end{document}

%% file: sections/0-abstract.tex
Watermarking the outputs of generative models has emerged as a promising approach for tracking their provenance.
Despite significant interest in autoregressive image generation models and their potential for misuse, no prior work has attempted to watermark their outputs at the token level.
In this work, we present the first such approach by adapting language model watermarking techniques to this setting.
We identify a key challenge: the lack of reverse cycle-consistency (RCC), wherein re-tokenizing generated image tokens significantly alters the token sequence, effectively erasing the watermark. 
To address this and to make our method robust to common image transformations, neural compression, and removal attacks, we introduce (i) a custom tokenizer-detokenizer finetuning procedure that improves RCC, and (ii) a complementary watermark synchronization layer. 
As our experiments demonstrate, our approach enables reliable and robust watermark detection with theoretically grounded p-values. Code and models are available at \url{https://github.com/facebookresearch/wmar}.

%% file: math_commands.tex
\usepackage{amsmath,amsfonts,bm}
\usepackage{amsthm}

\def\1{\bm{1}}

\DeclareMathAlphabet{\mathsfit}{\encodingdefault}{\sfdefault}{m}{sl}
\SetMathAlphabet{\mathsfit}{bold}{\encodingdefault}{\sfdefault}{bx}{n}

\newcommand{\lr}{\alpha}

\DeclareMathOperator*{\argmin}{arg\,min}

%% file: macros.tex
\usepackage{ifthen}
\newboolean{set_me_to_remove_todos}
\setboolean{set_me_to_remove_todos}{true} %
\newcommand{\todo}[1]{{\color{red} [\textbf{TODO}: #1]}}
\newcommand{\pierre}[1]{{\color{blue} [\textbf{Pierre}: #1]}}
\newcommand{\nj}[1]{{\color{orange} [\textbf{Nikola}: #1]}}
\newcommand{\ismail}[1]{{\color{purple} [\textbf{Ismail}: #1]}}

\ifthenelse{\boolean{set_me_to_remove_todos}}{
    \renewcommand{\todo}[1]{}
    \renewcommand{\pierre}[1]{}
    \renewcommand{\nj}[1]{}
    \renewcommand{\ismail}[1]{}
    
}{}

\definecolor{badred}{rgb}{0.8,0.2,0.2}
\definecolor{basecolor}{HTML}{E63946}
\definecolor{ftcolor}{HTML}{457B9D}
\definecolor{ftaugcolor}{HTML}{2A9D8F}
\definecolor{ftaugsynccolor}{HTML}{e2b33c}
\definecolor{textonlyorange}{HTML}{F4A261}

\def\1{\mathbf{1}}

\def\H{\mathcal{H}}

\usepackage{xspace} %

\newcommand{\eg}{e.g.,\@ }
\newcommand{\ie}{i.e.,\@ }

\newcolumntype{x}[2]{S[table-format=#1.#2,table-auto-round]}

\newcommand{\model}{\ensuremath{\mathcal{M}}}
\newcommand{\tok}[1]{\ensuremath{\mathcal{T}_{#1}}}
\newcommand{\detok}[1]{\ensuremath{\mathcal{D}_{#1}}}

\usepackage{tikz}

\usetikzlibrary{shapes.misc, positioning}
\usetikzlibrary{positioning, backgrounds, fit, calc}
\newcommand{\bsize}{0.15}
\newcommand{\boxyRed}{%
  \begin{tikzpicture}[every node/.style={inner sep=0,outer sep=0}]
    \protect\draw[fill={rgb,255:red,250;green,0;blue,0}] (0,0) rectangle ++(\bsize, \bsize);
  \end{tikzpicture}%
}
\newcommand{\boxyGreen}{%
  \begin{tikzpicture}[every node/.style={inner sep=0,outer sep=0}]
    \protect\draw[fill={rgb,255:red,0;green,200;blue,0}] (0,0) rectangle ++(\bsize, \bsize);
  \end{tikzpicture}%
}
\newcommand{\boxyWhite}{%
  \begin{tikzpicture}[every node/.style={inner sep=0,outer sep=0}]
    \protect\draw[fill={rgb,255:red,250;green,250;blue,250}] (0,0) rectangle ++(\bsize, \bsize);
  \end{tikzpicture}%
}

\newcommand{\loss}{\ensuremath{\mathcal{L}}}
\newcommand{\augset}{\ensuremath{\mathcal{A}}}
\newcommand{\expectation}{\mathop{\mathbb{E}}}

\newcommand{\taming}{\textsc{Taming}\xspace}
\newcommand{\cham}{\textsc{Chameleon}\xspace}
\newcommand{\rarxl}{\textsc{RAR-XL}\xspace}
\newcommand{\moshi}{\textsc{Moshi}\xspace}
\newcommand{\mimi}{\textsc{Mimi}\xspace}

\newcommand{\base}{{{\textsc{Base}}}\xspace}
\newcommand{\ft}{{{\textsc{FT}}}\xspace}
\newcommand{\ftaug}{{{\textsc{FT+Augs}}}\xspace}
\newcommand{\ftaugsync}{{{\textsc{FT+Augs+Sync}}}\xspace}

\newcommand{\baseTABLE}{{{\textsc{Base}}}\xspace}
\newcommand{\ftTABLE}{{{\textsc{FT}}}\xspace}
\newcommand{\ftaugTABLE}{{{\textsc{FT+Augs}}}\xspace}
\newcommand{\ftaugsyncTABLE}{{{\textsc{FT+Augs+Sync}}}\xspace}

\definecolor{green1}{HTML}{008000}
\definecolor{green2}{HTML}{03ad55}
\definecolor{red1}{HTML}{FF0000}
\definecolor{red2}{HTML}{880000}

\lstdefinestyle{mystyle}{
    breaklines=true,
    basicstyle=\scriptsize\sffamily,
    numbers=none,
    language={},
    framextopmargin=0pt,
    framexbottommargin=0pt,
    breakindent=0pt,
    showspaces = false,
    keywordstyle=\bfseries,
    showstringspaces=false,
    columns=fullflexible,
    morekeywords={acoustic, guitar, thatch},
    moredelim=**[is][\color{green1}]{@}{@},
    moredelim=**[is][\color{red2}]{^}{^}
}

\newtcblisting{prompt}[1][]{
    arc=3pt, outer arc=3pt,
    width=.93\linewidth,
    left=1mm,
    top=0mm,
    bottom=0mm,
    title=\userheader, 
    colback=red!5!white,
    colframe=red!50!black,
    fonttitle=\bfseries,
    listing only, 
    listing options={style=mystyle,deletekeywords={}},
    breakable,
    #1
}

\definecolor{oai}{HTML}{10a37f}
\newtcblisting{chameleon}[2][]{
    arc=3pt, outer arc=3pt,
    width=.93\linewidth,
    left=1mm,
    top=0mm,
    bottom=0mm,
    title=\chameleonheader, 
    colback=oai!5!white,
    colframe=oai!40!black,
    fonttitle=\bfseries,
    listing only, 
    listing options={style=mystyle,deletekeywords={acoustic,guitar,thatch}},
    breakable,
    after upper={
      \vspace{0.5em}
      \includegraphics[width=0.45\textwidth]{prompts/#2.png}
    },
    #1
}
\newtcblisting{detector}[3][]{
    arc=3pt, outer arc=3pt,
    width=.93\linewidth,
    left=1mm,
    top=0mm,
    bottom=0mm,
    title=\detectorheader{#3}, 
    colback=metablue!5!white,
    colframe=metablue!40!black,
    fonttitle=\bfseries,
    listing only, 
    listing options={style=mystyle,deletekeywords={acoustic,guitar,thatch}},
    breakable,
    after upper={
      \vspace{0.5em}
      \includegraphics[width=0.45\textwidth]{prompts/#2.png}
    },
    #1
}

\newcommand{\chameleonheader}{
    \begin{tikzpicture}
      \node[anchor=north] {\pgftext{\includegraphics[width=0.34cm]{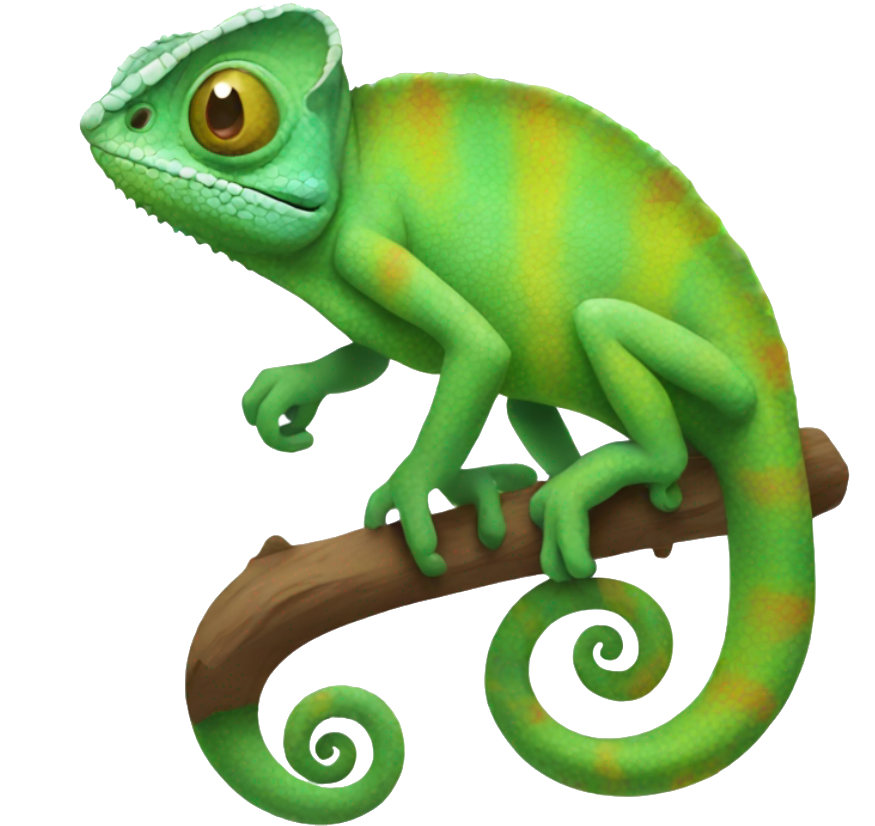}}};
    \end{tikzpicture}
    ~Chameleon
}
\newcommand{\userheader}{
    \begin{tikzpicture}
      \node[anchor=north] {\pgftext{\includegraphics[width=0.34cm]{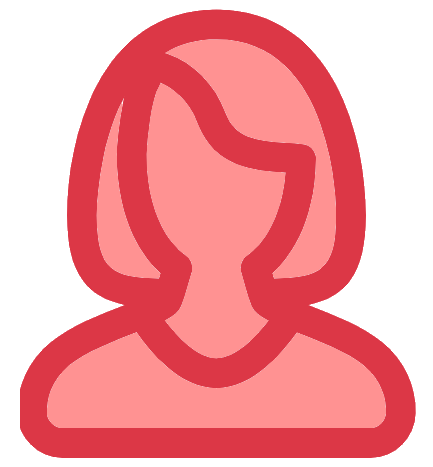}}};
    \end{tikzpicture}
    ~User Prompt
}
\newcommand{\detectorheader}[1]{
    \begin{tikzpicture}
      \node[anchor=north] {\pgftext{\includegraphics[width=0.34cm]{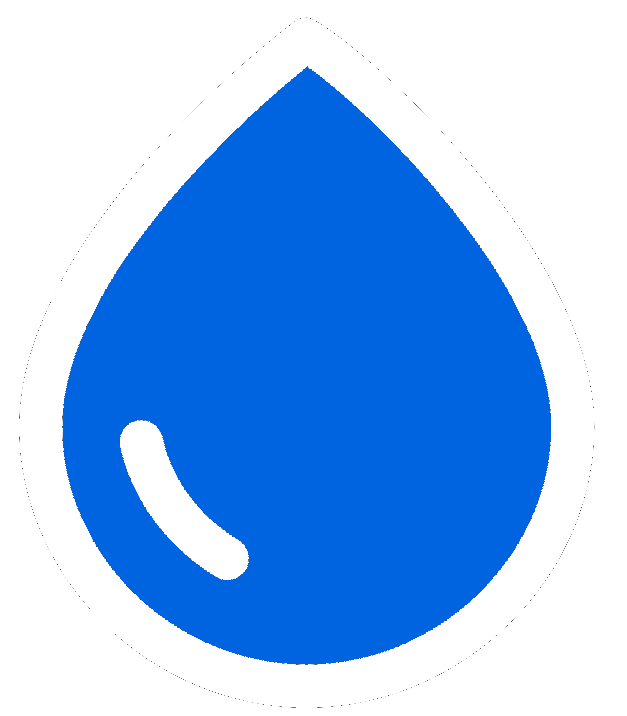}}};
    \end{tikzpicture}
    ~Watermark Detector (#1)
}

\newcommand{\nalive}{n_{\text{alive}}}

\newcommand{\aref}[1]{\hyperref[#1]{App.~\ref*{#1}}}

%% file: sections/1-introduction.tex
\begin{figure}[h]
    \centering
    \vspace{-1em}
    \includegraphics[width=\linewidth,clip,trim={0 7.1in 0in 0}]{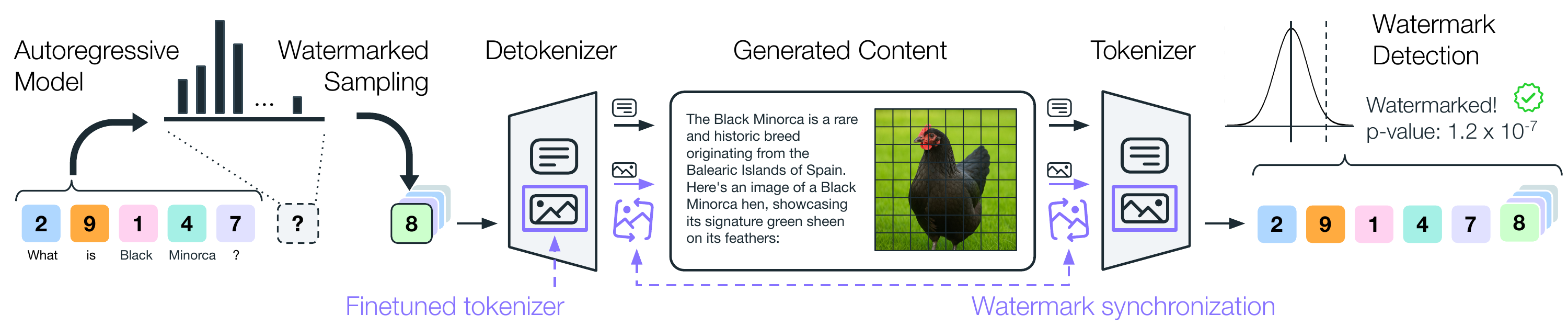}
    \caption{We watermark autoregressively generated images together with text in a theoretically principled way by adapting LLM watermarking. We identify and address the novel challenges present in this setting (\cref{sec:method}) via a custom (de)tokenizer finetuning procedure (\cref{ssec:method:ft}) and a watermark synchronization layer (\cref{ssec:method:sync}).}
    \label{fig:overview}
    \vspace{-1em}
\end{figure}

\section{Introduction}\label{sec:intro}

Autoregressive models are powerful frameworks for understanding and generating diverse content types.
By converting multiple modalities into discrete representations via custom tokenizers~\citep{van2017neural,razavi2019generating,soundstream}, a single transformer is able to seamlessly process multiple domains, including text, images~\citep{team2024chameleon,tian2024mm,chern2024anole,wu2024janus}, audio~\citep{Borsos2022AudioLMAL,defossez2024moshi}, and even molecules~\citep{hsu2022learning}. 
Following patterns observed in large language models (LLMs), established scaling laws~\citep{henighan2020scaling,aghajanyan2023scaling,shukor2025scaling} demonstrate that the performance of these models improves predictably with size and computational resources, leading to increasing adoption across research and industry~\citep{team2023gemini,openai2024gpt4o,meta2025llama}.
Most notably, in the image domain, autoregressive models are widely studied as an alternative to diffusion models for high-quality generation~\citep{ramesh2021zero,yu2022scaling,tian2024visual,sun2024autoregressive}.

\paragraph{Watermarking generative model outputs.}
Regardless of the specific method, the widespread deployment of high-quality generative models has made the detection of AI-generated content increasingly challenging.
This has raised significant concerns about misuse, including deepfakes, harmful content generation, and intellectual property violations.
One promising direction to help address these issues is \emph{generative AI watermarking}, in which the model provider proactively embeds imperceptible signals into generated content to verify its origin later.

Recent research in this area can be categorized into post-hoc methods, which modify already generated content in a model-agnostic way~\citep{zhu2018hidden,deepmind2023watermarking,bui2023trustmark,fernandez2024video,postmark,sanroman2024proactive}, and modality-specific generation-time methods, which alter the generation process of a specific model~\citep{kirchenbauer2023watermark,aaronson2023watermarking,wen2023tree,fernandez2023stable,yang2024gaussian,Dathathri2024}.
The latter are the standard in LLM watermarking, offering theoretically grounded watermark detection with provably low false positive rates.
However, most image watermarking research focuses on diffusion models, and no prior work has attempted to adapt LLM watermarks to other token types, which could be a way to develop a robust and theoretically principled watermark that is tailored to autoregressive image generation models.
This motivates our key question:
\begin{center}
\textit{Can we robustly watermark autoregressive image generation models at the token level?}
\end{center}

\paragraph{This work.}
To answer this question, we extend LLM watermarks to watermark autoregressively generated image tokens as illustrated in~\cref{fig:overview}.
We identify and address a key technical challenge.
Namely, while image tokenization is designed to be \emph{forward cycle-consistent}, \ie tokenizing and detokenizing an image does not significantly alter it, \emph{reverse cycle-consistency} (RCC) is often violated.
In particular, we show that decoding model-generated tokens and then re-tokenizing the resulting image leads to, on average, one-third of the tokens being different.
The tokens differ even more if the images are transformed between generation and watermark detection (\eg JPEG compressed or cropped), which is a common scenario in practice.
While RCC may not be important for generative model performance, it is a crucial prerequisite for robust generation-time watermarking.

To mitigate this, we adopt two main strategies, shown in~\cref{fig:overview}.
We introduce a lightweight finetuning procedure that optimizes the detokenizer and tokenizer to be more reverse cycle-consistent, improving watermark power and robustness to valuemetric transformations (\eg JPEG) and attacks such as diffusion purification~\citep{nie2022diffusion} and neural compression~\citep{bmshj18, cstk20, esser2021taming, rombach2022high}.
To then improve robustness to geometric transformations (\eg flips), we introduce a complementary post-hoc \emph{watermark synchronization} layer, repurposing localized watermarking~\citep{sanroman2024proactive} to detect and revert geometric transformations and recover original tokens. 
As our experiments show, this results in a watermark that is quality-preserving, effective, and robust.
To inspire future work, in \cref{sec:audio} we take first steps to extend our approach to audio, another modality where autoregressive generation via tokenization is popular~\citep{Borsos2022AudioLMAL,nguyen2025spirit,copet2024simple,defossez2024moshi}.

Our \textbf{main contributions} are: 
\emph{(i)} the first study of watermarking for autoregressive image generation, and the identification of reverse cycle-consistency (RCC) as a key challenge (\cref{sec:method}); 
\emph{(ii)} a lightweight finetuning procedure that improves RCC and watermark power (\cref{ssec:method:ft}); 
\emph{(iii)} a watermark synchronization layer for geometric robustness (\cref{ssec:method:sync}); 
and \emph{(iv)} a thorough empirical evaluation showing our watermark preserves generation quality and is robust to a wide range of attacks (\cref{sec:experiments}).

%% file: sections/2-3-bgr-relwork.tex
\section{Background and Related Work}\label{sec:bgr_relwork}

\paragraph{Autoregressive image models.}
A long-studied approach to image generation, and the focus of our work, is to first learn an image tokenizer and then train a (conditioned) autoregressive model $\model$ such as a transformer, to create images by generating corresponding token sequences~\citep{ramesh2021zero,yu2022scaling,tian2024visual,yu2024randomized,tschannen2024givt,wang2024emu3,sun2024autoregressive}.
Notable examples of such models include
DALL-E~\citep{ramesh2021zero}, 
Parti~\citep{yu2022scaling}, 
VAR~\citep{tian2024visual}, and
RAR~\citep{yu2024randomized}.
This approach is central to models for interleaved multimodal generation~\citep{chern2024anole, ge2023making, liu2024world, tian2024mm, lin2024moma, team2024chameleon, zhan2024anygpt, wu2024janus}, such as
Chameleon~\citep{team2024chameleon}, 
AnyGPT~\citep{zhan2024anygpt}, 
or Janus~\citep{wu2024janus}.
In this work, we do not consider other models that combine diffusion with autoregressive mechanisms~\citep{xie2024show,zhou2024transfusion,fan2024fluid,li2024autoregressive,ma2024janusflow}.

\paragraph{Tokenization.}
Formally, for a target modality $m$ (in this work primarily text or image, but also audio in \cref{sec:audio}), a \emph{tokenizer} $\tok{m}$ maps each data sample $x$ to a sequence of integer tokens $s = (s_1, \dots, s_T) \in V^T$, where $V$ is the predefined vocabulary.
The \emph{detokenizer} $\detok{m}$ attempts to reverse this process.
Most text tokenizers are based on \emph{byte-pair encoding} (BPE)~\citep{bpe}.
While alternative approaches have been explored~\citep{li2024autoregressive,tian2024visual}, the tokenization of images overwhelmingly relies on vector quantization (VQ)~\citep{gray1984vector,li2024autoregressive}.
Most models use VQ-VAE~\citep{van2017neural,razavi2019generating} or its variants VQGAN~\citep{esser2021taming}, ImprovedVQGAN~\citep{yu2021vector}, and FSQ~\citep{mentzer2023finite}.
VQ tokenizers generally consist of an encoder network $E$ and a quantizer $Q_C$. 
$E$ maps $x$ to a sequence of \emph{soft latents} $z = E(x) \in \mathbb{R}^{T \times d}$.
\looseness=-1
Then, $Q_C$ replaces each $z_i$ with the index of the nearest entry in a \emph{codebook} $C \in \mathbb{R}^{|V| \times d}$ to obtain discrete tokens $s \in V^T$:
\begin{align}
    s_i = Q_C(z_i) = \argmin_{j \in \{1, \ldots, k\}} \|z_i - C_j\|_2^2.
\end{align}
The detokenizer $\detok{m}$ replaces each $s_i$ with the corresponding $\hat{z}_{i} = C_{s_i}$ (\emph{hard latents}), and then applies a decoder network $D$ to obtain the detokenized sample $\hat{x} = D(\hat{z})$.
All components ($E$, $C$, and $D$) are typically trained jointly, primarily with the reconstruction objective.

\paragraph{Watermarking AI-generated outputs.}
Methods for watermarking of AI-generated outputs broadly fall into two main categories based on the stage at which the watermark is introduced.
\emph{Generation-time watermarks} directly alter generations to embed statistically detectable patterns, and are the standard for LLM watermarking~\citep{kirchenbauer2023watermark,kuditipudi2023robust,aaronson2023watermarking,christ2023undetectable}.
Similar approaches also exist for diffusion models~\citep{wen2023tree,yang2024gaussian,fernandez2023stable}. 
In contrast, \emph{post-hoc watermarks} modify previously generated outputs in a modular model-agnostic way, by paraphrasing text~\citep{blackbox,remark,postmark} or altering image pixels~\citep{zhu2018hidden,luo2020distortion,tancik2020stegastamp,CIN,MBRS,bui2023trustmark,bui2023rosteals}. 

For both of these categories, we say that a watermarking scheme is \emph{zero-bit} if only the presence of the watermark can be detected, and \emph{multi-bit} if it embeds a binary message.
This message embedding is sometimes done in a \emph{localized} way~\citep{sander2024watermark,sanroman2024proactive}, such that a different message can be extracted from different parts of the same data sample (\eg one per pixel).
While post-hoc watermarks have broad applicability, generation-time approaches that introduce semantic changes to the content often offer superior robustness to attacks such as diffusion purification~\citep{saberi2023robustness} and provide provable, key-based guarantees on false positive rates (unlike neural extraction methods, where recovered bits may be biased or correlated~\citep[App. B.5]{fernandez2023stable}). 

\paragraph{LLM watermarking.}
In this work, we focus on the LLM watermark of \citet{kirchenbauer2023watermark} (KGW, green/red scheme).
At each step $i$ of generation, this method uses a secret key and previous $h$ tokens of \emph{context} $s_{i-h:i}$ to pseudorandomly partition the vocabulary $V$ of the tokenizer into $\gamma |V|$ \emph{green} tokens $\mathcal{G}_i$ and other \emph{red} tokens $\mathcal{R}_i$. 
The logits corresponding to $\mathcal{G}_i$ are then increased by $\delta$, the \emph{watermark strength}.
The watermark detector computes the \emph{score} $S = \sum_{i=h+1}^T \mathds{1}(s_i \in \mathcal{G}_i)$ as the number of green tokens in the given sequence of $T$ tokens. 
Under the null hypothesis $\H_0$ (no watermark), $S$ follows a binomial distribution with parameters $T-h$ and $\gamma$. 
The p-value (often denoted as $p$) is calculated as:
\begin{align} \label{eq:pvalue}
    \text{p-value}(S,T,h,\gamma) = \text{Prob} \left( X \geq S \mid X \sim \text{Binomial}(T-h, \gamma) \right).
\end{align}
A low p-value proves that the content was generated with $\mathcal{M}$ (more details in \aref{app:watermarks})  .

\paragraph{Concurrent work.}
Several concurrent works also study watermarks for autoregressive image models: \textsc{IndexMark}~\citep{tong2025training} (by replacing generated tokens in a way inspired by KGW), \textsc{C-reweight}~\citep{wu2025watermarkimage} (via semantic partitioning similar to the one we describe at the bottom of~\cref{app:moreresults:ablations}), \textsc{LBW}~\citep{hui2025autoregressive} (via KGW-style biasing), and \textsc{WIAR}~\citep{meintz2025radioactive} (from the perpsective of radioactivity~\citep{sander2024watermark}).
\citet{wu2025watermark} concurrently apply KGW and DiPmark~\citep{wu2024dipmark} to autoregressive speech generation.
As elaborated on in the following sections, our work uniquely addresses the core issue of low reverse cycle-consistency via finetuning, proposes a synchronization mechanism that can boost geometric robustness beyond this work, and heavily centers the evaluation on robustness against a wide range of realistic transformations and attacks, such as geometric modifications and diffusion purification.

%% file: sections/4a-method.tex
\section{Watermarking Autoregressive Image Generation}\label{sec:method}

In this section, we present our approach to watermarking autoregressive image models.
We identify and address the key challenge of low \emph{reverse cycle-consistency (RCC)} via tokenization finetuning (\cref{ssec:method:ft}) and watermark synchronization (\cref{ssec:method:sync}). 
As our experiments in~\cref{sec:experiments} demonstrate, this leads to a strong and robust watermark that does not affect generation quality.

\paragraph{Setting.}
A model provider (Alice) deploys an autoregressive model $\model$ that may generate arbitrarily interleaved text and images, using a tokenizer $\tok{m}$ and detokenizer $\detok{m}$ for each modality $m \in \{\text{text}, \text{image}\}$.
In line with the most prominent choices, we assume BPE for text~\citep{bpe} and VQ for images~\citep{van2017neural,esser2021taming}.
Alice's goal is \emph{out-of-model, generation-time, zero-bit watermarking} (see \cref{sec:bgr_relwork}), \ie embedding a later detectable watermark in \emph{all} outputs of $\model$, without modifying the model's weights.
We assume that Bob has only black-box access to $\model$, and no access to any $\tok{m}$ or $\detok{m}$.
  \input{figures/gimmick}

\paragraph{Adapting LLM watermarking.}
When $\model$ generates text, we directly apply KGW (\cref{sec:bgr_relwork}) with context size $h=1$.
For images, using a fixed split ($h=0$), known to make watermarks insecure for text, \ie easy to reverse-engineer~\citep{kirchenbauer2024on, jovanovic2024watermark,milpstealing}, may in our case be a more viable choice due to the opacity of the VQ tokenizers.
We thus explore both $h \in \{0, 1\}$ in our experiments in \cref{sec:experiments}.
Another degree of freedom is the choice of watermark context---we did not find exploiting the $2$D structure of images to be beneficial, despite the intuition that using tokens that correspond to spatially close regions as the watermark context (instead of those last generated by the autoregressive model) may benefit robustness.
This idea and more values of watermark parameters are explored in \aref{app:moreresults:ablations}.

\paragraph{Detection.}
Given samples $x^{(i)}$ of varying modalities that Alice suspects were generated by $\model$ (\eg a post on a breed of hen as in \cref{fig:overview}), she can apply~\cref{eq:pvalue} in a unified way.
We first tokenize each $x^{(i)}$ to $s^{(i)}$ of length $T^{(i)}$ tokens, and score it using the corresponding $h^{(i)}$ to obtain a score $S^{(i)}$.
We next sum all $S^{(i)}, T^{(i)}$, and $h^{(i)}$, and deduplicate scored (context,~token) pairs across all samples to preserve statistical soundness~\citep{kirchenbauer2023watermark,fernandez2023three,radioactive,jovanovic2025ward}.
Then, we apply~\cref{eq:pvalue} to obtain a single p-value.
Notably, the same $\gamma$ must be used across all modalities.
Alice may reject $\H_0$ (flag content as watermarked) if the p-value is below the desired false positive rate (FPR). 
In~\cref{ssec:experiments:interleaved} we investigate the benefits of jointly watermarking multiple modalities, and discuss the involved tradeoffs.
  
In~\cref{fig:gimmick}, we visualize the watermark on an image generated with \taming~\citep{esser2021taming}, by applying it only on the second half of generated tokens.
As we later confirm in \cref{sec:experiments}, the watermark imperceptibly modifies images by altering semantics, while achieving high watermark confidence (low p-value).

\paragraph{Challenge: reverse cycle-consistency (RCC).}
The tokens $s$ shown in~\cref{fig:gimmick} as input to the detector are those generated by the autoregressive model, which is not realistic. 
In practice, to apply the detector to a sample $x'$, Alice must first tokenize it as $s' = \tok{m}(x')$.
If tokens $s'$ significantly differ from $s$, the watermark may be lost.
To quantify this, we define the \emph{token match} as:

\vspace{-1.5em}
\begin{align}\label{eq:tm}
\text{TM}(s, s') = \frac{1}{T} \sum_{i=1}^T \mathds{1}(s_i = s'_i), 
\\[-1.3em] \nonumber
\end{align}
where $s' = \tok{m}(\detok{m}(s))$.
We say that \emph{reverse cycle-consistency (RCC)} holds if $\text{TM}(s, s') \approx1$.
In~\aref{app:consistency:related} we discuss similar concepts studied in prior work and relate them to RCC.

Interestingly, RCC is not guaranteed to hold even in the text domain, despite BPE tokenizers ensuring \emph{forward cycle-consistency (FCC)}, \ie $\detok{\text{text}}(\tok{\text{text}}(x)) = x$ always holds.
For example, if tokens $A$, $B$ and their concatenation $AB$ all exist in the vocabulary, then the token sequence $[A, B]$ will be re-tokenized as a single token $AB$ due to the greedy tokenization algorithm of BPE, effectively violating RCC.
In~\aref{app:consistency:text} we discuss this in more detail and provide a real example.
Still, RCC in text largely holds in practice, which is also evidenced by the fact that no prior LLM watermarking work has cited related challenges.
We confirm this experimentally: across $1000$ completions generated with \textsc{Llama3.1-8B-Instruct} the average token match was $0.995$.

\paragraph{RCC in image generation.}
We repeat this experiment on image models, presenting the results in \cref{tab:tokens} (full details in \aref{app:expdetails:rccft}).
We also consider the case where images undergo transformations ($x \to a(x)$) before re-tokenization.
Without transformations (\emph{Original}), RCC is already weaker than expected with $\text{TM} = 0.66$.
As \cref{sec:experiments} will show, this often suffices under ideal conditions (\eg our example in~\cref{fig:gimmick} has a p-value of $10^{-9}$ after re-tokenization).
However, common valuemetric transformations (blur, noise, JPEG, brighten) lower TM (\eg to 0.31 for JPEG with $Q=25$), and geometric ones (rotate, flip, crop) cause a further drop to almost $0$.
Two key factors explain this behavior.
First, neural image tokenizers are trained for FCC, not RCC.
Their training data does not include detokenized samples, which often lie off-manifold.
Second, spatial sensitivity of the tokenizer causes semantic-preserving edits to easily alter most tokens.
We next show how to mitigate this.

\input{tables/tokens}

%% file: figures/gimmick.tex
\begin{figure}[t]
  \centering
  \resizebox{\linewidth}{!}{%
  \small
  \newcommand{\colspace}{\hspace{0.035\linewidth}}
    \begin{tabular}{c@{\colspace}c@{\colspace}c@{\colspace}c}
      \toprule
      Unwatermarked Generation & Tokens (\textcolor{red!90}{$p=0.73$}) & Watermarked Generation & Tokens (\textcolor{teal!75}{$p=10^{-11}$}) \\
      \midrule
      \includegraphics[width=0.21\linewidth]{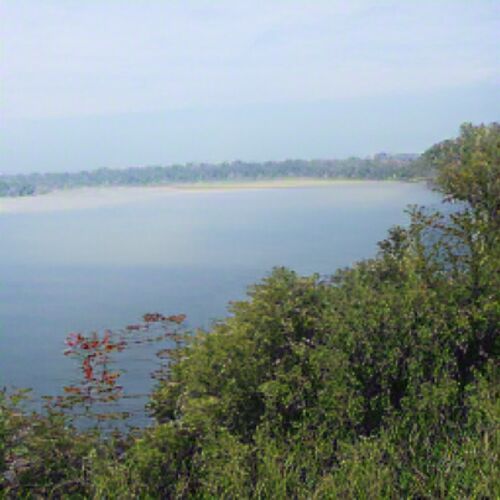} & 
      \includegraphics[width=0.21\linewidth]{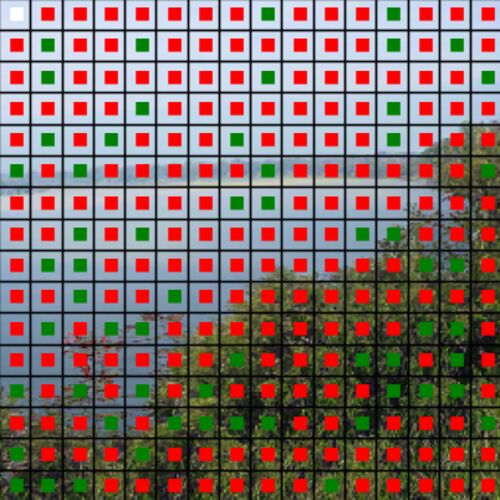} &  
      \includegraphics[width=0.21\linewidth]{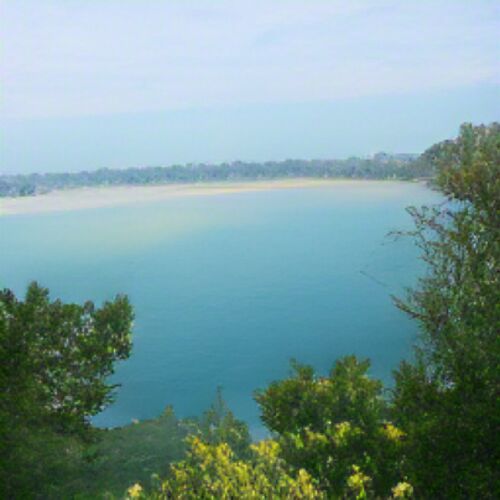} & 
      \includegraphics[width=0.21\linewidth]{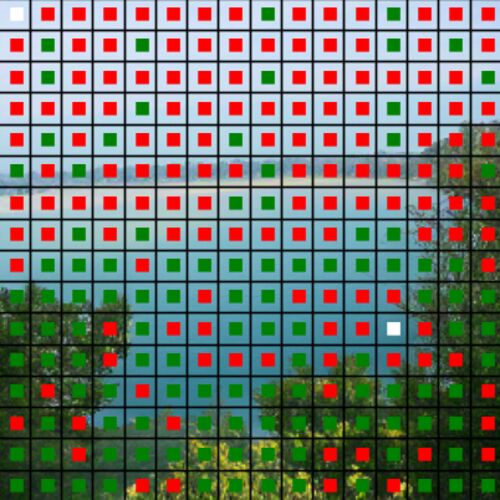} \\
      \bottomrule
    \end{tabular}
  }
  \caption{
    {Example of our watermark on an autoregressively generated image.} 
    We generate the upper half of the image without the watermark.
    We then complete the bottom half in the same way (\emph{left}) or with the watermark (\emph{right}).
    The overlay indicates generated image tokens detected as green (\boxyGreen), red (\boxyRed), or ignored as a duplicate (\boxyWhite).
    The watermark only alters semantics and could be detected even when applied only partially as in this case.
    }
  \label{fig:gimmick}
\end{figure} 

%% file: tables/tokens.tex
\begin{table}[t]
    \centering 
    \caption{Average token match between $1000$ image token sequences generated with \taming (see~\cref{sec:experiments}) and their re-tokenized versions, which may also undergo image transformations before re-tokenization.}
    \resizebox{0.9\linewidth}{!}{%
    \begingroup 
    \small
    \setlength{\tabcolsep}{5pt} %
    \begin{tabular}{
        x{1}{2}
        x{1}{2} 
        x{1}{2}
        x{1}{2}
        x{1}{2}
        x{1}{2}  
        x{1}{2}
        x{1}{2}
        }
        \toprule
    \textbf{Original} & \textbf{Blur}~\scriptsize{$ksz=9$} 
    & \textbf{Noise}~\scriptsize{$\sigma=0.1$} 
    & \textbf{JPEG}~\scriptsize{$Q=25$} 
    & \textbf{Brighten}~\scriptsize{$2 \times$} 
    & \textbf{Rotate}~\scriptsize{$10^{\circ}$} 
    & \textbf{Flip}~\scriptsize{$\leftrightarrow$} 
    & \textbf{Crop}~\scriptsize{$0.75$} \\
        \midrule
        0.66 & 0.26 & 0.17 & 0.31 & 0.11 & 0.02 & 0.01 & 0.01 \\
        \bottomrule
    \end{tabular}
    \endgroup
    }
    \label{tab:tokens}
\end{table}

%% file: sections/4b-finetuning.tex
\subsection{Finetuning for Reverse Cycle-consistency} \label{ssec:method:ft}

We propose a finetuning procedure (illustrated in \cref{fig:ft}) that improves RCC in image tokenizers.
Recall the VQ components (\cref{sec:bgr_relwork}): encoder $E$, quantizer $Q_C$ with codebook $C$, and decoder $D$. 
Let $D_0,E_0$ be the original weights of $D$ and $E$. 
To avoid costly retraining of $\mathcal{M}$, we must keep $(E,Q_C,C)$ fixed; otherwise, we risk modifying the codebook semantics (directly or by changing how images are encoded by $E$), which harms the autoregressive model.
Thus, we propose to only finetune $D$ and an encoder \emph{replica} $E'$ (initialized to $E_0$).
$E'$ is used only for watermark detection, while the original $E$ may be used to condition $\mathcal{M}$ on images.
Unlike usual VQ training that promotes FCC, we optimize RCC: we aim to learn a decoder $D$ whose outputs $E'$ can reliably invert. 

\paragraph{Finetuning objectives.}
We first precompute tokenizations $s$ from a set of images, which we use as our training data.
We encourage RCC by minimizing the following loss:
\begin{align} \label{eq:rcc}
\loss_{\text{RCC}}(s) =\expectation_{a \sim \augset} \| \hat{z} - E'(a(D(\hat{z})))  \|_2^2.
\end{align}
Its goal is to match the original \emph{hard latents} $\hat{z}=C_{s}$ to \emph{soft latents} obtained after detokenization and encoding using $E'$.
To ensure RCC holds robustly even under transformations, we uniformly sample an augmentation $a \sim \augset$ with preset probability $p_{\text{aug}}$ in each training step, or set it to identity otherwise.
Our augmentation set $\augset$ includes valuemetric (brighten, contrast, JPEG) and weak geometric transformations (e.g.\ $\pm1^\circ$ rotation), with strength ramped up over training (see \aref{app:expdetails:rccft}).

To retain decoder quality we introduce a regularization that keeps the effect of $D$ close to the effect of its initial weights $D_0$ via a mixture of MSE and LPIPS perceptual loss~\citep{zhang2018unreasonable}:
\begin{align} \label{eq:reg}
\loss_{\text{reg}}(s) = \| D(\hat{z}) - D_{0}(\hat{z}) \|_2^2 + \loss_{\text{LPIPS}}(D(\hat{z}), D_0(\hat{z})).
\end{align}
We found this sufficient as a quality constraint and easier to train compared to loss functions that compare reconstructions to the original images or the use of adversarial discriminators.

We jointly train $D$ and $E'$ to minimize: $\loss(s) = \loss_{\text{RCC}}(s) + \lambda \cdot \loss_{\text{reg}}(s)$,
where $\lambda$ is a tradeoff hyperparameter.
In \cref{sec:experiments} we demonstrate that this efficiently boosts RCC and watermark robustness against valuemetric attacks, neural compression, and diffusion purification~\citep{nie2022diffusion}.
\input{figures/ft.tex}

%% file: figures/ft.tex
\begin{figure}[t]
    \centering
    \includegraphics[width=0.95\linewidth, clip, trim={0 6.2in 1in 0}]{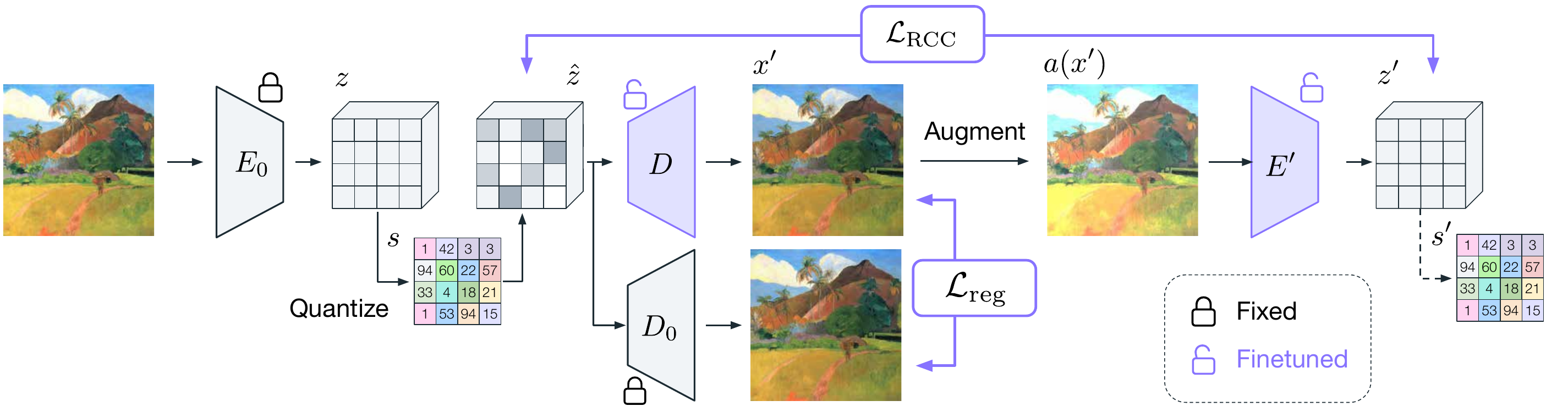}
    \caption{A replica $E'$ of the encoder and the decoder $D$ are jointly trained to improve reverse-cycle consistency, \ie make $E'(D(\hat{z}))$ close to $\hat{z}$ for most generations of the autoregressive model $\mathcal{M}$, even under transformations.
    }
    \label{fig:ft} 
\end{figure}

%% file: sections/4c-sync.tex
\subsection{Post-hoc Watermark Synchronization} \label{ssec:method:sync}

Semantic-preserving geometric transformations (e.g., flips) easily change image tokenization as each token loosely corresponds to a local image patch.
Therefore, RCC finetuning alone cannot recover the watermark. 
One could run the watermark detector on multiple transformed image copies (rescaled, rotated, etc.), but this is costly and significantly inflates false positives as noted in prior work~\citep{kirchenbauer2023watermark}.

\paragraph{Localization as synchronization.}
To address this in a more practical way, we repurpose localized watermarks (see \cref{sec:bgr_relwork}) as a synchronization signal.
More precisely, we locally embed a fixed set of messages whose detection estimates the applied transform, which we then aim to invert before applying the original watermark detector.  
To not degrade original performance, our watermark should be robust to the addition of this signal, which we verify in~\cref{sec:experiments}.
Detecting this signal could in principle be taken as evidence that the image is watermarked, as in some of the prior post-hoc watermarking schemes that explore synchronization~\citep{guo2023practical,luo2022leca}.
However, applying the original watermark detector is still necessary to obtain theoretically grounded p-values that can be combined with other samples across modalities as described above.
\looseness=-1 Moreover, as we will see in \cref{sec:experiments}, post-hoc watermarks are generally much more brittle to adversarial purification compared to the approach we propose.

\paragraph{Reverting transformations.}
In \cref{fig:sync}, we show our instantiation of this idea, where we embed four 32-bit synchronization messages \{\textcolor{magenta!90}{$0^{32}$}, \textcolor{red!95}{$0^{16}1^{16}$}, \textcolor{teal!90}{$1^{16}0^{16}$}, \textcolor{blue!95}{$1^{32}$}\} via the method of \citet{sander2024watermark} into the four image quadrants.
We observe that the original watermark is well preserved (with p-value $p=5\cdot10^{-38}$) after adding the synchronization signal.
However, a horizontal flip shuffles tokens and breaks detection ($p=0.66$).
To identify this, we apply an algorithm that searches over a grid of rotation angles, and for each fits the best axis-aligned pair of orthogonal lines that separate the four messages.
This is sufficient to identify and revert flips, rotations, and crops followed by upscaling to the respective model's original generation size, which we revert by downscaling and padding appropriately.
We note that our synchronization pattern (quadrants) assumes crops that preserve one corner---as we discuss in~\cref{sec:conclusion}, the same idea can be directly extended to arbitrary crops by using a more elaborate synchronization pattern. 
In our example in~\cref{fig:sync}, our procedure detects that a flip was applied and restores $p=5\cdot10^{-38}$.
Full algorithm and more examples are deferred to \aref{app:syncdetails:image}.

\looseness=-1
In the following, we empirically show that synchronization enhances geometric robustness, complementing RCC finetuning. 
This step is further aided by the use of small geometric augmentations during RCC finetuning, as they effectively compensate for minor errors in our transformation estimates.

\input{figures/sync.tex}

%% file: figures/sync.tex
\begin{figure}[t]
  \centering
  \includegraphics[width=0.95\linewidth, clip, trim={0.3in 6.8in 0.6in 0}]{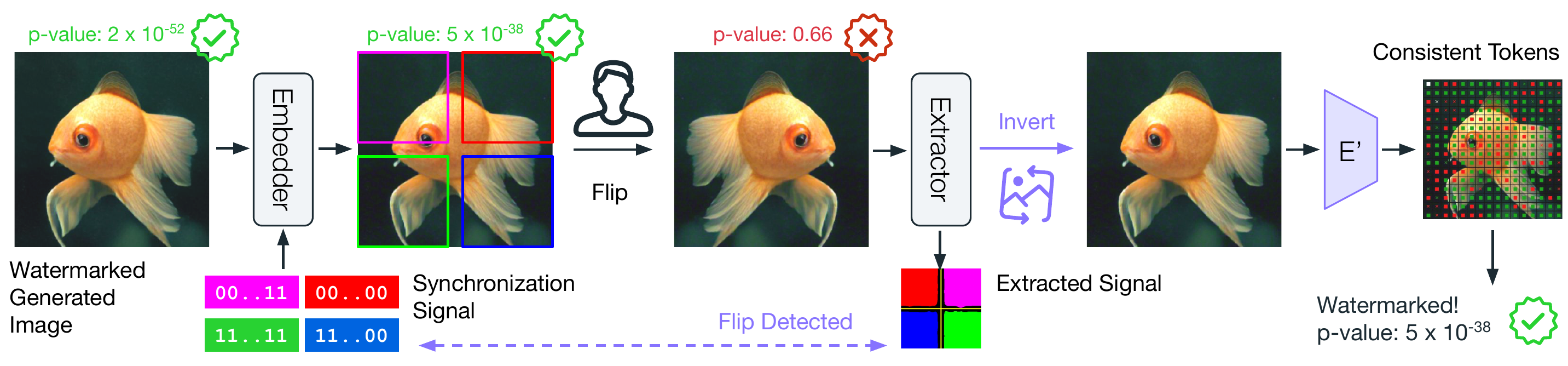}
  \caption{Watermark synchronization. Localized messages are embedded into a generated watermarked image and later used to discover the unknown transformation and revert it, which recovers the original watermark.}
  \label{fig:sync}
\end{figure}

%% file: sections/5-experiments.tex
\section{Experimental Evaluation}\label{sec:experiments}
In \cref{ssec:experiments:main}, we measure the effect of RCC finetuning (\cref{ssec:method:ft}) and the synchronization layer (\cref{ssec:method:sync}) on RCC, quality, and the power of our watermark.
\cref{ssec:experiments:robustness} studies robustness under common transformations and attacks, while \cref{ssec:experiments:interleaved} studies joint watermarking of text and images.
Additional experimental details and results are given in \aref{app:expdetails} and \aref{app:moreresults}, respectively.

\paragraph{Setup.}
We consider three autoregressive image generation models. 
First, the class-conditional ImageNet transformer from~\citet{esser2021taming} that generates images at resolution $256 \times 256$ with a VQGAN tokenizer with $|V|=16384$ and downsampling factor $f=16$, denoted \taming below.
Next, the $7$B variant of the mixed-modal \cham~\citep{team2024chameleon} that can generate interleaved text and $512 \times 512$ images, with $|V|=8192$ and $f=16$.
Finally, \rarxl, the $955$M configuration of the state-of-the-art \textsc{RAR} model~\citep{yu2024randomized}, generating $256 \times 256$ images conditioned on an ImageNet class, as \taming, with $|V|=1024$ and $f=16$.
In each experiment, we generate $1000$ samples per model ($100$ samples per each of $10$ ImageNet classes for \taming and \rarxl, and $1000$ COCO prompts for \cham).

We evaluate $4$ variants of our method: \base, which uses original models and tokenizers, \ft and \ftaug, which apply the same watermark after RCC finetuning (\cref{ssec:method:ft}) without and with augmentations in training, respectively, and \ftaugsync, which also uses our watermark synchronization (\cref{ssec:method:sync}) on top of augmented-RCC finetuning.
We use $\delta=2$ and $\gamma=0.25$ in all experiments, $h=1$ for \taming, \rarxl, and \cham on text, and $h=0$ for \cham on images.
\looseness=-1
We finetune models on tokens derived from $50{,}000$ ImageNet training samples for $10$ epochs ($2$h on $16$ {V100} for \taming, $2.5$h on $8$ H200 for \cham, and $0.5$h on $8$ H200 for \rarxl).
Computational efficiency is discussed further in \aref{app:moreresults:efficiency}.
 
\subsection{Reverse Cycle-Consistency, Watermark Power, and Generation Quality} \label{ssec:experiments:main}
 
The key question raised in~\cref{sec:method} is if our proposed finetuning procedure can alleviate the lack of reverse cycle-consistency (RCC) in image tokenizers, and in turn improve watermark power. 
We also measure the effect of watermarking and finetuning on generation quality.
In the following, we present results with \taming.
In~\aref{app:moreresults:powerandrobustnessmoremodels}, we repeat the same experiments on \cham and \rarxl with the same finetuning hyperparameters (see~\aref{app:expdetails:rccft}) and reach similar conclusions.

\paragraph{Finetuning improves RCC.}
We generate $1000$ class-conditioned ImageNet samples using each of our $4$ variants, and measure token match (TM, \cref{eq:tm}) between the generated tokens and those obtained by re-tokenizing the image.
In our results in \cref{fig:power} (left), we observe that TM is consistently below $0.8$, as previously seen in~\cref{tab:tokens}, while for all finetuned variants it is generally \emph{above} $0.8$.
This demonstrates that finetuning is successful in improving RCC.
\textsc{Augs} and \textsc{Sync} slightly reduce TM on unmodified images, but significantly increase robustness (see \cref{ssec:experiments:robustness}).

\paragraph{Finetuning improves watermark power.}
In \cref{fig:power} (right), we report the true positive rate (TPR) of the watermark detector for different false positive rates (FPR).
The \base variant already has practically viable power, achieving TPR of ${\approx}{1}$ at FPR of $10^{-2}$ (dashed line), the setting commonly considered in prior work~\citep{wen2023tree,ci2024ringid,zhao2023provable,Dathathri2024}.
However, RCC gains directly translate to improvements in watermark power: for all $3$ variants, the TPR at lower FPR settings is significantly higher.

\paragraph{Watermarking and finetuning do not harm generation quality.}
To measure the quality of the generated samples, we compute FID~\citep{fid} on $50{,}000$ generations ($50$ per ImageNet-1K class) for all variants.
We find that none of \base, \ft, and \ftaug have FID above $16.7$, which is the FID of an \emph{unwatermarked} \base model.
This confirms that our watermark preserves generation quality even after finetuning.
The FID of \ftaugsync is $17.3$, a minor increase inherited from the localized watermark used for synchronization.
We complement this with a comparison of the original and finetuned decoders using PSNR in~\aref{app:moreresults:decoderquality} and with qualitative samples on all models in \aref{app:qualitative}.

\subsection{Watermark Robustness} \label{ssec:experiments:robustness}
An important requirement for a generative model watermark is robustness to common domain-specific transformations, as well as to removal attacks, which have shown to be effective against other watermarks~\citep{fernandez2023stable,saberi2023robustness,an2024waves}.
To evaluate this, in~\cref{fig:robustness} we report the watermark TPR for a fixed FPR of $1\%$ on a range of transformations of different strength, in the same setting as in~\cref{fig:power}, using \taming (the equivalent results for \cham and \rarxl are deferred to~\aref{app:moreresults:powerandrobustnessmoremodels}).

We summarize the results of this experiment for all three models in~\cref{tab:baselines}, where, as in prior work~\citep{wen2023tree}, we average TPR over a set of (transformation, parameter) pairs, detailed in~\aref{app:expdetails:main}.
Our main conclusions, discussed next, hold across all models, despite the variations in \eg transformer or tokenizer codebook size, which demonstrates the transferability of our approach. 

\paragraph{Finetuning enables valuemetric and attack robustness.}
In~\cref{fig:robustness}, we see that the watermark is fragile to valuemetric and geometric transformations when used on \base.
When we use it on \ftaug, robustness to valuemetric transformations greatly improves, validating our focus on RCC finetuning. 
Surprisingly, finetuning also improves robustness to (i) neural compressors~\citep{mbt18,cstk20,bmshj18} of different strengths (see details in~\aref{app:expdetails:main}), including FLUX and SD VAEs~\citep{rombach2022high,sdxl,flux2024,dcae}, and (ii) the challenging diffusion purification attack~\citep{nie2022diffusion}.
We remark that high values such as $t=0.3$ were found to excessively alter images, making this regime less relevant~\citep{saberi2023robustness}.
\looseness=-1
This effect holds for \rarxl but is less pronounced for \cham in~\cref{tab:baselines}, where our watermark is already robust to these attacks even without RCC finetuning, likely due to the detector scoring more tokens for larger images.

\input{figures/power.tex}
\paragraph{Synchronization enables geometric robustness.}

Geometric robustness (\emph{bottom left} in \cref{fig:robustness}) remains low across all models, even with \ftaug, which is expected as autoregressive models cannot preserve token sequences under semantic changes like flips, and the watermark is lost.
This motivated our synchronization layer (\cref{ssec:method:sync}), which estimates and reverts such changes. 
As~\cref{fig:robustness} and \cref{tab:baselines} show, synchronization significantly improves geometric robustness while preserving watermark power on unmodified images.
However, this comes at the cost of a minor drop in valuemetric robustness. 
Namely, moderate valuemetric transformations sometimes disrupt the synchronization signal, causing us to estimate and revert a non-existent geometric transformation.
Such a mistake corrupts most tokens and breaks the watermark. 
This does not happen for strong valuemetric transformations that often fully destroy the synchronization signal, as in this case we keep the image intact.
A more elaborate and robust synchronization layer could likely minimize this undesirable effect while further extending the scope of supported transformations; we propose several concrete directions in~\cref{sec:conclusion}.
\input{figures/robustness.tex}

\input{tables/baselines_small.tex}
\paragraph{Comparison to post-hoc methods.} 
As noted above, no prior work targets watermarking of autoregressive image generation.
Thus, in~\cref{tab:baselines} we compare to \emph{post-hoc} methods~\citep{CIN,MBRS,bui2023trustmark,sander2024watermark} applied on top of generated images.
While they are comparably or more robust than our watermark on valuemetric transformations, each post-hoc watermark is either fully removed by geometric ones or not robust to attacks (adversarial purification and neural compressors).
Further, our watermark yields p-values grounded in randomness, with theoretical guarantees inherited from LLM watermarking~\citep{kirchenbauer2024on,zhao2023provable,fernandez2023three} and empirically validated in~\aref{app:moreresults:fpr}.
In contrast, post-hoc methods use neural extractors to recover messages and may introduce bias in their p-value estimators~\citep[App. B.5]{fernandez2023stable} and~\citep[App. B]{sanroman2024proactive}.
Finally, as a token-level generation-time method, our method is the only one able to watermark content via semantic modifications (see \cref{fig:gimmick}).
\looseness=-1
In~\aref{app:moreresults:gentime} we provide an additional comparison to generation-time watermarks for diffusion models, despite their inapplicability to our target models.

\subsection{Joint Watermarking of Interleaved Modalities} \label{ssec:experiments:interleaved}
Finally, we explore joint watermarking of multiple modalities generated by the same autoregressive model.
\cref{eq:pvalue} shows that scoring more tokens that are all equally watermarked improves power.
However, acquiring more tokens is not always possible---in such cases, jointly watermarking multiple modalities may be necessary to reliably detect the watermark.
For example, consider that Alice aims to prove if an online article was generated by her model $\model$. 
To simulate this, we run \cham (\ftaug) in interleaved mode on $1000$ prompts to produce text and an image, and we model attempts to conceal the use of $\model$ by randomly changing text tokens (a proxy for paraphrasing~\citep{krishna2023paraphrasing}). 

\definecolor{purple1}{HTML}{B99BCA}
\definecolor{purple2}{HTML}{8A58A7} 
\definecolor{purple3}{HTML}{533464}

\paragraph{Benefits of joint watermark detection.}  
The orange line in~\cref{fig:interleaved} shows TPR at $1\%$ FPR when only text is watermarked, quickly degrading with more text corruption. 
As Alice uses all text available to her in this scenario, it is hard for her to improve detection.

\input{figures/interleaved.tex}
However, if both text and the image were originally watermarked with our method, detection on combined tokens as described in \cref{sec:method} significantly boosts watermark power (\textcolor{purple1}{top purple}, \emph{Clean}).
At $10\%$ text corruption TPR improves from ${\approx}{0.9}$ to $1.0$, and stays above $0.94$ even in the hardest case, where text-only TPR drops to ${\approx}{0}$.
Alice also gets a rigorous p-value, which would be hard if modalities were watermarked separately.
As our method is robust to moderate image transformations, a similar trend holds when adding Gaussian noise with $\sigma=0.1$ (\textcolor{purple2}{middle purple}, \emph{Weak Noise}). 

Importantly, there is a tradeoff---\emph{integrating a weak watermarking signal can degrade detection}. 
We see this for $\sigma=0.3$ (\textcolor{purple3}{bottom purple}, \emph{Strong Noise}), where TPR drops below $0.6$ at $10\%$ corruption, \ie text-only detection is preferable.
In a related study in~\aref{app:moreresults:entropy} we analyze the entropy of text and image tokens, and in~\aref{app:interleaved} we provide a more detailed discussion of joint watermarking and present extended results and example interactions with \cham.
\looseness=-1
Inspired by this, in the following section we explore the extension of our method to additional modalities.

%% file: figures/power.tex
\begin{figure}[t]
    \centering
    \resizebox{0.8 \textwidth}{!}{
    \includegraphics[width=0.4\textwidth]{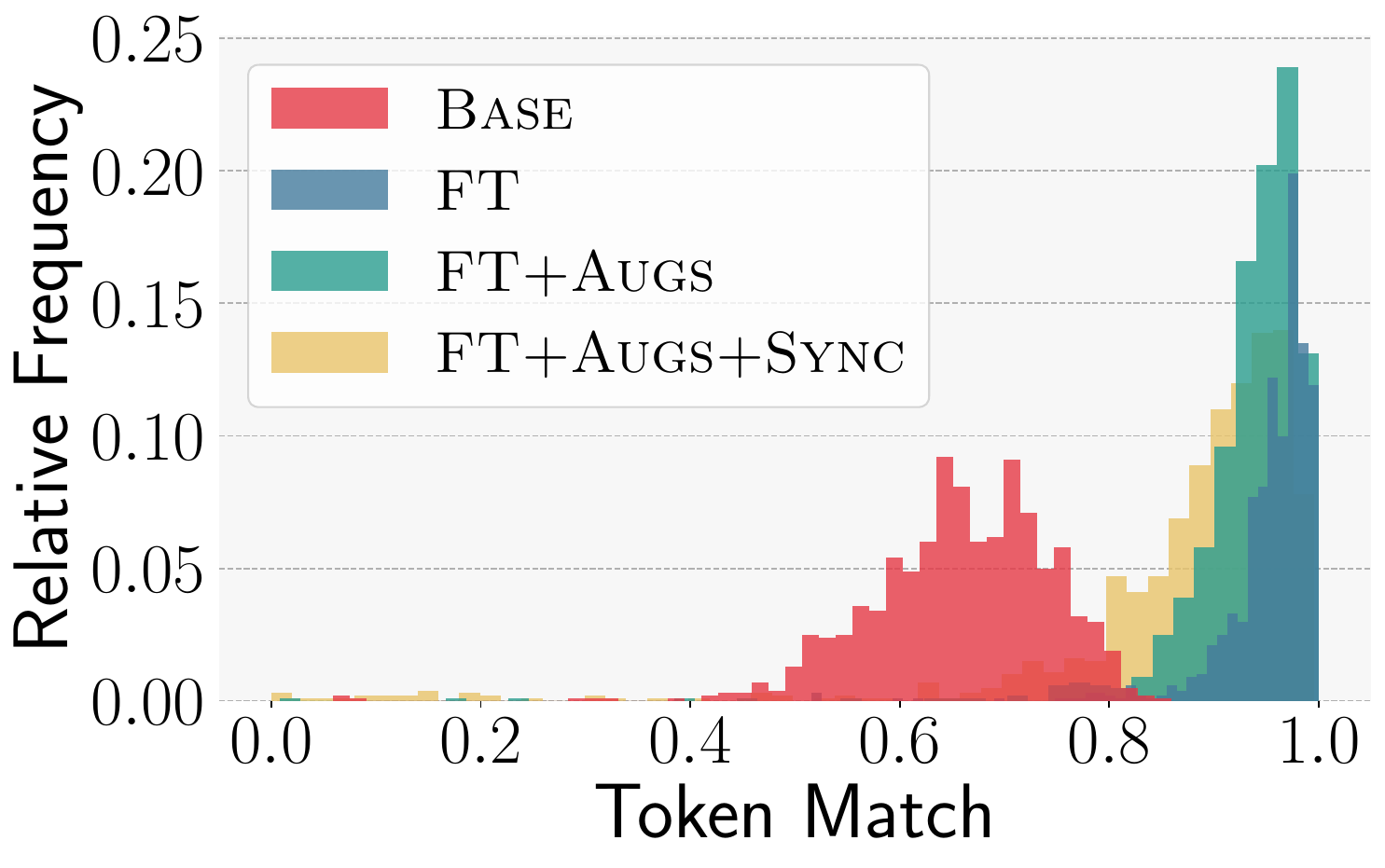}
    \quad \quad \quad \quad 
    \includegraphics[width=0.4\textwidth]{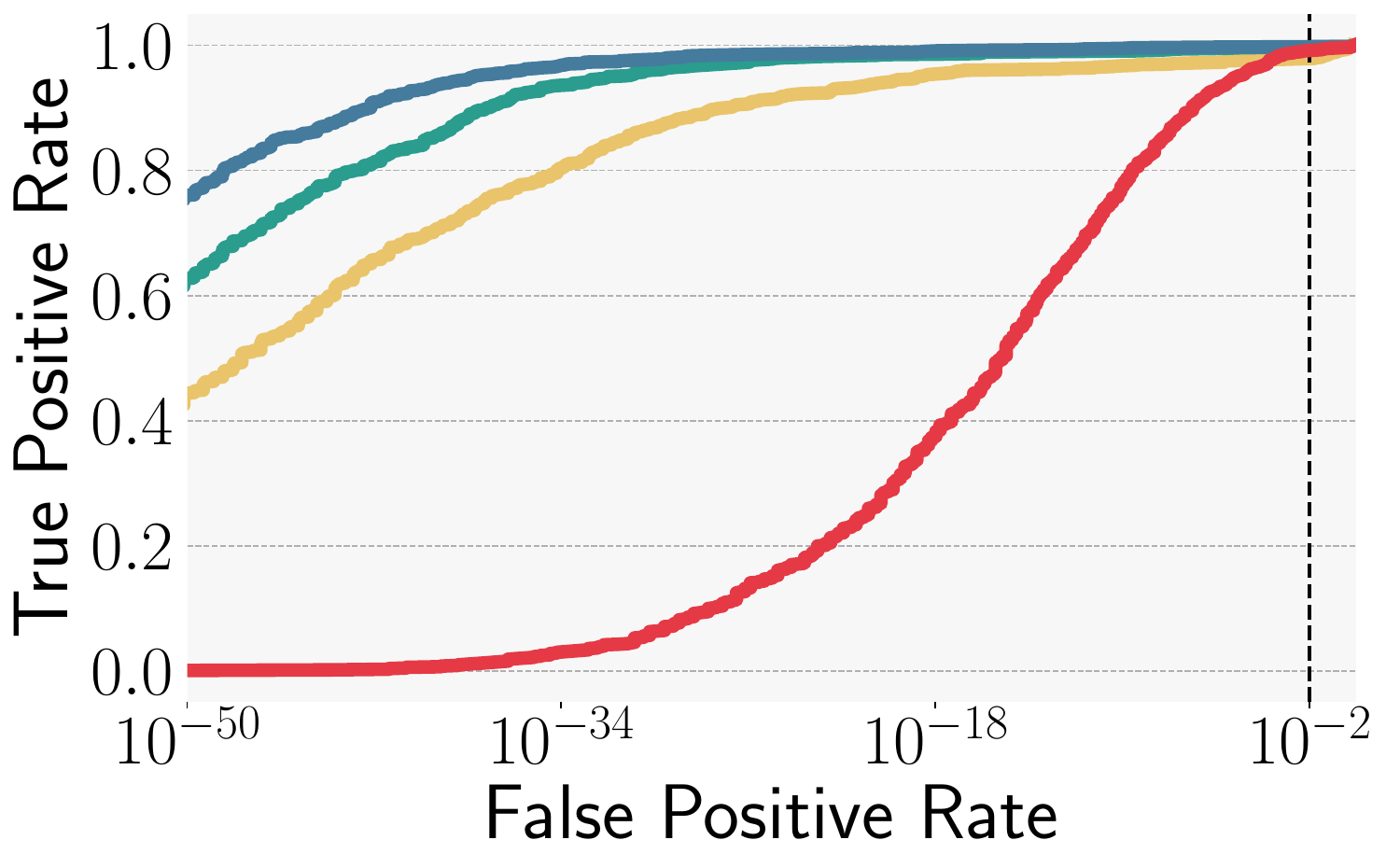}
    }
    \caption{
        \emph{Left}: Finetuning improves token match (\cref{eq:tm}) between original and re-tokenized image tokens.
        \emph{Right}: All variants achieve TPR~${\approx}{1}$ at FPR of $1\%$. Finetuning further boosts detection in low-FPR settings.
    }
    \label{fig:power}
\end{figure}  

%% file: figures/robustness.tex
\begin{figure}[t!]
    \centering
    \includegraphics[width=\textwidth]{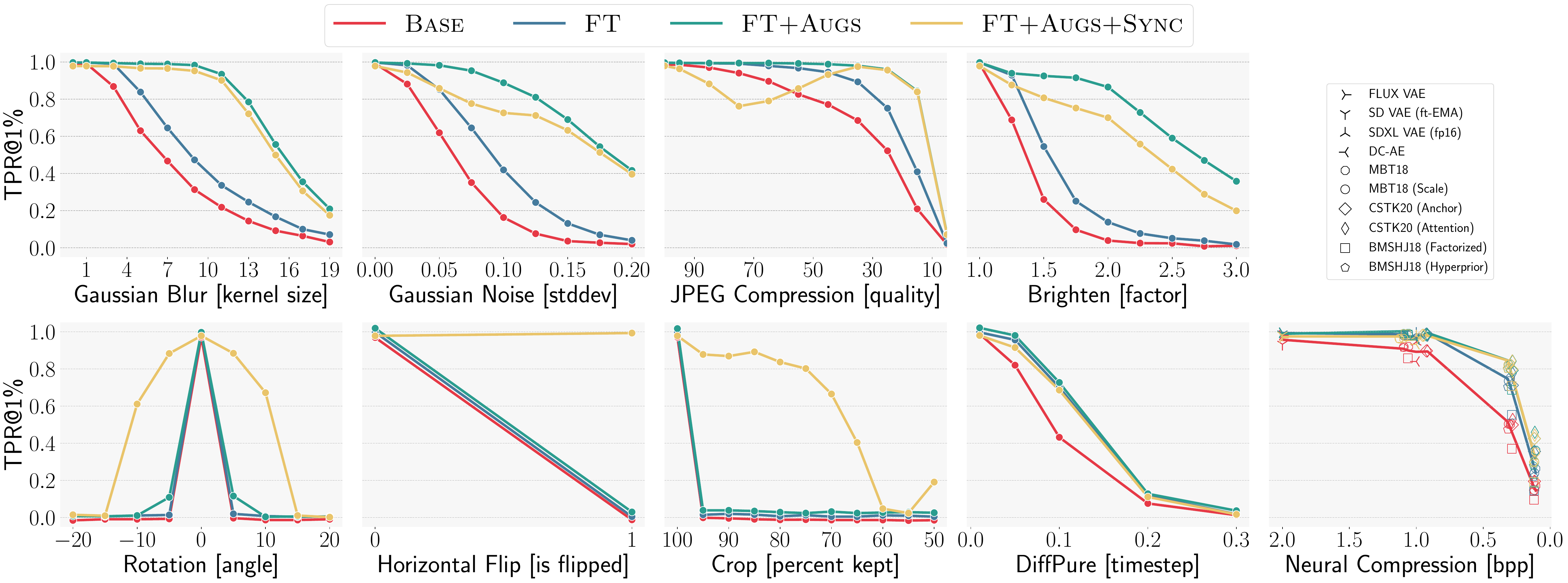}
    \caption{\emph{Top:} RCC finetuning improves robustness to valuemetric transformations. \emph{Bottom Left:} Watermark synchronization unlocks robustness to geometric transformations. \emph{Bottom Right:} Our watermark is also fairly robust to realistic strengths of diffusion purification~\citep{nie2022diffusion,saberi2023robustness} and neural compression~\citep{mbt18,cstk20,bmshj18,rombach2022high,sdxl,flux2024,dcae}.}
    \label{fig:robustness}
\end{figure}

%% file: tables/baselines_small.tex
\begin{table}[t]
    \centering 
    \caption{TPR at 1\% FPR of our watermark and post-hoc baselines, under valuemetric (\textbf{Val.}) and geometric (\textbf{Geo.}) transformations, adversarial purification attacks (\textbf{Adv.}) and neural compression (\textbf{NC}). Scores (see~\aref{app:expdetails:main}) below $0.6$ are marked \textcolor{badred}{red}.
    RCC finetuning and the synchronization layer lead to a strong and robust watermark.}
    \newcommand{\fivecol}[1]{\multicolumn{5}{c}{#1}}  
    \resizebox{\linewidth}{!}{%
    \begingroup 
    \setlength{\tabcolsep}{4pt} %
    \begin{tabular}{cl
        x{1}{2}
        x{1}{2}
        x{1}{2}
        x{1}{2}
        x{1}{2}
        @{}p{0.5cm}@{}
        x{1}{2}
        x{1}{2}
        x{1}{2}
        x{1}{2}
        x{1}{2}
        @{}p{0.5cm}@{}
        x{1}{2}
        x{1}{2}
        x{1}{2}
        x{1}{2}
        x{1}{2}
        }
        \toprule

        && \fivecol{\taming~{\small ($256\times256$)}} && \fivecol{\cham~{\small($512\times512$)}} && \fivecol{\rarxl~{\small ($256\times256$)}} \\ 
        \cmidrule(l{3pt}r{0pt}){3-7}
        \cmidrule(l{0pt}r{0pt}){9-13}
        \cmidrule(l{0pt}r{3pt}){15-19}
        \textbf{} & \textbf{} 
        & {\textbf{None}} 
        & {\textbf{Val.}} 
        & {\textbf{Geo.}} 
        & {\textbf{\shortstack[c]{Adv.}}}
        & {\textbf{\shortstack[c]{NC}}} 
        && {\textbf{None}} 
        & {\textbf{Val.}} 
        & {\textbf{Geo.}} 
        & {\textbf{\shortstack[c]{Adv.}}}
        & {\textbf{\shortstack[c]{NC}}}
        && {\textbf{None}} 
        & {\textbf{Val.}} 
        & {\textbf{Geo.}} 
        & {\textbf{\shortstack[c]{Adv.}}}
        & {\textbf{\shortstack[c]{NC}}} \\

        \midrule
        \multirow{4}{*}{\shortstack[c]{Ours}} & \baseTABLE & 0.99 & \textcolor{badred}{0.26} & \textcolor{badred}{0.01} & \textcolor{badred}{0.43} & \textcolor{badred}{0.48} && 0.98 & \textcolor{badred}{0.50} & \textcolor{badred}{0.02} & 0.80 & 0.82 && 0.98 & \textcolor{badred}{0.22} & \textcolor{badred}{0.01} & \textcolor{badred}{0.29} & \textcolor{badred}{0.27} \\
        & \ftTABLE & 1.00 & \textcolor{badred}{0.45} & \textcolor{badred}{0.01} & 0.70 & 0.71 && 0.99 & \textcolor{badred}{0.53} & \textcolor{badred}{0.03} & 0.85 & 0.87 && 1.00 & 0.74 & \textcolor{badred}{0.03} & \textcolor{badred}{0.58} & 0.76 \\
        & \ftaugTABLE & 1.00 & 0.92 & \textcolor{badred}{0.01} & 0.70 & 0.79 && 0.99 & 0.89 & \textcolor{badred}{0.02} & 0.82 & 0.88 && 1.00 & 0.98 & \textcolor{badred}{0.03} & 0.79 & 0.95 \\
        & \ftaugsyncTABLE & 0.98 & 0.83 & 0.82 & 0.69 & 0.80 && 0.97 & 0.76 & 0.64 & 0.81 & 0.86  && 1.00 & 0.89 & 0.82 & 0.78 & 0.94 \\
        
        \midrule

        \multirow{4}{*}{\shortstack[c]{Post-\\hoc}} & \textsc{CIN} & 1.00 & 0.96 & \textcolor{badred}{0.00} & \textcolor{badred}{0.03} & \textcolor{badred}{0.02} && 1.00 & 0.99 & \textcolor{badred}{0.00} & \textcolor{badred}{0.14} & \textcolor{badred}{0.16} && 1.00 & 0.95 & \textcolor{badred}{0.00} & \textcolor{badred}{0.04} & \textcolor{badred}{0.01} \\
        & \textsc{MBRS} & 1.00 & 0.98 & \textcolor{badred}{0.02} & \textcolor{badred}{0.36} & \textcolor{badred}{0.31}&& 1.00 & 0.99 & \textcolor{badred}{0.02} & \textcolor{badred}{0.27} & \textcolor{badred}{0.56} && 1.00 & 0.99 & \textcolor{badred}{0.01} & \textcolor{badred}{0.38} & \textcolor{badred}{0.34} \\
        & \textsc{Trustmark} & 1.00 & 0.98 & 0.75 & \textcolor{badred}{0.40} & 0.86  && 1.00 & 0.97 & 0.74 & 0.64 & 0.99  && 1.00 & 0.99 & 0.75 & \textcolor{badred}{0.42} & 0.90 \\
        & \textsc{WAM} & 1.00 & 0.89 & 0.98 & \textcolor{badred}{0.06} & \textcolor{badred}{0.02} && 1.00 & 0.97 & 0.95 & \textcolor{badred}{0.26} & \textcolor{badred}{0.48} && 1.00 & 0.92 & 0.98 & \textcolor{badred}{0.08} & \textcolor{badred}{0.02} \\

        \bottomrule
    \end{tabular}
    \endgroup
    }
    \vspace{-0.5em}
    \label{tab:baselines}
\end{table} 

%% file: figures/interleaved.tex
\begin{wrapfigure}[17]{r}{0.4\textwidth} \centering
		\vspace{-0.05in}
		\includegraphics[width=\linewidth]{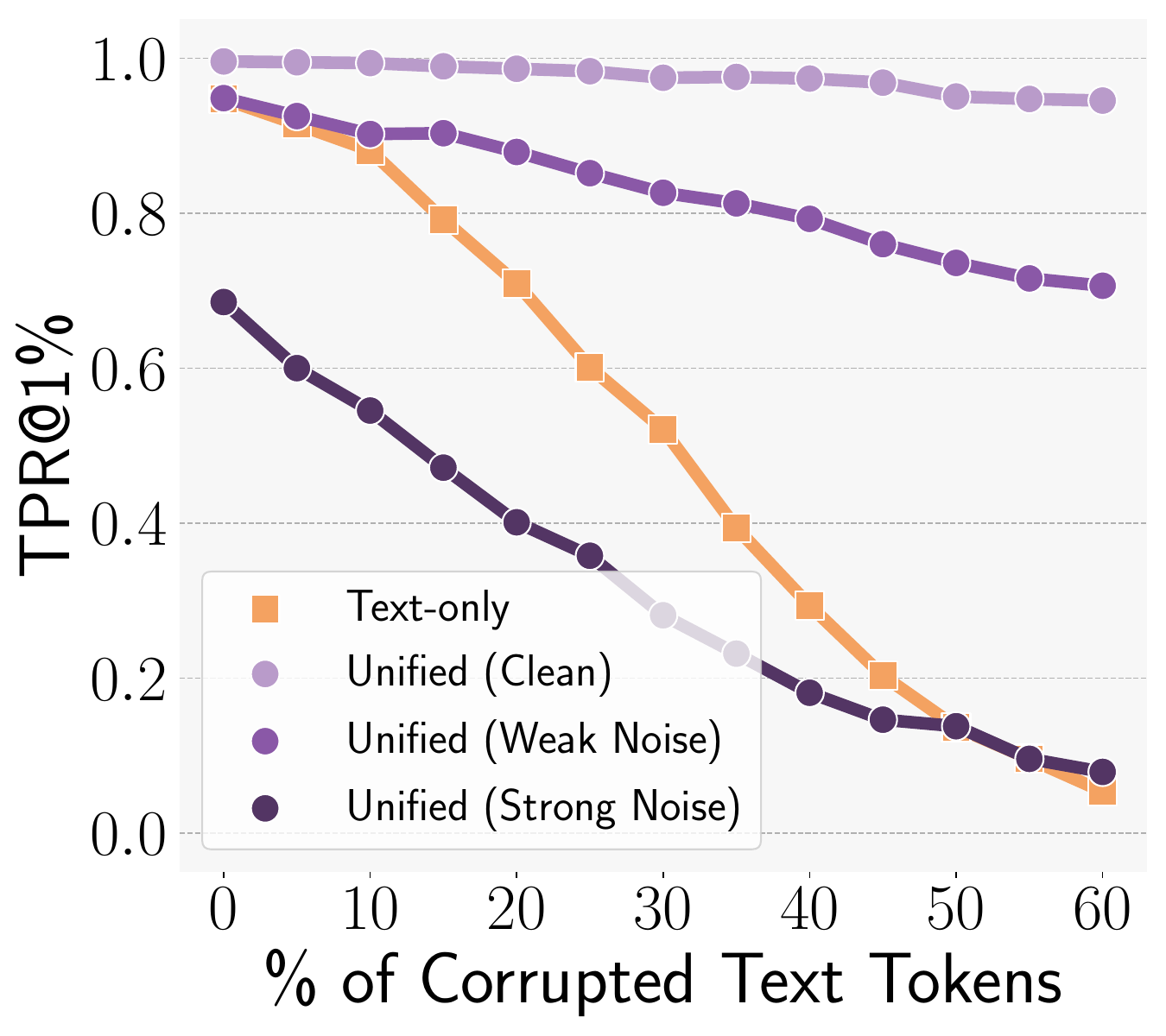}
        \caption{Joint watermark detection on text and image generations. 
        }
  \label{fig:interleaved} 
\end{wrapfigure}
    

%% file: sections/6-audio.tex
\section{Extension to Additional Modalities: Audio Case Study}\label{sec:audio}

In this final section we ask: \emph{Can our approach be extended to other modalities?} 
In an attempt to answer this, we conduct a preliminary study on autoregressive {audio} generation~\citep{Borsos2022AudioLMAL,copet2024simple,zhang2024speechlm,zhang2023speechgpt,nguyen2025spirit,defossez2024moshi} focusing on \moshi~\citep{defossez2024moshi}, a transformer-based speech-text foundation model.
We observe similar challenges and main results as for images, while noting several important differences.
We defer many details to \aref{app:consistency:audio} (RCC experiment), \aref{app:syncdetails:audio} (synchronization details), \aref{app:expdetails:audio} (experimental details), and present extended results of audio experiments in \aref{app:moreresults:audio}.

\paragraph{Audio tokenization.}
\moshi's tokenizer (\textsc{Mimi}) relies on \emph{residual} vector quantization (RVQ)~\citep{rvq,kumar2023high,soundstream,defossez2022high}. 
RVQ iteratively quantizes the residuals of the previous quantizer, such that $s_i = (s_{i}^{1},.., s_{i}^{K})$ for $K$ different codebooks $C^{1}, .., C^{K}$ ($K$ \emph{streams})
Each token here represents $\approx$80\,ms.

\paragraph{RCC in audio.} 
\citet[Sec. 6.4]{defossez2024moshi} already investigate the RCC properties of \moshi's tokenizer.
They observe that the first stream is somewhat cycle-consistent, while this degrades for later ones.
We obtain similar results and also show that TM further worsens under transformations (highpass, speedup). 
For instance, in an experiment on $1000$ generated sequences, we measure average $\text{TM}$ of $0.36$ (original), $0.21$ (highpass $500$~Hz), and $0.16$ ($1.1 \times$ speedup).
This motivates an adaptation of RCC finetuning~(\cref{ssec:method:ft}) and synchronization~(\cref{ssec:method:sync}) to audio.

\paragraph{RCC finetuning.} 
To instantiate the finetuning procedure from~\cref{ssec:method:ft} we make the following changes to~\cref{eq:rcc,eq:reg}.
We use the pre-projection soft latents as target since the quantization is done in a projected space~\citep{kumar2023high}.
We replace LPIPS with multi-resolution STFT loss~\citep{yamamoto2020parallel}. 
Finally, during training we apply augmentations from a set $\mathcal{A}$ that includes audio-specific valuemetric edits (high/low/bandpass, gaussian/pink noise, etc.) as well as small ($1$-$10$\,ms) time-frequency shifts.

\paragraph{Synchronization.} 
In contrast to images, no localized audio embedder proved robust: we found~\citet{sanroman2024proactive} to not be precise enough under time-frequency edits (see~\aref{app:syncdetails:audio}).
Thus, our following experiment focuses primarily on the influence of finetuning.
\looseness=-1
Future work on audio-specific localized watermarking could improve on this, motivated by the synchronization usecase.

\paragraph{Watermarking multiple streams.}
Early RVQ streams are more reverse cycle-consistent and thus more likely to preserve the watermark signal. 
We observe however that limiting watermarking to a single stream provides too few tokens for reliable statistical testing, significantly increasing p-values.
On the other hand, watermarking all streams introduces noise due to the lack of RCC in later codes. 
Empirically, we find that watermarking the first four streams achieves a good balance.

\paragraph{Experimental setting.}
We perform RCC finetuning to obtain \ft and \ftaug using the VoxPopuli~\citep{wang2021voxpopuli} dataset, such that final PESQ~\citep{rix2001perceptual} (perceptual speech quality metric) is $4.3$ w.r.t. \base samples.
For evaluation, we generate $12$s watermarked audio samples with \moshi using $1000$ text prompts generated by \textsc{Llama3.1-8B-Instruct} and synthesized to audio with \textsc{SeamlessV2}~\citep{barrault2023seamlessm4t}.
We set $h=0$ and $\delta=2$ and watermark the first four audio streams as described above.
As in~\cref{sec:experiments}, we evaluate TPR at $1\%$ FPR and quality, for which in this case we use the MOSNet~\citep{defossez2024moshi} metric.

\input{tables/audio.tex}
\paragraph{Results.}
We present the results in \cref{tab:audio}.
As for images, we do not observe notable quality degradation due to watermarking, and obtain high watermark strength even without RCC finetuning.
We find that \base already has nonzero time-frequency robustness, likely due to non-semantic streams being used to carry the watermark. 
Finetuning without augmentations, interestingly, impairs time-frequency RCC, which is later recovered by \ftaug.
We hypothesize that this drop is due to catastrophic forgetting~\citep{kirkpatrick2017overcoming} as the model learns to detokenize the audio in a way that is not robust to time-frequency transformations.
This suggests that augmentations are a key component of finetuning, matching our results on images.
Yet, while \ftaug improves valuemetric robustness, it fails to raise time-frequency robustness to satisfactory levels, leaving this question open for future work.
Finally, while we are not aware of audio equivalents of diffusion purification used in~\cref{sec:experiments}, we evaluate robustness to neural compression (DAC~\citep{kumar2023high} and EnCodec~\citep{defossez2022high}).
We observe results comparable to post-hoc \textsc{AudioSeal}~\citep{sanroman2024proactive}, even though in contrast to \textsc{AudioSeal} we do not explicitly train against EnCodec during finetuning.

%% file: tables/audio.tex
\begin{wraptable}[11]{r}{0.49\linewidth}
    \vspace{-1.1em}
    \centering
    \caption{TPR at 1\% FPR and MOSNet~\citep{mosnet} of our watermark, under valuemetric (\textbf{Val.}), time-frequency (\textbf{Time}) transformations, and neural compression (\textbf{NC}).
    MOSNet is $3.80$ for unwatermarked generation.
    }
    \newcommand{\fourcol}[1]{\multicolumn{4}{c}{#1}}  
    \definecolor{badred}{rgb}{0.8,0.2,0.2}
    \resizebox{1.0\linewidth}{!}{%
    \begingroup 
        \setlength{\tabcolsep}{5pt}
        \begin{tabular}{cl
                x{1}{2}
                x{1}{2}
                x{1}{2}
                x{1}{2}
                | @{\hspace{4pt}}x{1}{2}
                }
            \toprule
             &  & {\textbf{None}} & {\textbf{Val.}} & {\textbf{Time}} & {\textbf{NC}} & {\textbf{MOSNet}} \\
            \midrule
            \multirow{3}{*}{Ours}
                & \baseTABLE    & 0.97 & 0.62 & \textcolor{badred}{0.24} & 0.80 & 3.82 \\
                & \ftTABLE      & 0.99 & 0.64 & \textcolor{badred}{0.14} & 0.84 & 3.83 \\
                & \ftaugTABLE   & 0.99 & 0.80 & \textcolor{badred}{0.24} & 0.86 & 3.73 \\
            \midrule
            Post-hoc & \textsc{AudioSeal} 
            & 1.00 & 0.84 & \textcolor{badred}{0.55} & 0.85 & 3.78 \\
            \bottomrule
        \end{tabular}
    \endgroup
    }
    \label{tab:audio}
\end{wraptable}

%% file: sections/7-conclusion.tex
\section{Conclusion and Limitations} \label{sec:conclusion}
Our work successfully applies watermarking to the previously unexplored setting of autoregressive image generation, addressing low reverse cycle-consistency (RCC) through a custom finetuning stage and a synchronization layer.
Experiments demonstrate the power, robustness, and practicality of our watermark across a range of settings.
By broadening the scope of generative model watermarking, we believe this work takes an important step towards more reliable content provenance.

\paragraph{Limitations.}
Our method's scope could be extended in several important ways.
As noted in~\cref{sec:bgr_relwork}, we target the most prominent models that tokenize images via VQ. 
Our method is thus not applicable to models that use continuous representations or hybrids that combine autoregressive and diffusion models~\citep{zhou2024transfusion,fan2024fluid,li2024autoregressive,ma2024janusflow}.
Another dimension is modality: we present initial audio experiments in~\cref{sec:audio}, but this direction could be investigated further, \eg finding ways to improve time-frequency robustness.
Next, our synchronization relies on off-the-shelf localized watermarks to embed a fixed pattern (\emph{quadrants}, see \cref{fig:sync}), which is suboptimal for several reasons.
First, localized watermarks are trained to embed arbitrary patterns.
Training a custom synchronization layer\footnote{We explore this in our follow-up work \textsc{SyncSeal}~\citep{fernandez2025geometric}.}, potentially integrated with RCC finetuning, would be a more principled approach that could significantly improve robustness. 
Second, using more elaborate synchronization could both improve the quality of our transformation estimation step, but also expand its scope, \eg to support arbitrary crops as discussed in~\cref{ssec:method:sync}.
Finally, our method is not robust to combined removal attacks (to disrupt synchronization) and geometric attacks (to decrease token match)---to the best of our knowledge, this attack would also break most other contemporary watermarks.

%% file: sections/app-other.tex
\section{Ethical Statement}\label{app:ethical}

\subsection{Societal Impact}
Watermarking in general improves the traceability of content, be it AI-generated or not.
It can have positive consequences, for example when it is used to trace the origin of fake news or to protect intellectual property.
This traceability can also have negative consequences, for example when it is used to trace political opponents in authoritarian regimes or whistleblowers in secretive companies. 
Besides, it is not clear how to disclose watermark detection results, which may foster a closed ecosystem of detection tools. 
It may also exacerbate misinformation by placing undue emphasis on content that is either not detected, generated by unknown models, or authentic but used out of context.
We however believe that the benefits of watermarking outweigh the risks, and that the development of robust watermarking methods is a positive step for society.

\subsection{Environmental impact}
The cost of experiments and model training is high, though order of magnitude less than training the generative models themselves.
Finetuning the image tokenizer as done in the paper takes $\leq 32$~GPU-hours.
We also roughly estimate that the number of GPU-days used for running all our experiments is around $500$, \ie $12$k GPU-hours.
This amounts to total emissions in the order of 1 ton of CO$_2$eq.
Estimations are conducted using the \href{https://mlco2.github.io/impact#compute}{Machine Learning Impact Calculator} presented by~\citet{lacoste2019quantifying}.
Namely, using a default grid, we compute $250$W $\times$ $12000$h = $3000$ kWh $\times$ $0.3$ kg eq. CO2/kWh = $900$ kg eq. CO2.
We do not consider memory storage, CPU-hours, production cost of GPUs/CPUs, etc.

%% file: sections/app-watermarks.tex
\section{Technical Details of LLM Watermarking} \label{app:watermarks}

We here more thoroughly introduce LLM watermarking, following the notation in \cref{sec:bgr_relwork}.

\paragraph{Generation.} 
We consider an autoregressive model $\mathcal{M}$ generating a sequence of tokens $s = (s_1, s_2, \ldots, s_T)$, where each token $s_t$ is sampled from a probability distribution conditioned on the previous tokens $p(s_t | s_{<t})$. In practice, the model outputs a vector of logits $\ell \in \mathbb{R}^{|V|}$, where $V$ is the vocabulary (which we can assume in the most general case can contain text, audio or image tokens), which is transformed into a probability distribution $p = \text{softmax}(\ell / \tau)$, with $\tau$ being a temperature parameter.

The watermark scheme modifies the token selection process using a secret key $\xi$. A cryptographic hash function takes as input $h$ previous tokens $(s_{t-h}, \ldots, s_{t-1})$ (the context window) and the secret key $\xi$, producing a seed for a random number generator (RNG) that influences the choice of the next token $s_t$.

Two prominent LLM watermarking approaches are:

\begin{itemize}
    \item The method of \citet{kirchenbauer2023watermark} (KGW), which uses RNG to randomly partition the vocabulary $V$ into a greenlist $\mathcal{G}_t$ and a redlist $\mathcal{R}_t$, where $\mathcal{G}_t$ contains a proportion $\gamma$ of the vocabulary. The logit of each token in the greenlist is increased by $\delta > 0$, effectively boosting the probability of selecting tokens from the greenlist.
    
    \item The method of \citet{aaronson2023watermarking}, which uses a different approach based on the RNG to sample secret values for each token. 
    Although we do not present it in this work for simplicity, it could have been adapted in the same way to watermark autoregressive models.
\end{itemize}

In this paper we focus on KGW. As the first study of watermarking for autoregressive image models, choosing a relatively simple, well-studied and well-understood scheme allowed us to focus on key issues particular to our setting (\eg RCC). We do think that it would be interesting to explore other schemes and improvements of KGW as well---our key contributions (RCC finetuning, synchronization) could be directly used, as they are orthogonal to the underlying scheme choice.

\paragraph{Detection.} For the KGW method that we focus on, the \emph{watermark detection} process analyzes a token sequence $s$ and computes a score $S$ based on the count of green tokens:
\begin{equation}
    S = \sum_{t=h+1}^T \mathds{1}(s_t \in \mathcal{G}_t),
\end{equation}
where $\mathcal{G}_t$ is the greenlist for position $t$, which depends on the $h$ preceding tokens and the secret key $\xi$.

\paragraph{Statistical hypothesis testing.}
Detection uses a statistical hypothesis test distinguishing between $\mathcal{H}_0$: ``the sequence is not watermarked with secret key $\xi$'' and the alternative $\mathcal{H}_1$: ``the sequence was generated with a watermark with secret key $\xi$.''
Previous approaches, such as those by \citet{kirchenbauer2023watermark} and \citet{aaronson2023watermarking}, relied on a $Z$-test to compare the count of green tokens $S$ to its expected value under the null hypothesis $\mathcal{H}_0$.
In this work we instead adopt an exact test~\citep{fernandez2023three}, which is more accurate, especially for short sequences. 

Under $\mathcal{H}_0$, $S$ follows a binomial distribution $\mathcal B$ with parameters $(T-h)$ and $\gamma$, where $\gamma$ is the expected proportion of green tokens, $T$ is the total number of tokens, and $h$ is the size of the watermark context window.
The p-value determines the likelihood of observing a score as extreme as $S$ under $\mathcal{H}_0$, and is calculated as:
\begin{equation}
\text{p-value}(S,T,h,\gamma) = \text{Prob} \left( X \geq S \mid X \sim \mathcal{B}(T-h, \gamma) \right) = I_{\gamma}(S, T-h-S+1),
\end{equation}
where $I_x(a,b)$ is the regularized incomplete Beta function.

Sequences are flagged as watermarked if the p-value falls below the desired false positive rate.

\paragraph{Main parameters.}
The main parameters of the watermarking method are the context window size $h$, the watermark strength factor $\delta$ and the proportion of green tokens $\gamma$.

The context window size $h$ determines how many previous tokens determine the greenlist. 
A smaller $h$ increases robustness against text modifications but may bias generation as the same hash is used more frequently.
It typically reduces security since recurring greenlists make the watermark easier to spoof~\citep{jovanovic2024watermark,milpstealing}.
When $h=0$, the RNG seed depends solely on the secret key $\xi$, creating fixed green/red lists for all tokens.
For non-text tokens, we hypothesize that $h=0$ maintains security since tokenizer access is restricted and image tokenizers have more degrees of freedom than text ones.

The watermark strength factor $\delta$ determines the amount by which the logits of green tokens are boosted.
A higher $\delta$ increases the robustness of the watermark, but also increases the risk of generating low-quality text/images.
It is tuned for every model and application.

The proportion of green tokens $\gamma$ affects both detection sensitivity and generation quality.
With low $\delta$, a smaller $\gamma$ reduces green token selection during generation, resulting in lower watermark power.
With high $\delta$, it restricts token choice and may lower output quality.
During detection, lower $\gamma$ values yield more significant p-values since green tokens are less likely to appear by chance~\citep{kirchenbauer2023watermark}.
At fixed watermark power, higher $\gamma$ distributes the watermark evenly, while lower values concentrate it on fewer tokens.
We set $\gamma$ to $0.25$ in our experiments, as it is a common choice in the literature~\citep{aaronson2023watermarking,kirchenbauer2023watermark} and consistently yields good results in our experiments.

%% file: sections/app-consistency.tex
\section{More on Reverse Cycle-consistency} \label{app:consistency}

In this section, we elaborate on the case of text tokenizers not being perfectly reverse cycle-consistent (RCC), discuss audio tokenizers and our experiment measuring RCC in this setting, expanding on~\cref{sec:audio}, and discuss related topics studied in prior work.

\subsection{RCC in Text Tokenizers} \label{app:consistency:text}
In BPE tokenizers, the vocabulary is initialized with all characters in the training set, and common character pairs are iteratively merged and added to the vocabulary until the predefined size is reached. 
Tokenization is performed greedily from left to right, by always selecting the longest possible token from $V$.
Detokenization is simply performed by a lookup into $V$.

\paragraph{RCC can be violated.}
Text tokenizers are not immune to the RCC issue. 
For example, consider the following subset of the \textsc{gpt-4o} tokenizer: $\{\text{\texttt{cons}}\colon 9673, \text{\texttt{istent}}\colon 20908, \text{\texttt{consistent}}\colon 173878\}$. 
Due to the greedy property of BPE tokenizers, $\detok{\text{{text}}}$ is guaranteed to always invert $\tok{\text{{text}}}$, \eg $$\detok{\text{{text}}}(\tok{\text{{text}}}(\text{\texttt{consistent}})) = \detok{\text{{text}}}([173878]) = \text{\texttt{consistent}},$$ guaranteeing \emph{forward cycle-consistency (FCC)}. 
In contrast, \emph{reverse cycle-consistency (RCC)}, necessary for a strong watermark, may be violated, \eg $$\tok{\text{{text}}}(\detok{\text{{text}}}([9673, 20908])) = \tok{\text{{text}}}(\text{\texttt{consistent}}) = [173878].$$
RCC approximately holds for text tokenizers in practice: while it is also a prerequisite for successful watermarking in text, no prior art has highlighted this as a hurdle.
Some works have even shown that adversaries learning about the watermark can still be successful even if they use a different tokenizer~\citep{jovanovic2024watermark}, which is only possible if the tokens match across tokenizers.

\paragraph{Experiment.}
As discussed in the main text, we confirmed this experimentally. 
We used \textsc{Llama3.1-8B-Instruct} to generate $1000$ answers to prompts from the Open~Assistant dataset~\citep{kopf2023openassistant}. 
We then compared the generated sequences of tokens with the re-tokenized sequences. 
Specifically, we took the token IDs from the model's generation, detokenized them to text, then re-tokenized this text and computed the Levenshtein distance between the original and the new token sequence. 
Our results showed that the average token match is 99.5\%, confirming that text tokenizers exhibit very high reverse cycle-consistency in practice.

\subsection{RCC in Audio Tokenizers} \label{app:consistency:audio}
A study of the RCC issue in \moshi's tokenizer is already given by the authors~\citep{defossez2024moshi} (called \emph{idempotence}).
We however observe some differences in our study, as well as other key findings, such as the effect of augmentations, that we summarized in \cref{sec:audio}, and that we discuss in more detail in the following.

\paragraph{RVQ tokenizer.}
As a reminder, \moshi's tokenizer (\textsc{Mimi}) utilizes \emph{residual vector quantization} (RVQ)~\citep{rvq,kumar2023high,soundstream,defossez2022high}.
In RVQ, the quantization process happens iteratively, where each step quantizes the residual error from the previous quantization. 
Formally, for each step $i$, representing an audio frame of 1920 samples, the tokenization results in a sequence of tokens $(s_{i}^{1},.., s_{i}^{K})$ corresponding to $K$ different codebooks $C^{1}, .., C^{K}$ (referred to as $K$ \emph{streams}). 
Each token represents approximately 80\,ms of audio.

The first token (or stream) is referred to as \emph{semantic}, because there is a distillation loss during training with a non-causal model that encourages this first codebook to capture the most semantically relevant information. 
\citet{defossez2024moshi} note that this semantic token exhibits higher cycle-consistency compared to later streams, which are assumed to progressively capture more fine-grained details, and to be less consistent.

\begin{table}[t]
\centering
\caption{Token Match across different streams for $1000$ sequences, where audios are subject to transformations before re-tokenization.
Sequences of tokens are generated either as reconstructions of $10$ second VoxPopuli audios with the \mimi tokenizer, or by the \moshi model with audio prompts (described in \aref{app:expdetails:audio}).
}
\label{tab:audio-rcc}
\resizebox{1.0\textwidth}{!}{
    \input{tables/appendix/audio_rcc}

}
\end{table}

\paragraph{Experimental setup.}
We measure Token Match (TM) for sequences either (a) generated as reconstructions of 10-seconds audios from VoxPopuli with the \mimi tokenizer, or (b) generated by the \moshi model, as described in \aref{app:expdetails:audio}.
This corresponds to $\approx$125 time-steps for both cases, so $\approx$1,000 audio tokens (counting all the streams).
The audio is subjected to various transformations before re-tokenization, which include the three categories: valuemetric (lowpass filtering at 3kHz, addition of strong Gaussian noise at 0.01 amplitude), temporal-frequency (speed modification by 1.25x, cropping 90\% of the original audio), and compression-based (MP3 compression at 16kbps, EnCodec compression).

\paragraph{Results.}
\Cref{tab:audio-rcc} presents the results that supplement the study by \citet{defossez2024moshi}.
Notably, their study only focuses on pre-existing audio sequences, while we also include generated sequences, which behave differently.
For instance, the second stream sometimes shows higher consistency than other streams contradicting the expectation that only the first (semantic) stream could maintain high consistency. 
Different augmentations affect streams differently: \eg lowpass has less impact on streams 2 and 3 compared to strong noise addition, while EnCodec strongly decreases TM of the first stream. 

Importantly, temporal-frequency augmentations (speed, cropping) reduce TM less dramatically for \moshi (0.04-0.15) than would be expected given our image watermarking results where such transformations typically yield near-zero consistency.
This multi-stream aspect presents challenges and opportunities for watermarking: while complicating RCC analysis, it enables potential development of more sophisticated techniques leveraging complementary properties across streams.

\subsection{Related Concepts} \label{app:consistency:related}
There are several concepts related to RCC that were studied in prior work. 

\paragraph{Codec idempotence.}
In the context of codecs a relevant property is \emph{codec idempotence}~\citep{o2025code,kim2020instability,imageidempotence}.
Directly applying the mathematical definition of idempotence to our case, we let $f(\cdot)$ denote the encoder-decoder pair (\eg $f$ compresses an image to a JPEG file and then decompresses it back to pixels), and say that a codec is idempotent if it satisfies:
\begin{equation}
    f(f(x)) = f(x).
\end{equation}
This property is naturally of interest to codecs.
While we assume that JPEG compressing an image is lossy (\ie we do not expect $f(x)=x$), we want our codec to not further degrade image quality on successive applications, which can commonly occur in practice (\ie $f(f(x))=f(x)$).
This is in stark contrast with the notion of RCC relevant to our work.
In our case, $x$ are the tokens, while $f(\cdot)$ is the detokenization followed by the tokenization. 
The first application of $f(\cdot)$ is crucial for us: as explained above, we require $f(x) \approx x$ as otherwise the watermark that was present in the tokens of $x$ is lost. 
$f(f(x))$, \ie re-tokenizing the image several times successively, on the other hand is not of particular interest in this case, thus idempotence is not an important concern.

\paragraph{Consistency of tokenizers.}
Another related concern is the \emph{consistency of tokenizers}~\citep{liu2024analyzing,tokenizationconsistency}.
Intuitively a tokenizer is \emph{consistent} if the tokenization of a particular string (assuming the text domain) does not change depending on the surrounding context. 
While the cited works show that this is a desirable property, it is not as relevant to our motivation of preserving the watermark as RCC.
In particular, for generated token sequence $x$, if RCC is satisfied the watermark will be entirely preserved, even if the tokenization was context-dependent. 
This may be a concern in the context of various attacks: for example, infilling a part of the image before re-tokenization should ideally change only the tokens corresponding to the infilled part, and not the entire image, which may happen depending on the setup of the convolutions in the tokenizer.
We do not explore this angle as part of this work. 

\paragraph{Cycle-consistency in other contexts.}
Finally, a line of works studies cycle-consistency in various generative models~\citep{zhu2017unpaired,cc}, most commonly in the context of style transfer: a single \emph{cycle} is the translation from a style $A$ to a style $B$ and back to $A$, and cycle-consistency can in this case be beneficial as a constraint for the model. 
Finally, \citet{invertibleae} explicitly parametrize encoder-decoder pairs to be inverses of each other and \citet{huang2020cycle} study cycle-consistency in the context of disentangled representations. 

%% file: tables/appendix/audio_rcc.tex
\begin{tabular}{
    l 
    *{9}{>{\centering\arraybackslash}p{0.5cm}} 
    @{\hskip 15pt} 
    *{9}{>{\centering\arraybackslash}p{0.5cm}}
}
\toprule
& \multicolumn{9}{c@{\hskip 15pt}}{\mimi tokenizer} & \multicolumn{9}{c}{\moshi} \\
\cmidrule(lr{1em}){2-10} \cmidrule(rr){11-19}
& 1 & 2 & 3 & 4 & 5 & 6 & 7 & 8 & Avg. & 1 & 2 & 3 & 4 & 5 & 6 & 7 & 8 & Avg. \\
\midrule
Identity                
& 0.56 & 0.31 & 0.21 & 0.22 & 0.18 & 0.20 & 0.18 & 0.16 & 0.25 
& 0.60 & 0.55 & 0.38 & 0.28 & 0.26 & 0.26 & 0.24 & 0.25 & 0.35 \\
\midrule
{\hspace{-4pt} \footnotesize \textit{Transformations}} \\
Lowpass 3~kHz            
& 0.38 & 0.15 & 0.15 & 0.17 & 0.14 & 0.16 & 0.15 & 0.12 & 0.18 
& 0.50 & 0.39 & 0.30 & 0.21 & 0.19 & 0.21 & 0.20 & 0.21 & 0.28 \\
Noise 0.001 
& 0.50 & 0.33 & 0.19 & 0.20 & 0.17 & 0.18 & 0.17 & 0.14 & 0.24 
& 0.51 & 0.34 & 0.19 & 0.18 & 0.17 & 0.18 & 0.18 & 0.13 & 0.23 \\
MP3 16~kbps              
& 0.44 & 0.19 & 0.16 & 0.18 & 0.15 & 0.17 & 0.16 & 0.13 & 0.20 
& 0.54 & 0.41 & 0.29 & 0.20 & 0.18 & 0.20 & 0.20 & 0.21 & 0.28 \\
Encodec                 
& 0.24 & 0.13 & 0.10 & 0.10 & 0.07 & 0.07 & 0.05 & 0.04 & 0.10 
& 0.28 & 0.38 & 0.24 & 0.16 & 0.14 & 0.15 & 0.11 & 0.15 & 0.20 \\
Speed $\times$1.25      
& 0.02 & 0.01 & 0.01 & 0.00 & 0.00 & 0.00 & 0.00 & 0.00 & 0.01 
& 0.04 & 0.11 & 0.08 & 0.05 & 0.05 & 0.06 & 0.04 & 0.07 & 0.06 \\
Crop (90\% kept)        
& 0.03 & 0.01 & 0.01 & 0.01 & 0.01 & 0.01 & 0.01 & 0.00 & 0.01 
& 0.08 & 0.15 & 0.11 & 0.07 & 0.07 & 0.08 & 0.05 & 0.09 & 0.09 \\
\bottomrule
\end{tabular}

%% file: sections/app-syncdetails.tex
\section{More on Watermark Synchronization} \label{app:syncdetails}
In~\aref{app:syncdetails:image} we provide a more detailed description of our watermark synchronization layer (\cref{ssec:method:sync}), and show additional examples. 
In~\aref{app:syncdetails:audio} we describe our attempt to use AudioSeal~\citep{sanroman2024proactive} for synchronization in audio.

\subsection{Image Synchronization Details} \label{app:syncdetails:image}
We remark that the problem of watermark synchronization was studied in the past, before the advent of generative models, in the context of digital watermarking.
These works suggest approaches such as multiple testing with a carefully controlled number of tests to avoid the false positive rate increase we mentioned in \cref{ssec:method:sync}~\citep{hartung2000spread}, or similarly to us, embedding a synchronization pattern in addition to the original watermarking pattern to revert the transformation~\citep{tirkel98,csurka99,pereira99}.
As noted above, we are aware of two works that study this in the context of post-hoc generative model watermarks~\citep{guo2023practical,luo2022leca}, however their code is not publicly available.

\paragraph{Full algorithm description.}
Our algorithm consists of two main procedures: embedding a synchronization pattern into the generated and decoded image, and estimating the transformation from an incoming image where previously the watermark and the synchronization pattern were embedded.
We assume access to a localized watermark module $L$ that can embed a different message in every pixel of an image, and recover the probability that each pixel has the watermark along with the most probable message in it. 
As noted above, we instantiate this using~\citet{sander2024watermark}.

To embed the synchronization pattern, we use four 32-bit messages \{\textcolor{magenta!90}{$m_1=0^{32}$}, \textcolor{red!95}{$m_2=0^{16}1^{16}$}, \textcolor{teal!90}{$m_3=1^{16}0^{16}$}, \textcolor{blue!95}{$m_4=1^{32}$}\}. 
For each message, the mask is one of the quadrants (as in~\cref{fig:sync}), where given parameter $\mu$, we keep a horizontal and a vertical strip of width $\mu$ pixels in the middle of the image free of messages (we use $\mu=18$ for \taming and \rarxl, and $\mu=36$ for \cham as we work with images of twice the resolution). 
Using this mask, we embed the pattern using $L$.

Given an incoming image, we first obtain and postprocess the predictions of $L$.
Namely, for each pixel, we take the closest message in Hamming distance from the four fixed messages above, as long as the Hamming distance is below $6$ bits and the probability of the pixel being watermarked as predicted by $L$ is above $0.5$.
Then, as a heuristic, we proceed only if we found at least one pixel for each of the $4$ messages, and if the total area of the pixels with messages is at least $70\%$ of the image---attempting to estimate the transformation otherwise proved too unreliable.

To estimate the transformation, we sweep over rotations in $[-20, 20]$ degrees, for each rotation \emph{rotate the grid of extracted messages back} by the inverse value, and find the best-fitting pair $(i,j)$ such that the row $i$ of the rotated message grid best separates pixels with messages $(m_1, m_3)$ as well as $(m_2, m_4)$, and the column $j$ of the rotated message grid best separates pixels with messages $(m_1, m_2)$ as well as $(m_3, m_4)$.
For example, to find $j$ that best separates $(m_1, m_2)$, we compute the \emph{cost} of each candidate $j'$ as the number of \emph{wrongly positioned} pixels, \ie pixels with message $m_1$ that are to the right of $j'$ and pixels with message $m_2$ that are to the left of $j'$.
We repeat the same cost computation for the horizontally flipped message grid: if we find that this leads to a lower cost, we estimate that the image was flipped. 
The lowest-cost estimate for $j$ and the estimate if the image was flipped or not are then aggregated over $(m_1, m_2)$ and $(m_3, m_4)$ proportionally to the number of pixels with these messages in the image. 
The process for $i$ is analogous, with the final result for the fixed suspect rotation being the tuple $(i, j, isFlipped, cost)$.
We finally take such tuple with the minimal $cost$, returning the corresponding rotation and $(i,j,isFlipped)$ as our final estimate. 

\paragraph{Examples.}
Complementing~\cref{fig:sync}, in \cref{fig:sync_examples} we show real examples of recovered synchronization patterns and estimated transformations for horizontal flip, positive and negative rotation, crop, as well as Gaussian noise, that we found to be the most challenging valuemetric transformation in terms of disturbing the synchronization signal.

\subsection{Audio Synchronization with Localized Audio Watermarking} \label{app:syncdetails:audio}

\begin{figure}[t]
    \centering
    \begin{subfigure}[b]{0.32\textwidth}
        \centering
        \includegraphics[width=0.95\textwidth]{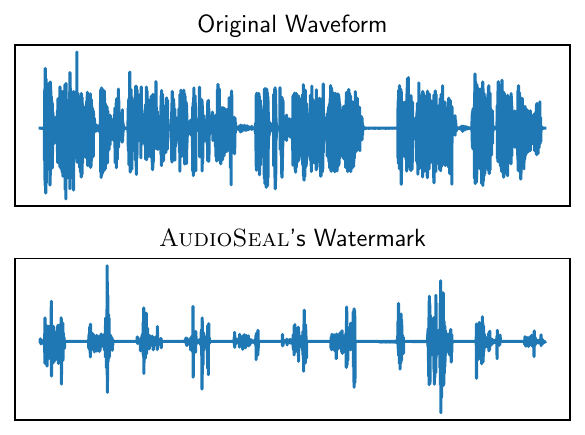}
        \caption{Visualization of the synchronization mask pattern}
    \end{subfigure}
    \hfill
    \begin{subfigure}[b]{0.32\textwidth}
        \centering
        \includegraphics[width=0.95\textwidth]{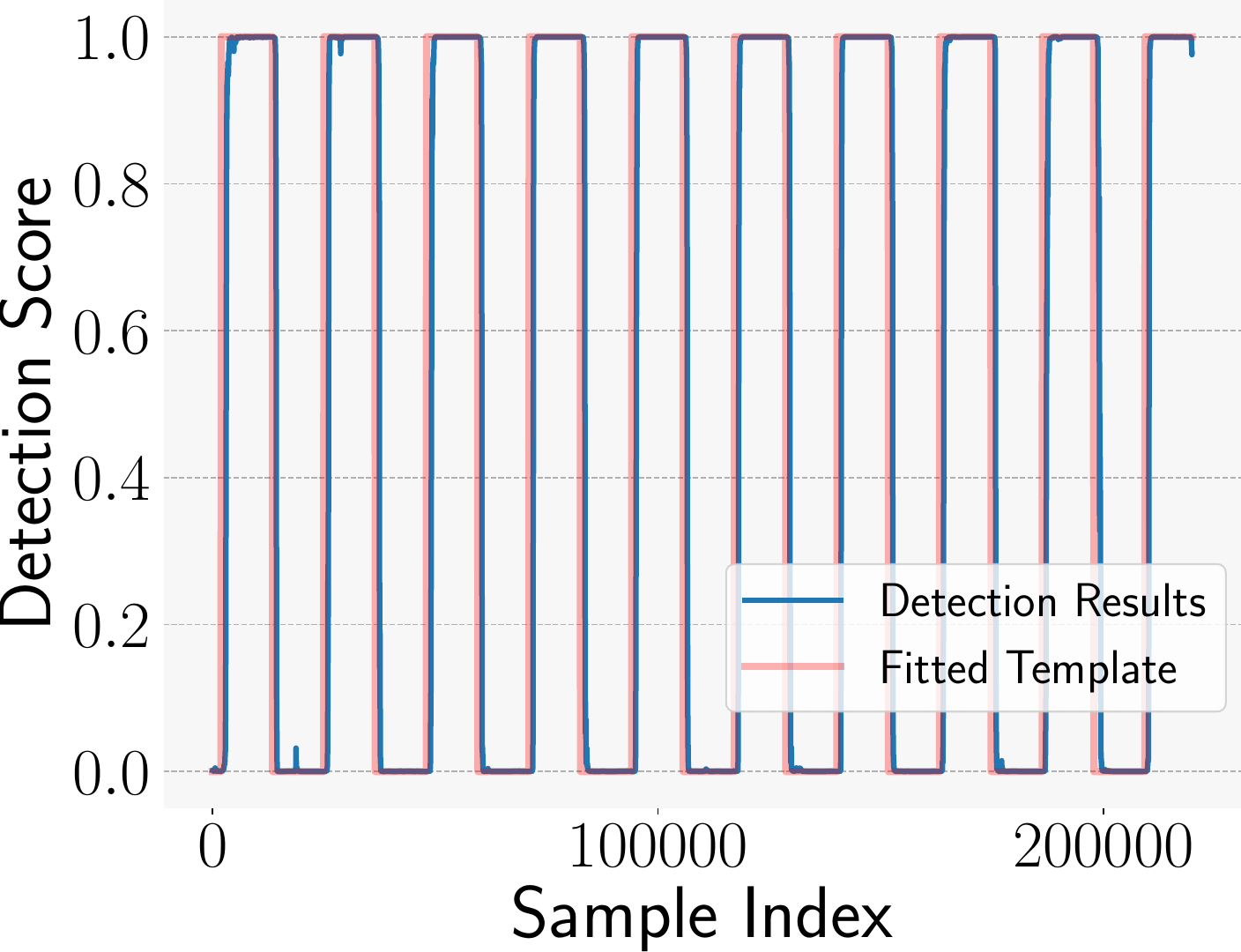}
        \caption{Example of successful detection after a small temporal crop}
    \end{subfigure}
    \hfill
    \begin{subfigure}[b]{0.32\textwidth}
        \centering
        \includegraphics[width=0.95\textwidth]{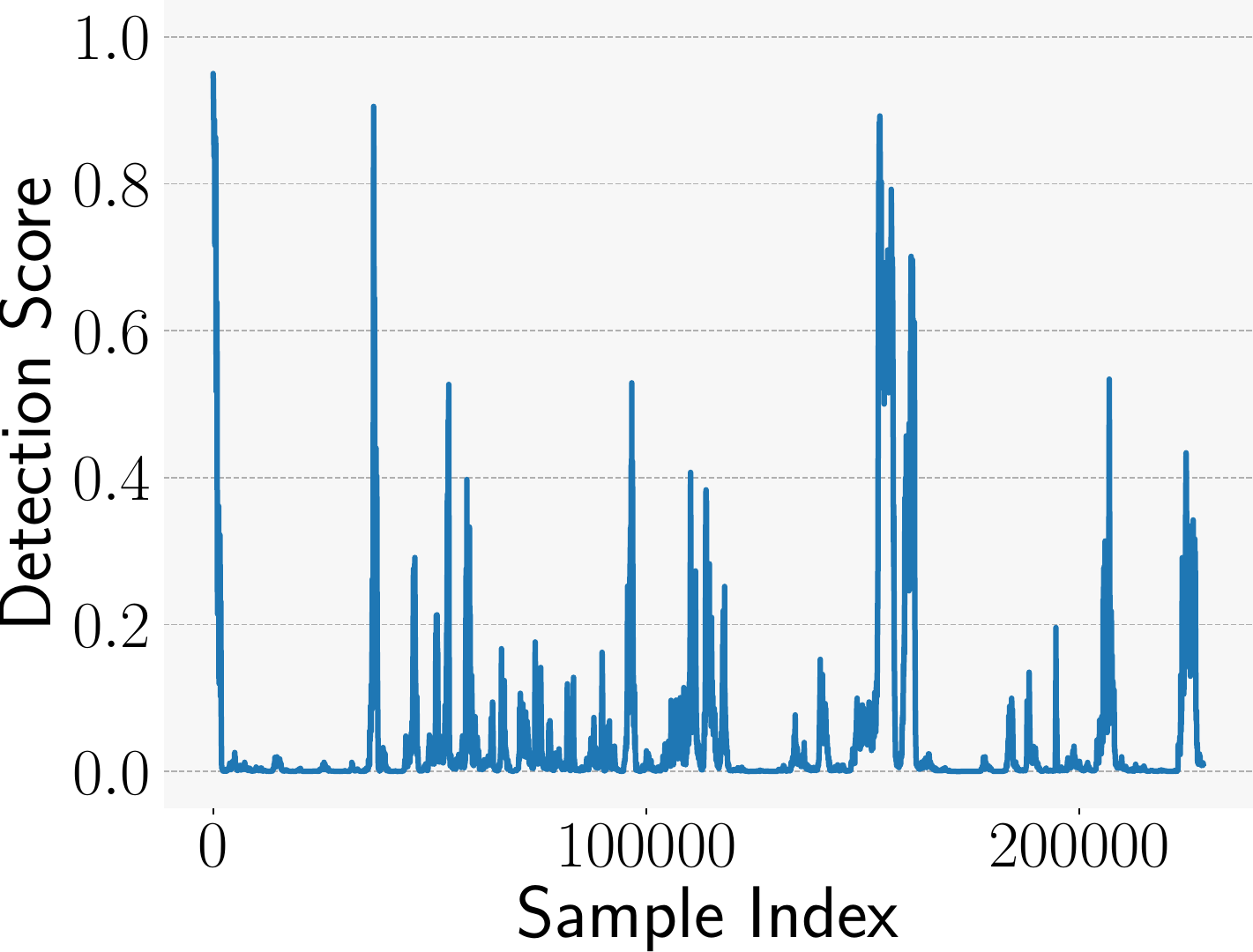}
        \caption{Example of unsuccessful detection after 1.05$\times$ speedup}
    \label{fig:audio_sync_c}
    \end{subfigure}
    \caption{The audio watermark synchronization method we attempted to incorporate. (a) how we embed the watermark periodically in the audio; (b) successful case with a clear periodic pattern detected when the audio is cropped for the first $0.84$ seconds; (c) a $1.05\times$ speedup creates a detection signal too noisy to reliably extract the synchronization pattern.}
    \label{fig:audio_sync}
    \vspace{-1em}
\end{figure}

As explained in \cref{sec:audio}, we attempted to use the localization property of AudioSeal~\citep{sanroman2024proactive} for synchronization similar to \cref{ssec:method:sync}, but this approach proved less successful than with images.
We describe below the method and experiments supporting this claim and summarize the results in \cref{fig:audio_sync}.

\paragraph{Synchronization approach and transformation detection.}
We use AudioSeal's embedder to embed watermarks with a periodic mask pattern across the signal. 
This enables detection of transformations like time-stretching and phase shifts through cross-correlation analysis. 
We applied a square wave template with periods of 6 frames, each frame of 1920 samples at 24kHz, as in \mimi.

When audio undergoes transformations, the periodic pattern distorts predictably. 
Through cross-correlation between the detection signal and template patterns, we can estimate the speedup factor (identified by the period maximizing cross-correlation) and phase shift (located by finding optimal alignment). 
Once estimated, we can invert the transformation by resampling to original speed and applying phase correction. More specifically, in our implementation, the detection results are first downsampled by a factor of $8$ and we sweep the template period from $0.5\times$ to $1.5\times$ the nominal half-period (6 frames = $0.48$s) in coarse steps of $10$ samples, then refine over $\pm10$ samples around the best match; this two-stage cross-correlation yields precise estimates of speedup (from the best period) and phase shift, which are used to resample back to $24$kHz and correct the alignment.

\input{figures/appendix/sync_examples.tex}
\paragraph{Challenges and limitations.}
As shown in \cref{fig:audio_sync_c}, the detection signal extracted from AudioSeal is not yet robust enough for general use, as it fails to detect the watermark reliably after a speedup of $1.05\times$. 
Future work could explore more sophisticated synchronization methods.

%% file: figures/appendix/sync_examples.tex
\clearpage
\begin{figure}[t]
  \centering
  \resizebox{0.65 \linewidth}{!}{%
  \small
  \newcommand{\colspace}{\hspace{0.035\linewidth}}
    \begin{tabular}{c}
      \toprule
      \includegraphics[width=\linewidth]{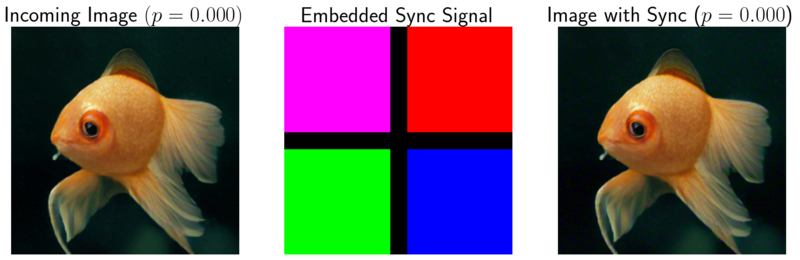} \\
      \includegraphics[width=\linewidth]{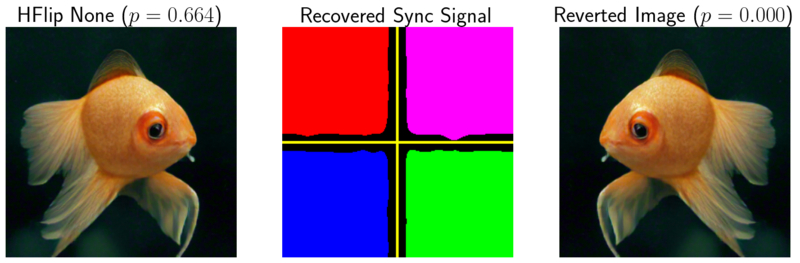} \\
      \includegraphics[width=\linewidth]{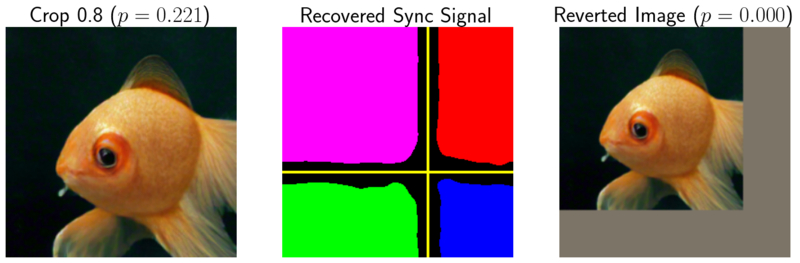} \\
      \includegraphics[width=\linewidth]{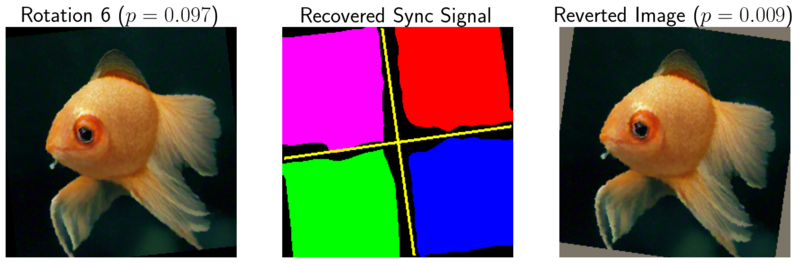} \\
      \includegraphics[width=\linewidth]{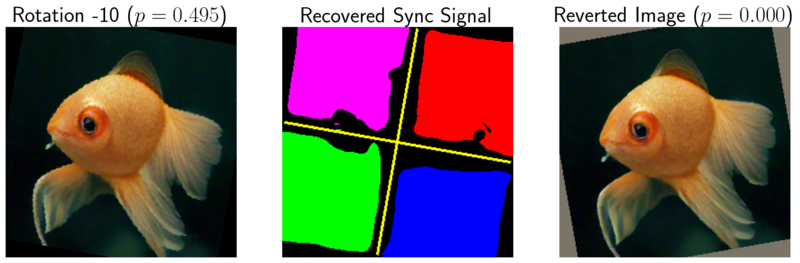} \\
      \includegraphics[width=\linewidth]{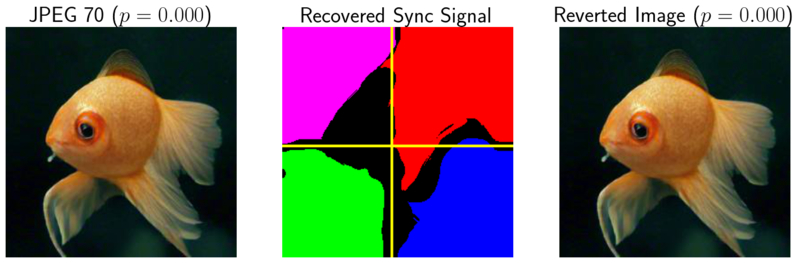} \\ 
      \bottomrule
    \end{tabular}}
  \caption{Visualization of our synchronization layer (\cref{ssec:method:sync}) on a real example from our experiments. In the four middle rows we see that the watermark detection would have failed on original geometrically transformed images, but has eventually succeeded after the synchronization signal was detected and reverted. In the bottom row we see that a valuemetric transformation can disrupt the signal---in this case a JPEG compression. While this did not hamper detection in this example, it can be problematic in practice as evidenced by the drop in valuemetric robustness with synchronization shown in~\cref{tab:baselines}.}
  \label{fig:sync_examples}
\end{figure} 
\clearpage

%% file: sections/app-expdetails.tex
\section{Experimental Details} \label{app:expdetails}

\subsection{Omitted Details of Image RCC Evaluation and Finetuning} \label{app:expdetails:rccft}
Here we provide more details on our RCC evaluation experiments shown above in \cref{tab:tokens} and details related to RCC finetuning introduced in~\cref{ssec:method:ft}.

\paragraph{RCC evaluation.}
For~\cref{tab:tokens}, for simplicity, we re-use the watermarked \base model of \taming from our main experiments.
We confirmed that running the non-watermarked version results in very similar values. 
We use the full set of valuemetric and geometric transformations as in the main experiments, and for each transformation use the same parameter that was chosen for summarized scores in~\cref{tab:baselines}, as detailed below in \aref{app:expdetails:main}.

\paragraph{RCC finetuning.}
To complete our RCC finetuning description from~\cref{ssec:method:ft} we provide the omitted details. 
We note that the exact same hyperparameters were used for all models which demonstrates the transferability of our finetuning procedure. 
The finetuning is done for $10$ epochs with distributed data parallel training on $16$ V100 GPUs (\taming, training takes $2$h) and $8$ H200 GPUs (\cham and \rarxl, training takes $2.5$h and $0.5$h respectively).
We use the Adam optimizer~\citep{kingma2015adam} with a learning rate of $10^{-4}$, multiplied by a factor of $0.9$ each epoch (\texttt{StepLR}).
We use a total batch size across all GPUs of $64$ ($4$ per gpu for \taming and $8$ per gpu for \cham and \rarxl), and always set $\lambda=1$.
As noted above, we use a set of augmentations $\mathcal{A}$ to improve robustness of our watermark to transformations and attacks. 
JPEG is not differentiable, therefore we backpropagate only through the difference between the uncompressed and compressed images (straight-through estimator):
$x'= x_{\mathrm{aug}} + \mathrm{nograd}(x_{\mathrm{aug}, \mathrm{JPEG}} - x_{\mathrm{aug}})$~\citep{zhang2021asl,yin2019understanding}.
We define three progressively harder sets: $\mathcal{A}_1$, $\mathcal{A}_2$, $\mathcal{A}_3$, and use no augmentations for $1$ epoch, then $\mathcal{A}_1$ for $1$ epochs, then $\mathcal{A}_2$ for $4$ epochs, and finally $\mathcal{A}_3$ for the last $4$ epochs. 

$\mathcal{A}_1$ uses JPEG compression with qualities $\{90, 80, 70\}$, Gaussian blur with kernel sizes $\{1, 3\}$, Gaussian noise with standard deviations $\{0.005, 0.01, 0.015, 0.02\}$, Brigthening with factors $\{1.0, 1.1, 1.2\}$, Rotation with angles $\{-1, 1\}$ degrees, and Cropping with \% kept from $\{80, 90\}$.
$\mathcal{A}_2$ uses JPEG with qualities $\{80, 60, 40\}$, Gaussian blur with kernel sizes $\{3, 5\}$, Gaussian noise with standard deviations $\{0.02, 0.04, 0.06\}$, Brigthening with factors $\{1.2, 1.3, 1.4\}$, Rotation with angles $\{-3,-2,-1,1,2,3\}$ degrees, and Cropping with \% kept from $\{50,60,70,80,90\}$.
Finally, $\mathcal{A}_3$ uses JPEG with qualities $\{40, 30, 20\}$, Gaussian blur with kernel sizes $\{5, 7, 9\}$, Gaussian noise with standard deviations $\{0.06, 0.08, 0.1\}$, Brighten with factors $\{1.4, 1.7, 2.0\}$, and the same geometric augmentations as in $\mathcal{A}_2$.

\subsection{Details of Main Experiments} \label{app:expdetails:main}
We provide full details of our main experiments (\cref{ssec:experiments:main} and \cref{ssec:experiments:robustness}).

\paragraph{Models.}
For \taming, we use the \textsc{VQGAN ImageNet (f=16), 16384} version available in the authors' repository.
For \cham, we use the 7B model. 
Since the open-weight version does not include image generation capabilities (as noted in the original paper), we obtained the necessary weights directly from the authors.
Alternatively, image generation with \cham can be approximated using the Anole model~\citep{chern2024anole} and its associated repository: \url{https://github.com/GAIR-NLP/anole}, though we note that its output quality is somewhat lower.
For \rarxl we use the official model and image tokenizer from the authors' repository.

\paragraph{Parameters.} \looseness=-1
The results in~\cref{fig:power}, \cref{fig:robustness} and \cref{tab:baselines} are obtained from the same experiment, repeated on \taming, \cham, and \rarxl.
For \taming and \rarxl we set $\delta=2, \gamma=0.25, h=1$ and evaluate (for each transformation/attack) on $1000$ generations, $100$ per each of the following ImageNet class indices: $[1,9,232,340,568,656,703,814,937,975]$.
For \cham we set $\delta=2, \gamma=0.25, h=0$. We again use $1000$ generations, conditioning the model on a text prompt each time. 
Following the standard protocol in the literature~\citep{ramesh2022hierarchical, ramesh2021zero, rombach2022high, saharia2022photorealistic} we use the prompts from the validation set of MS-COCO~\citep{lin2014microsoft}.
To do so, we first retrieve all the captions from the validation set, keep only the first one for each image, and select the first $1000$ (or $5000$ when computing FID for \cham).
While we did not benchmark this in detail, the computational overhead of our watermarked generation matches that of the LLM watermarking scheme we inherit from \emph{and} the localized watermark we use as the synchronization signal.

\input{figures/appendix/stratification.tex}
\vspace{-0.3em}
As discussed in~\cref{sec:conclusion} more thoroughly integrating these two components could also make our watermark more efficient. 
A single run (\eg \base with all (augmentation, parameter) pairs detailed below on $1000$ generations) with \taming was executed on $25$ V100 GPUs, lasting ${\approx}30$ minutes for \base, \ft, \ftaug and ${\approx}1.5$h for \ftaugsync.
For \cham and \rarxl, we use $10$ H200 GPUs ($50$ for \ftaugsync), taking comparable time as for \taming.
 
\paragraph{Split stratification.}
As noted in the literature~\citep{esser2021taming,yu2021vector}, trained VQGANs often suffer from \emph{low codebook utilization}, meaning that a certain percentage of the codebook is effectively not used and those tokens (\emph{dead codes}) are in practice never emitted by the transformer nor used when tokenizing images.
While later work addresses this issue~\citep{zhu2024scaling}, the VQGAN used in \taming and in our experiments suffers from this issue and has only $971$ \emph{alive codes}, despite the codebook size of $16384$.

This can affect the soundness of the watermark, \ie of the statistical test used to detect the watermark (\cref{eq:pvalue}).
Namely, the test's assumed null hypothesis is that the ratio of green tokens in texts produced without the model $\model$ is distributed according to a Binomial distribution with mean $\gamma$. 
However, the true parameter $\gamma'$ can be different if the number of alive codes $\nalive$ is much smaller than the codebook size $|V|$.
In other words, when we choose the green tokens as a uniformly random subset of $|V|$ of size $\gamma |V|$, we may have significantly more or less than $\gamma \nalive$ green tokens among the alive codes.
As these are the only tokens ever emitted by the tokenizer in practice, \cref{eq:pvalue} makes a mistake by using $\gamma$ instead of $\gamma'$ in its calculation. 
While for $h>0$ we can hope that this mistake averages out across different contexts (as the expected alive green ratio is still $\gamma$), for $h=0$ (fixed red/green split) this can introduce a constant bias. 

In particular, assume for simplicity that $\gamma |V|$ and $\gamma \nalive$ are both integers.
What can we say about the distribution of $\gamma'$ with respect to the random green/red split?
The probability that a uniformly random split of $|V|$ into $\gamma |V|$ green and $|V|-\gamma |V|$ red tokens results in \emph{exactly} $g$ green tokens among alive ones (resulting in $\gamma' = g / \nalive$) is given by a \href{https://en.wikipedia.org/wiki/Hypergeometric_distribution}{hypergeometric distribution} with PMF $f$:
\begin{equation}
    \mathbb{P}(\gamma' = g / \nalive) = f(g; |V|, \gamma |V|, \nalive) = \frac{\binom{\gamma |V|}{g} \cdot \binom{|V| - \gamma |V|}{\nalive-g}}{\binom{|V|}{\nalive}}.
\end{equation}
In~\cref{fig:stratification} we plot the distribution of {actual green ratios} $\gamma'$, \ie plot $\mathbb{P}(\gamma' = g / \nalive)$ as a function of $g / \nalive$, for $|V|=192, \nalive=12, \gamma=0.25$ (corresponding roughly to the ratio $\nalive / |V|$ of the VQGAN we use in our experiments).
We see (\emph{in red}) that there is in fact only ${\approx}25\%$ chance that the green ratio among alive tokens $\gamma'$ is equal to $\gamma$, as the statistical test expects. 
To resolve this, for \taming we use a \emph{stratified} split, \ie we separately sample a red/green split on alive and dead codes, ensuring that $\gamma'=\gamma$ (\emph{green in~\cref{fig:stratification}}).

An interesting question, to the best of our knowledge not explored before, is if similar effects can be observed in LLM watermarking.
For example, a uniformly random split of a large multilingual vocabulary may introduce a particularly biased split on \eg Cyrillic tokens, which are effectively the majority of the alive ones when the LLM is prompted to write in a language that uses the Cyrillic script.
Especially for $h=0$, this may point at unfairness towards certain subdomains, where for a particular subdomain the watermark is overly conservative or more importantly has a much higher FPR than stated theoretically. 

\paragraph{Image transformations.}
We next list all image transformations and their parameters used in our main experiments.
We evaluate $90$ variants (the original image and $89$ transformations described below) for each image, \ie $90{,}000$ images in total per evaluation. 
For valuemetric transformations we use:
\begin{itemize}[itemsep=1pt, topsep=1pt, leftmargin=16pt]
    \item Gaussian Blur: kernel sizes $[\underline{0}, 1, 3, 5, 7, {\bf 9}, 11, 13, 15, 17, 19]$. 
    \item Gaussian Noise: standard deviations $[\underline{0}, 0.025, 0.05, 0.075, {\bf 0.1}, 0.125, 0.15, 0.175, 0.2]$.
    \item JPEG Compression: quality factors $[\underline{100}, 95, 85, 75, 65, 55, 45, 35, {\bf 25}, 15, 5]$.
    \item Brighten: factors $[\underline{1}, 1.25, 1.5, 1.75, {\bf 2}, 2.25, 2.5, 2.75, 3]$. 
\end{itemize}
For geometric transformations we use: 
\begin{itemize}[itemsep=1pt, topsep=1pt, leftmargin=16pt]
    \item Rotation: angles $[-20, -15, -10, -5, \underline{0}, 5, {\bf 10}, 15, 20]$. 
    \item Horizontal Flip: parameters $[\underline{0}, {\bf 1}]$, where $1$ indicates that a flip was performed. 
    \item Crop: percent of each side kept $[\underline{1.0}, 0.95, 0.9, 0.85, 0.8, {\bf 0.75}, 0.7, 0.65, 0.6, 0.55, 0.5]$. For example, for $0.75$ we crop from the top-left corner of the image, keeping $75\%$ of the width and height, and then resize back to the original size.
\end{itemize}
Finally, we use the following attacks:  
\begin{itemize}[itemsep=1pt, topsep=1pt, leftmargin=16pt]
    \item DiffPure: timesteps $[0.01, 0.05, {\bf 0.1}, 0.2, 0.3]$ with the $256 \times 256$ ImageNet diffusion model used in the original attack of~\citet{nie2022diffusion}. 
    \item Neural Compression: a range of $22$ models with different quality factors; see details below.
\end{itemize}
The \underline{underlined} values above correspond to transformations that do not change the image (showing the maximum of robustness in each subplot of~\cref{fig:robustness}).
The \textbf{bold} values are used to, following prior work~\citep{wen2023tree}, summarize the results to a single score per transformation/attack type in~\cref{tab:baselines}, where we average the $4$ valuemetric scores and $3$ geometric scores independently. 
For neural compression we describe how we compute the score below.
Visual examples of each bold transformation/attack are shown on a real \taming generation in~\cref{fig:transformation_examples}.
\input{figures/appendix/transformation_examples.tex}

\paragraph{Neural compression.}
For neural compression we use models from the \texttt{CompressAI}~\citep{begaint2020compressai} library:
\begin{itemize}[itemsep=1pt, topsep=1pt, leftmargin=16pt]
    \item \textsc{BMSHJ18 (Factorized)}~\citep{bmshj18} with quality factors $q \in \{1,3,6\}$.
    \item \textsc{BMSHJ18 (Hyperprior)}~\citep{bmshj18} with quality factors $q \in \{1,3,6\}$.
    \item \textsc{CSTK20 (Anchor)}~\citep{cstk20} with quality factors $q \in \{1,3,6\}$.
    \item \textsc{CSTK20 (Attention)}~\citep{cstk20} with quality factors $q \in \{1,3,6\}$.
    \item \textsc{MBT18}~\citep{mbt18} with quality factors $q \in \{1,3,6\}$.
    \item \textsc{MBT18 (Scale)}~\citep{mbt18} with quality factors $q \in \{1,3,6\}$.
\end{itemize}
To sort these by compression strength we compute \emph{bpp} (bits per pixel) as done in the library:
\begin{equation}
    \frac{\sum_i \log L_i}{-\ln 2 \cdot n_{\text{pix}}},
\end{equation}
where $L$ is the likelihood vector and $n_{\text{pix}}$ is the number of pixels in the image.
Empirically we observe bpp of around $0.1$ ($q=1$), $0.3$ ($q=3$), and $1.0$ ($q=6$).
When reporting a single score for neural compression we average the six scores with $q=3$. 
Additionally, we evaluate the following four autoencoders from the \texttt{diffusers}~\citep{von-platen-etal-2022-diffusers} library for which we compute bpp manually by considering the downscaling factor in the latent space, the latents size, and $16$-bit/$32$-bit floating precision. 
We see that our calculations are consistent with the results of neural compressors from \texttt{CompressAI}:
\begin{itemize}[itemsep=1pt, topsep=1pt, leftmargin=16pt]
    \item The Stable Diffusion VAE~\citep{rombach2022high} (\texttt{stabilityai/sd-vae-ft-ema}; \textsc{SD VAE (ft-EMA)}), with bpp $2$.
    \item The Stable Diffusion XL VAE in half precision~\citep{sdxl} (\texttt{madebyollin/sdxl-vae-fp16-fix}; \textsc{SDXL VAE (fp16)}), with bpp $1$.
    \item Deep Compression AE~\citep{dcae} (\texttt{mit-han-lab/dc-ae-f64c128-in-1.0-diffusers}; \textsc{DC-AE}), with bpp $1$.
    \item VAE of Flux~\citep{flux2024} (from the \texttt{black-forest-labs/FLUX.1-schnell} pipeline; \textsc{FLUX VAE}), with bpp $1$.
\end{itemize}

\subsection{Details of Audio Experiments}\label{app:expdetails:audio}

\paragraph{Audio prompt generation.}

We observed that \moshi frequently generates brief responses and typically expects human interaction to continue the conversation. 
When using conventional text prompts such as those from Alpaca~\citep{alpaca} or Open Assistant Conversations~\citep{kopf2023openassistant} datasets, the model rarely produced audio outputs of sufficient length (\eg $10$ seconds). 
We therefore synthesized specialized prompts designed for this particular use case.
These prompts are used when prompting the \moshi model to generate (possibly watermarked) audio, such as in \cref{sec:audio}, \aref{app:consistency:audio}, and \aref{app:moreresults:audio}.

To create a diverse collection of audio monologue topics, we leveraged \textsc{Llama 3.1-8B-Instruct} to generate 1000 unique text prompts. 
We guided the model using a system+user template to produce concise single-sentence requests (each beginning with action verbs like ``\emph{Describe}'', ``\emph{Talk about}'', etc.) covering distinct subjects. We then filtered out near-duplicates by calculating pairwise Rouge-L scores~\citep{lin2004rouge} (using a threshold of $0.7$) and eliminated texts that fell outside our desired length parameters.
Representative examples include: ``\emph{Describe the life cycle of a butterfly and the symbolic meanings associated with it.}'', ``\emph{Explain the process of photosynthesis in plants and its importance to ecosystems.}'', or ``\emph{Discuss the cultural significance of traditional Japanese tea ceremonies.}''.
Finally, we converted these text prompts into audio using the \textsc{SeamlessV2}~\citep{barrault2023seamlessm4t} (large) model, saving each resulting waveform alongside its corresponding source prompt.
The resulting audio prompts average approximately $4$ seconds in length.

\paragraph{Audio transformations.}
We evaluate robustness to a set of audio edits grouped into valuemetric, time-frequency, and neural compression transformations. 
When evaluating (\eg in \cref{tab:audio}), each is applied with the following fixed strengths:
\begin{itemize}[itemsep=1pt, topsep=1pt, leftmargin=16pt]
    \item Valuemetric:
    \begin{itemize}
        \item Bandpass Filter: ($300$,$3000$), ($500$,$5000$), ($1000$,$8000$) Hz.
        \item Highpass Filter: $100$, $500$, $1000$ Hz.
        \item Lowpass Filter: $1000$, $3000$, $8000$ Hz.
        \item Noise Injection (white): std = $0.001$, $0.01$, $0.05$.
        \item Pink Noise: std = $0.01$, $0.05$, $0.1$.
        \item Echo: (delay = $0.1$ s, vol = $0.2$), ($0.3$ s, $0.5$), ($0.5$ s, $0.7$).
        \item Smooth: window fraction = $0.001$, $0.005$, $0.01$.
        \item Boost Audio: $+50$ \%, $+90$ \%.
        \item Duck Audio: $-50$ \%, $-90$ \%.
        \item MP3 Compression: bitrate = $16$, $64$, $128$ kbps.
    \end{itemize}
    \item Time-frequency:
    \begin{itemize}
        \item Speed: factor = $0.75$, $0.9$, $1.0$, $1.1$, $1.25$.
        \item Temporal Crop: keep $50$ \%, $70$ \%, $90$ \% of duration.
        \item Time Shift: shift = $10$ ms, $20$ ms, $40$ ms.
        \item Up/Down Resample: intermediate = $24$ kHz, $36$ kHz, $48$ kHz.
    \end{itemize}
    \item Neural Compression:
    \begin{itemize}
        \item DAC Compression ($24$ kHz): full model pass.
        \item EnCodec Compression ($24$ kHz): full model pass.
    \end{itemize}
\end{itemize}

We use the same implementation as in AudioSeal~\citep{sanroman2024proactive} when the augmentations are available.
For DAC~\citep{kumar2023high} and EnCodec~\citep{defossez2022high} we use the official models at 24~kHz.

\paragraph{RCC finetuning.}
We perform finetuning for 200 epochs with 1000 steps per epoch on batches of 64 audio clips of 10-seconds from VoxPopuli~\citep{wang2021voxpopuli}, using 2 H200 GPUs for 1 day.
We use the AdamW optimizer~\citep{kingma2015adam} with a base learning rate of $2\times10^{-5}$, linear warmup over 5 epochs, and cosine annealing down to $2\times10^{-7}$. 
We set $\lambda$ to 0.01 for the regularization loss in the \ftaug model (with transformations), while using 0.001 in the \ft model (without transformations).
For the regularization loss we use the Multi-Resolution STFT between the audios reconstructed either with the original decoder $D_0$ or the finetuned decoder $D$.
Following notations from \cref{ssec:method:ft}, 
for the RCC loss we use the MSE between $z$, the soft latents before the projection and quantization step,  and $z'$, the soft latents generated by the encoder replica $E'$.
To improve robustness for the \ftaug model, we apply augmentations $\mathcal{A}$ from the start, sampling one augmentation per batch. 
The augmentations are chosen randomly at each step, and the parameters are sampled uniformly from the ranges below:
\begin{itemize}[itemsep=1pt, topsep=1pt, leftmargin=16pt]
  \item Lowpass filter: cutoff 2000-6000\,Hz
  \item Highpass filter: cutoff 200-600\,Hz
  \item White noise injection: std 0.001-0.01
  \item Pink noise: std 0.001-0.01
  \item Smooth: window fraction 0.001-0.005
  \item Time shift: 0.3-10\,ms
\end{itemize}

%% file: figures/appendix/stratification.tex
\begin{wrapfigure}[20]{r}{0.38\textwidth}
      \centering
          \includegraphics[width=\linewidth]{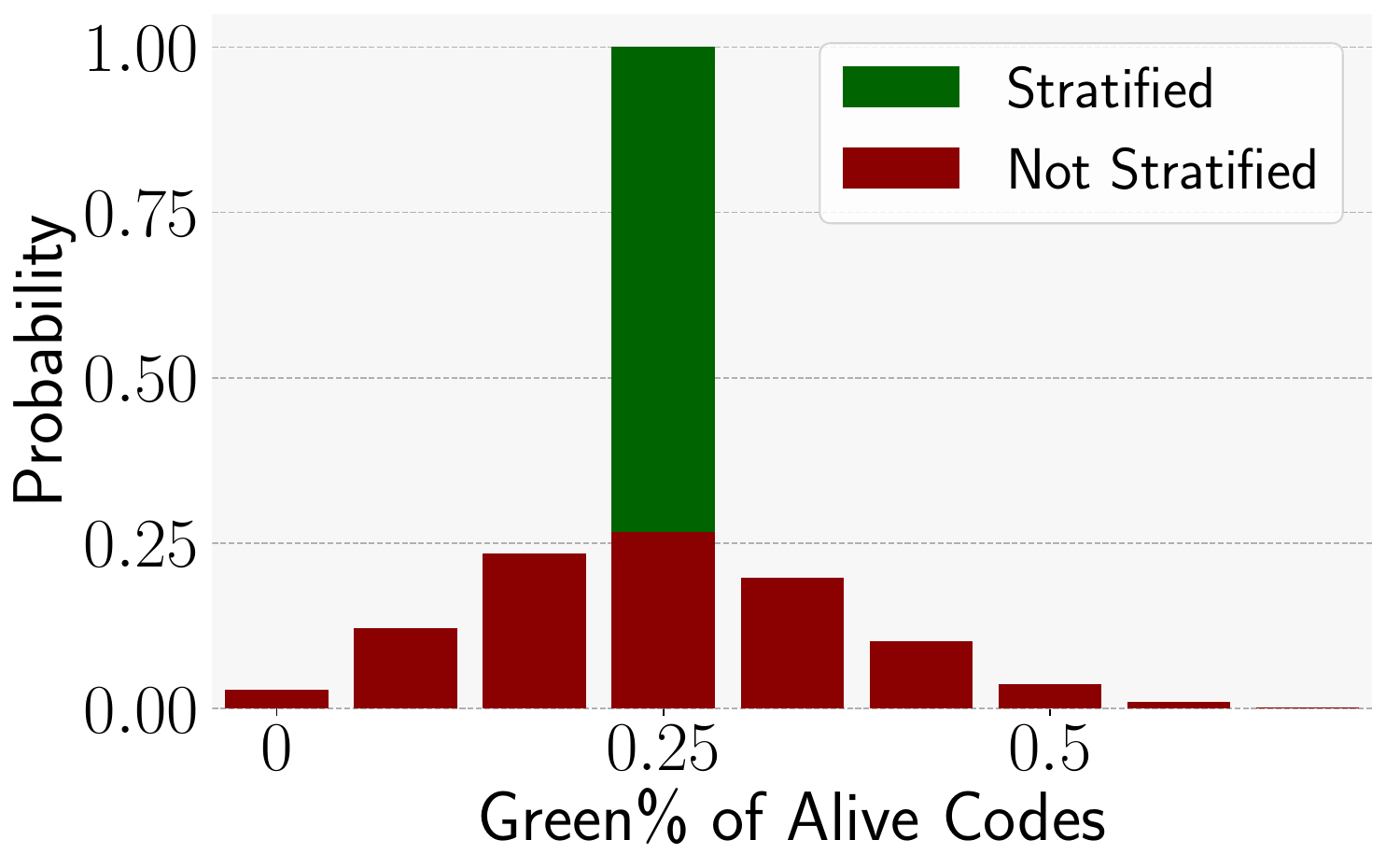}
          \caption{When the number of alive codes $\nalive$ is much smaller than the codebook size $|V|$, green/red splits may cause the green ratio within alive codes $\gamma'$ to be different from the expected $\gamma$ (red). In this figure, $\nalive=12, |V|=192, \gamma=0.25$. Stratification, \ie separate splitting of red and green tokens, resolves this issue (green).
          }
    \label{fig:stratification} 
  \end{wrapfigure}
     

%% file: figures/appendix/transformation_examples.tex
\begin{figure}[t]
    \centering
    \includegraphics[width=0.9\linewidth]{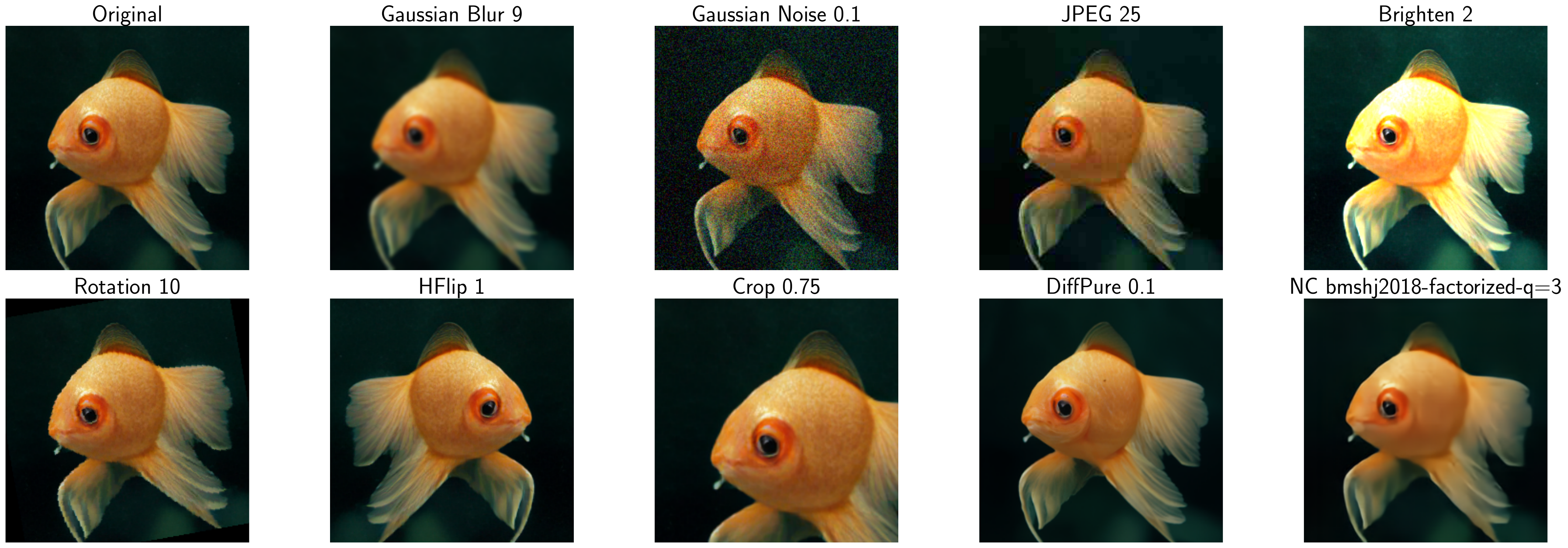}
    \caption{Examples of transformations with parameters used to compute the scores in~\cref{tab:baselines}.}
    \label{fig:transformation_examples}
    \vspace{-1em}
  \end{figure}

%% file: sections/app-moreresults.tex
\section{Additional Experimental Results} \label{app:moreresults}
In this section, we present additional results: ablations of RCC finetuning (\aref{app:moreresults:ft}), investigations of different watermarking parameters (\aref{app:moreresults:ablations}), additional results for \cham and \rarxl on token match, watermark power, quality, and robustness (\aref{app:moreresults:powerandrobustnessmoremodels}), comparison of the original and final decoders via PSNR (\aref{app:moreresults:decoderquality}), validation of statistical test correctness (\aref{app:moreresults:fpr}), comparison to generation-time watermarks (\aref{app:moreresults:gentime}), analysis of token entropy (\aref{app:moreresults:entropy}), efficiency measurements (\aref{app:moreresults:efficiency}), and omitted audio results (\aref{app:moreresults:audio}).

\subsection{Finetuning Ablations}\label{app:moreresults:ft}

\input{tables/appendix/ft_ablations.tex}

We train five more finetunes of \taming to test the influence of different parameters: 
\begin{itemize}[itemsep=1pt, topsep=1pt, leftmargin=8pt]
    \item $\lambda=10$ uses a higher regularization weight, \ie puts less weight on the RCC loss.
    \item $\lambda=0.1$ uses a lower regularization weight, \ie puts more weight on the RCC loss.
    \item $lr=10^{-5}$ uses a lower learning rate.
    \item $lr=10^{-3}$ uses a higher learning rate.
    \item \textsc{FT+Augs-All} finetunes all components of the VQGAN, including the codebook. 
\end{itemize} 
The results are presented in \cref{tab:ft_ablations} and visual examples in \cref{fig:ft_ablations} where the first row shows a detokenized output and the second row zooms in on the top-left $64 \times 64$ pixel region.
Our baseline here is our \ftaug variant (we do not consider synchronization in this experiment). 
We see that increasing $\lambda$ to $10$ slightly degrades the results (likely within the level of experimental noise) and reducing the learning rate reduces robustness. 
We also experimented with training for more epochs with a lower learning rate, but this did not lead to better results than our best variant. 

We also see that $\lambda=0.1$, $\lr=10^{-3}$, and \textsc{FT+Augs-All} lead to much better results.
However, the visual results in \cref{fig:ft_ablations} show that all three of these variants lead to degradation in image quality.
For \textsc{FT+Augs-All} this is the most evident, which motivates our discussion above regarding the importance of carefully choosing which modules to finetune.
The artifacts in $\lr=10^{-3}$ clearly show that the learning rate is too high for stable training.
For $\lambda=0.1$ the artifacts are clearly visible in the second row of the figure, showing that this setting puts too much weight on the RCC loss.

\input{figures/appendix/ft_ablations.tex}
\paragraph{Training on transformer-generated tokens.}
Finally, we hypothesized that using token sequences generated by the transformer as the training set instead of tokenizations of ImageNet images would improve finetuning, as the former more closely matches the distribution of inputs that the detokenizer sees at evaluation time.
Another experiment we tried was using a mixture of these two token sequences.
We did not observe any benefits of this approach.

\subsection{Watermark Parameters} \label{app:moreresults:ablations}
In the following experiments, starting from \ftaug on \taming, we vary the main watermarking parameters: context size $h$, strength $\delta$, green ratio $\gamma$, the choice of watermark context, and the partitioning strategy, presenting the key results in \cref{tab:wm_ablations}.

For the context size $h$, we observed that $h>1$ generally led to non-robust watermarks.
We explore both $h=2$ and $h=3$ and both the standard choice of watermark context (preceding $h$ tokens) and the image-specific one (the $h$ tokens spatially close, \eg for $h=3$ the tokens above the current token, to the left of it, and top-left of it in the image; denoted \textsc{Spatial}).
As \cref{tab:wm_ablations} shows, all of these variants significantly degrade robustness compared to the baseline (\ftaug).
\looseness=-1
For adversarial purification and neural compression, the results are comparable to the non-ablated results \emph{without finetuning}. 

This is in line with the intuitive understanding of $h$ from prior work~\citep{kirchenbauer2023watermark,zhao2023provable}: large $h$ makes watermark removal easier, as changing any of the preceding $h$ tokens changes the red/green split at the following token.
On the other hand, low $h$ makes the watermark less secure, \ie easier to forge.
As noted above, we hypothesize that $h=0$ is a more viable choice for images (and thus use it for \cham), as reverse-engineering of the watermark rules (as successfully done for $h=0$ in text~\citep{jovanovic2024watermark,milpstealing}) is likely much more difficult due to the complex image tokenizer being hidden.

Regarding strength, we found that $\delta=2$ is for both our models the strongest watermark that does not degrade quality.
As seen in \cref{tab:wm_ablations}, $\delta=1$ degrades robustness, while $\delta=4$ greatly boosts it at the cost of generation quality: for all rows in \cref{tab:wm_ablations}, FID is lower or comparable to the baseline value, while $\delta=4$ increases it to $18.7$.
Across our experiments we found that $\gamma=0.25$ leads to slightly better tradeoffs overall than $\gamma=0.5$, as illustrated in the last row of \cref{tab:wm_ablations}.

Finally, we briefly experimented with a semantic partitioning strategy, related to similar attempts for LLMs~\citep{hou2023semstamp}.
In particular, instead of choosing green tokens $\mathcal{G}$ uniformly at random from the vocabulary at each partitioning step, we use k-means clustering to partition the hard embeddings of each token in the vocabulary into $k=100$ clusters and assign colors such that all tokens in the same cluster have the same color, while keeping the overall green ratio at $\gamma$.
In theory, this should make the watermark more robust to modifications that do not change the semantics, as the resulting change in the token would hopefully not leave the k-means cluster and thus remain green.
On the other hand, it might make it harder for the decoding-time watermark to replace a green token with a red one, as all tokens from the same k-means cluster, which may be good alternatives, are also red. 
In practice, we consistently observed higher robustness but at the cost of significant quality degradation.
More work is needed to understand how to find a good tradeoff with this approach.
\input{tables/appendix/wm_ablations.tex}

\subsection{Full Results for \cham and \rarxl} \label{app:moreresults:powerandrobustnessmoremodels}

In~\cref{fig:cham_power} and~\cref{fig:cham_robustness} we present token match and watermark power results for \cham and \rarxl, complementing the \taming results presented above in~\cref{fig:power}.
Similarly, in~\cref{fig:rarxl_power} and~\cref{fig:rarxl_robustness} we present robustness results for \cham and \rarxl, similar to those in~\cref{fig:robustness}.

Additionally, we report FID results on \cham and \rarxl using $50{,}000$ images as for \taming, where for \cham we use $10$ independent generations per each of the $5000$ COCO validation prompts (see~\aref{app:expdetails:main}).
Our conclusions match those made in the main text.
None of \base, \ft, or \ftaug exceed the unwatermarked FID of $19.7$ for \cham and $11.5$ for \rarxl.
For these models synchronization does not increase FID.
We note that we obtain slightly higher baseline FID for \rarxl compared to \citet{yu2024randomized}; we suspect that this is due to sampling differences (we used the default parameters from the project repository) and did not investigate this further, as for our study only the \emph{difference} in FID is of interest.
For all models, we also visually confirm the quality of the generated images; we present examples in \aref{app:qualitative}.

\subsection{Distance to the Original Decoder}\label{app:moreresults:decoderquality}

FID assesses the quality of the generative model by comparing distributions of generated and real images.
Our generation-time watermark modifies each sampling step, which results in a completely different generation (\emph{semantic watermarking}, see~\cref{fig:gimmick}). 
This makes direct comparison between image pairs not viable as a quality metric in our case.
However, using the same latent representation and watermarking parameters, we can directly compare the image decoded by the original decoder with the one decoded by the RCC-finetuned decoder.
While not necessarily the best proxy for image quality, this experiment gives us some insight into how much the decoder behavior changes after finetuning.

To this end, we compute the average PSNR over $1000$ image pairs, where in each pair, one image is decoded using the decoder from \base, and the other using the decoder from \ft or \ftaug, or using both the decoder and synchronization layer from \ftaugsync.
Even with synchronization, which as expected reduces the PSNR the most, the average PSNR is $37.6$ for \taming, $39.5$ for \cham, and $29.0$ for \rarxl, indicating that the final decoding does not significantly deviate from the original.
\clearpage
\input{figures/appendix/cham_power.tex}
\input{figures/appendix/cham_robustness.tex}
\input{figures/appendix/rarxl_power.tex}
\input{figures/appendix/rarxl_robustness.tex}
\clearpage

\subsection{Empirical Validation of Statistical Test Correctness} \label{app:moreresults:fpr}
\input{figures/appendix/fpr.tex}
We empirically validate the correctness of our statistical test by computing p-values of our watermark detector on unwatermarked images. 
In~\cref{fig:fpr} we show the distribution of such p-values on $50{,}000$ unwatermarked images generated with the base model of \taming, using the corresponding watermark ($\gamma=0.25,h=1$) across $10$ different random seeds for the watermark.
We observe that the distribution is roughly uniform; this holds also for each individual seed. 

We push this investigation further by running our detector on huge token sequences, as in~\citet{radioactive}, despite those not being crucial to our usecase.
Namely, for $10$ random seeds, we $10$ times independently concatenate $50{,}00$ images to obtain a sequence of above $1$M tokens, and compute the p-value on prefixes of this sequence of increasing length.
We show the results in \cref{fig:fpr_extra} (\emph{left}, mean and standard deviation over $100$ runs described above).
Interestingly, while we would expect convergence around $0.5$, the p-values of extremely long token sequences become as high as $0.8$.

The reason for this is the overlooked relationship between the effective vocabulary size (\ie the set of alive codes, see~\aref{app:expdetails}) $\nalive=971$ and the watermark parameter $\gamma=0.25$.
Specifically, as the number of green tokens $\gamma \cdot \nalive = 242.75$ in the vocabulary assumed by our test (\cref{eq:pvalue}) is not an integer, it is impossible for us to make a red/green split that will exactly match $\gamma=0.25$.
Instead we have to either select $242$ or $243$ green tokens.
We conservatively choose $242$, which is the choice that keeps the statistical test sound: Now, the \emph{real} expected green ratio $\gamma'=242/971 \approx 0.2492$ is smaller than $\gamma=0.25$ used by the test, which only sacrifices some power as we use a one-sided test.

This seems overlooked in prior work as well but happens to be rarely relevant as $\gamma \cdot |V|$ is an integer for common vocabulary sizes and choices of $\gamma$.
To further confirm that this is the main cause for the observed behavior in~\cref{fig:fpr_extra}, we repeat the experiment using the correct $\gamma' \approx 0.2492$ instead of $\gamma = 0.25$ in~\cref{eq:pvalue} and present the results in \cref{fig:fpr_extra} (\emph{right}).
We observe that the p-values now converge to slightly below $0.5$, which matches results in prior work~\citep{radioactive}.
\input{figures/appendix/fpr_extra.tex}

\subsection{Comparison to Generation-time Watermarks for Diffusion Models} \label{app:moreresults:gentime}
\input{tables/appendix/gentime.tex}
In~\cref{tab:gentime} we present results on generation-time watermarks for image models.
As no prior work studies autoregressive models, we show results for diffusion models.
In particular, we study \texttt{stabilityai/stable-diffusion-2-1-base}~\citep{rombach2022high}. 
For \textsc{Tree-Ring}~\citep{wen2023tree} we use the official implementation and set 
the watermark pattern to \texttt{ring}, \texttt{w\_channels} to $3$ and use $50$ inference steps for generation and testing.
For \textsc{Stable Signature}~\citep{fernandez2023stable} we use the official implementation and set \texttt{unconditional\_guidance\_scale} to $9$, \texttt{steps} to $50$, and use \texttt{PLMSSampler} with \texttt{ddim\_eta} set to $0$. 
For \textsc{Gaussian Shading}~\citep{yang2024gaussian} we use the official implementation, enable \texttt{chacha\_encryption}, set the number of inference steps for generation and inversion to $50$ and the number of bits to $256$ with \texttt{channel\_copy} set to $1$ and \texttt{hw\_copy} to $8$.

We observe that each watermark is either fragile to geometric transformations or to adversarial purification and neural compression.

\subsection{Token Entropy Analysis}
\label{app:moreresults:entropy}

\begin{figure}[h!]
    \centering
    \includegraphics[width=0.9\linewidth]{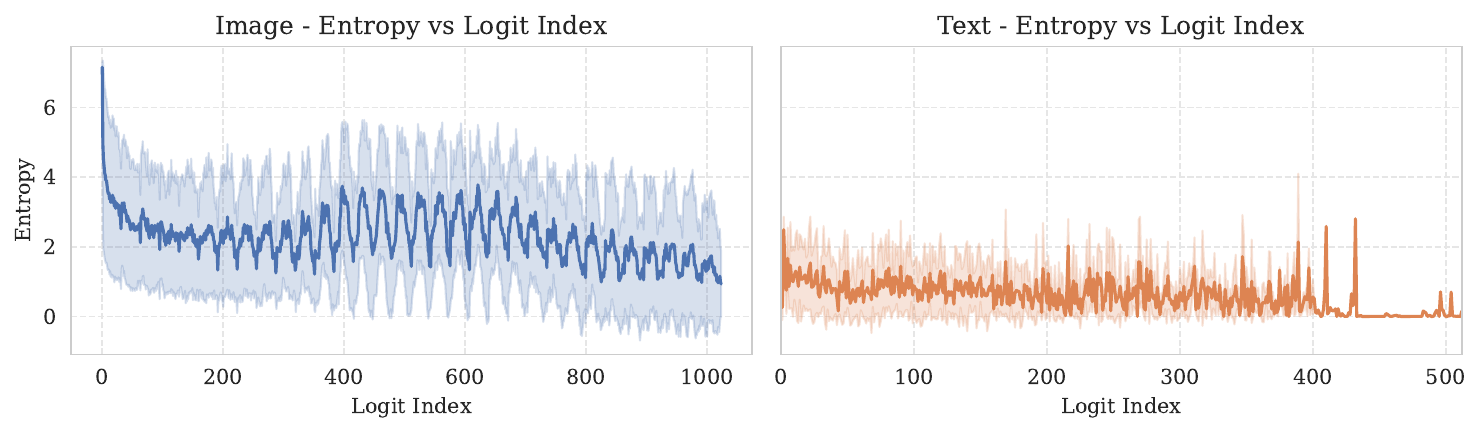}
    \caption{
        Entropy of image tokens as a function of token index.
        A clear periodic pattern appears, corresponding to the row-wise generation process 
        (period $\approx 32$ tokens per row).
    }
    \label{fig:entropy_pattern}
\end{figure}

We analyze the entropy of next-token distributions for image and text modalities to better understand modality-specific differences relevant to token-level watermarking. 
Entropy is measured in the same setup as our joint-modality experiment (\cref{ssec:experiments:interleaved}), using the \cham model, across $20$ generations ($\sim 20k$ logits) for each modality. 
The results are presented in \cref{fig:entropy_pattern}.
Both distributions exhibit a high peak near zero entropy, but image tokens show higher mean and a broader spread, with (mean = $2.93$, std = $1.90$) for images and (mean = $0.45$, std = $0.74$) for text. 
Thus, image tokens generally have higher uncertainty, potentially making them easier to watermark more reliably.

An interesting phenomenon emerges for autoregressive image models: token entropy follows a periodic pattern aligned with the raster (row-major) generation order. 
Specifically, entropy spikes at the start of each new image row (every $\approx 32$ tokens), reflecting structured uncertainty due to local context resets during generation. 
This behavior suggests that entropy-aware watermarking strategies could exploit such periodicity to improve robustness and detectability.

\subsection{Efficiency Considerations}
\label{app:moreresults:efficiency}

We evaluate the computational efficiency of the proposed watermark detection pipeline using the \taming model on a single H100 GPU. 
Average detection times are $0.8$s for synchronization, $6$ms for tokenization, and $0.5$s for watermark detection, 
resulting in a total latency of approximately $1.3$s per image. 
This is comparable to diffusion-model watermarks such as Tree-Ring~\cite{wen2023tree}, 
which we benchmark at around $1.9$s on an NVIDIA V100 GPU using default parameters from the official implementation.
Note that this detection time could be sped up with a better synchronization algorithm~\citep{fernandez2025geometric} and with an optimized detection implementation (which does not loop over tokens).

The tokenizer fine-tuning procedure introduces a modest one-time computational overhead of at most $32$ GPU hours per model, 
which is negligible relative to the $> 850{,}000$ GPU hours required to train models such as \cham~\cite{team2024chameleon} and makes the approach practical and scalable for deployment.

\subsection{Omitted Audio Results}\label{app:moreresults:audio}

\paragraph{RCC finetuning.}
We presented in \aref{app:expdetails:audio} the details of our finetuning approach. 
Here we discuss the validation metrics and demonstrate how finetuning influences RCC and token match.
We first evaluate the perceived audio quality of the generated samples using the PESQ and STOI metrics.
After finetuning, we achieve a PESQ~\citep{rix2001perceptual} score of $4.3$ for both \ft and \ftaug when compared to \base samples. 
STOI~\citep{taal2010short} scores reach $0.98$ for \ft and $0.99$ for \ftaug (we fixed the audio regularization loss weight to maintain approximately similar values for both approaches).
The resulting audio is very hard to discriminate from the original, although we observe that it sometimes lead to light humming artifacts.
\input{figures/appendix/ft_rcc_moshi_mimi.tex}
\input{figures/appendix/audio_big_boy.tex}

RCC and TM results are presented in \cref{fig:ft_rcc_moshi_mimi}.
We observe significant token match improvements after finetuning when considering sequences of tokens generated by reconstructing 10-seconds VoxPopuli audios with the \mimi tokenizer.
However, the finetuning process does not substantially improve token match for sequences generated by the \moshi model with audio prompts.
This may explain why finetuning does not significantly enhance watermark power for the \moshi model, as observed in \cref{tab:audio} of \cref{sec:audio}.
A potential approach for improving watermark TPR would be to specifically finetune the model on generated sequences, which we leave for future work.
(We attempted this approach for images but did not observe significant improvements, see \aref{app:moreresults:ft} for details.)

\paragraph{Detailed robustness results.}
In \cref{tab:audio} of the main paper, we report the average TPR at $1\%$ FPR over multiple audio-specific augmentations. 
\cref{tab:full-audio-robustness} provides the TPR for each individual augmentation that contributes to these averages, with a detailed view of how each transformation impacts watermark robustness.
\cref{tab:full-audio-robustness} also reports TPRs for different $\delta$ values. 
For instance, increasing $\delta$ to $4.0$ pushes TPR above $0.9$ across most augmentations, but the resulting audio quality deteriorates noticeably compared to lower-strength settings.

%% file: tables/appendix/ft_ablations.tex
\begin{table}[b]
    \centering
    
    \caption{TPR at 1\% FPR of finetuning ablations on \taming in the setting of~\cref{tab:baselines}. The ablations are described in~\aref{app:moreresults:ft}. Three of the variants improve the results; however,~\cref{fig:ft_ablations} shows that they also lead to significant degradation in image quality.}
    \newcommand{\fivecol}[1]{\multicolumn{5}{c}{#1}}  

    \definecolor{badred}{rgb}{0.8,0.2,0.2}
 
    \resizebox{0.65\linewidth}{!}{%
    \begingroup 
    \setlength{\tabcolsep}{5pt} %
    \begin{tabular}{l
        x{1}{2}
        x{1}{2}
        x{1}{2}
        x{1}{2}
        x{1}{2}
        }
        \toprule

        & {\textbf{None}} 
        & {\textbf{Valuemetric}} 
        & {\textbf{Geometric}} 
        & {\textbf{\shortstack[c]{Adversarial \\ Purification}}}
        & {\textbf{\shortstack[c]{Neural \\ Compression}}} \\

        \midrule

        \base & 0.99 & \textcolor{badred}{0.26} & \textcolor{badred}{0.01} & \textcolor{badred}{0.43} & \textcolor{badred}{0.48} \\
        \ftaug & 1.00 & 0.92 & \textcolor{badred}{0.01} & 0.70 & 0.79 \\
        \midrule
        $\lambda=10$ & 1.00 & 0.91 & \textcolor{badred}{0.01} & 0.68 & 0.77 \\
        $\lambda=0.1$ & 1.00 & 0.98 & \textcolor{badred}{0.01} & 0.85 & 0.96 \\
        $lr=10^{-5}$ & 0.99 & 0.75 & \textcolor{badred}{0.01} & 0.63 & 0.76 \\
        $lr=10^{-3}$ & 0.98 & 0.98 & \textcolor{badred}{0.01} & 0.81 & 0.78 \\
        \textsc{FT+Augs-All} & 1.00 & 1.00 & \textcolor{badred}{0.02} & 0.85 & 0.98 \\

        \bottomrule
    \end{tabular}
    \endgroup}
    \label{tab:ft_ablations}
\end{table}

%% file: figures/appendix/ft_ablations.tex
\begin{figure}[t]
      \centering
          \includegraphics[width=0.9\linewidth]{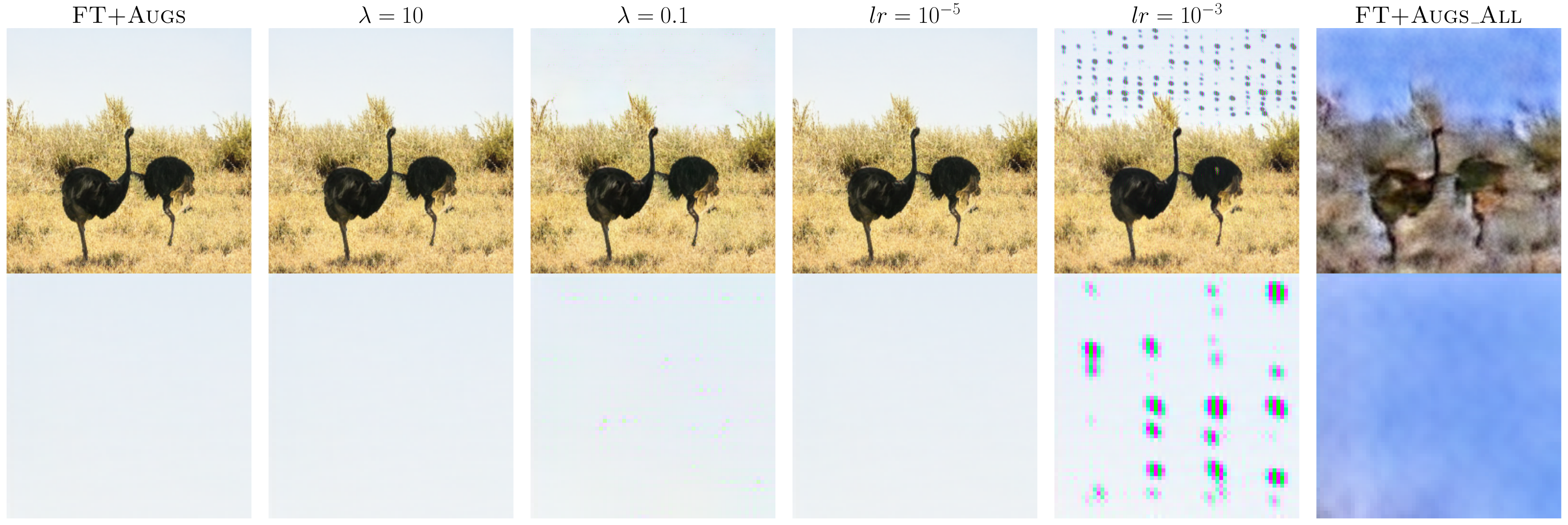}
          \caption{Visual examples of images produced by finetuning ablations introduced in~\aref{app:moreresults:ft}. The first row shows a detokenized image (based on the same token sequence as we fix the seed) and the second row zooms in on the top-left $64 \times 64$ pixel region to more clearly show artifacts.
          We see that the $\lambda=0.1$, $\lr=10^{-3}$, and \textsc{FT+Augs-All} variants lead to significant degradation in image quality.}
      \vspace{-1em}
    \label{fig:ft_ablations} 
  \end{figure}
      

%% file: tables/appendix/wm_ablations.tex
\begin{table}[t]
    \centering
     
    \caption{TPR at 1\% FPR of watermark parameter ablations ($\delta, h, \gamma$, context choice) on \taming in the setting of~\cref{tab:baselines} using \ftaug. The ablations and their results are discussed in~\aref{app:moreresults:ablations}.}
    \newcommand{\fivecol}[1]{\multicolumn{5}{c}{#1}}  

    \definecolor{badred}{rgb}{0.8,0.2,0.2} 
 
    \resizebox{0.7 \linewidth}{!}{%
    \begingroup 
    \setlength{\tabcolsep}{5pt} %
    \begin{tabular}{l
        x{1}{2}
        x{1}{2}
        x{1}{2}
        x{1}{2}
        x{1}{2}
        x{1}{2}
        }
        \toprule

        & {\textbf{None}} 
        & {\textbf{Valuemetric}} 
        & {\textbf{Geometric}} 
        & {\textbf{\shortstack[c]{Adversarial \\ Purification}}}
        & {\textbf{\shortstack[c]{Neural \\ Compression}}} & \textbf{FID} \\
 
        \midrule
        \textsc{\textsc{FT+Augs}} & 1.00 & 0.92 & \textcolor{badred}{0.01} & 0.70 & 0.79 & 16.33 \\
        \midrule
        \textsc{$h=2$} & 1.00 & 0.82 & \textcolor{badred}{0.01} & \textcolor{badred}{0.41} & \textcolor{badred}{0.57} & 15.70 \\
        \textsc{$h=3$} & 1.00 & 0.73 & \textcolor{badred}{0.01} & \textcolor{badred}{0.24} & \textcolor{badred}{0.39} & 15.87 \\
        \textsc{$h=3$\textsc{~(Spatial)}} & 1.00 & 0.69 & \textcolor{badred}{0.01} & \textcolor{badred}{0.22} & \textcolor{badred}{0.36} & 16.12 \\
        \textsc{$\delta=1$} & 1.00 & 0.69 & \textcolor{badred}{0.01} & \textcolor{badred}{0.29} & \textcolor{badred}{0.40} & 16.22 \\
        \textsc{$\delta=4$} & 1.00 & 0.97 & \textcolor{badred}{0.01} & 0.86 & 0.93 & 18.76 \\
        \textsc{$\gamma=0.5$} & 1.00 & 0.86 & \textcolor{badred}{0.01} & \textcolor{badred}{0.51} & 0.63 & 15.71 \\

        \bottomrule
    \end{tabular}
    \endgroup}
    \label{tab:wm_ablations}
\end{table}

%% file: figures/appendix/cham_power.tex
\begin{figure}[t]
    \centering
    \resizebox{0.75 \textwidth}{!}{
    \includegraphics[width=0.4\textwidth]{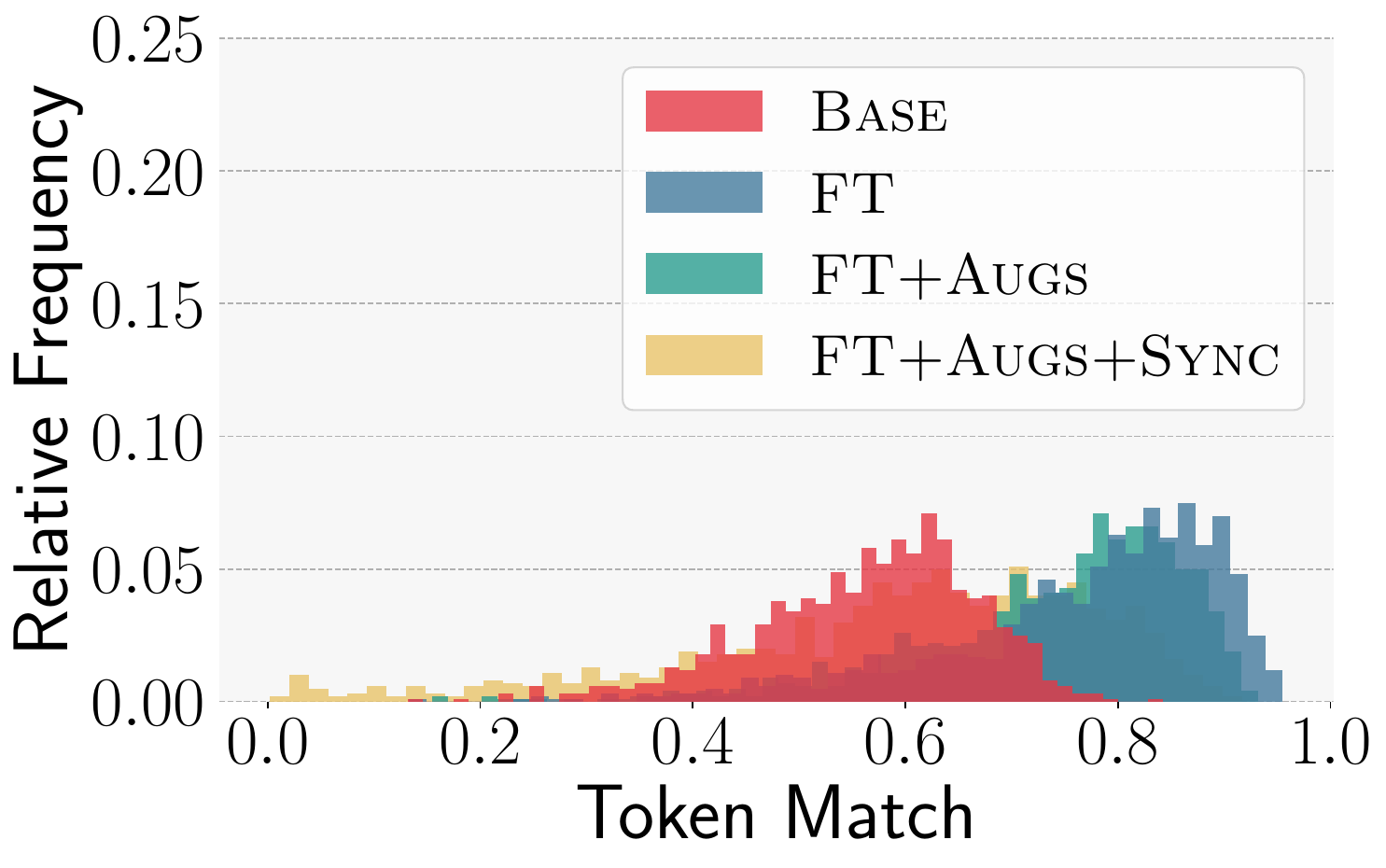}
    \quad \quad \quad \quad 
    \includegraphics[width=0.4\textwidth]{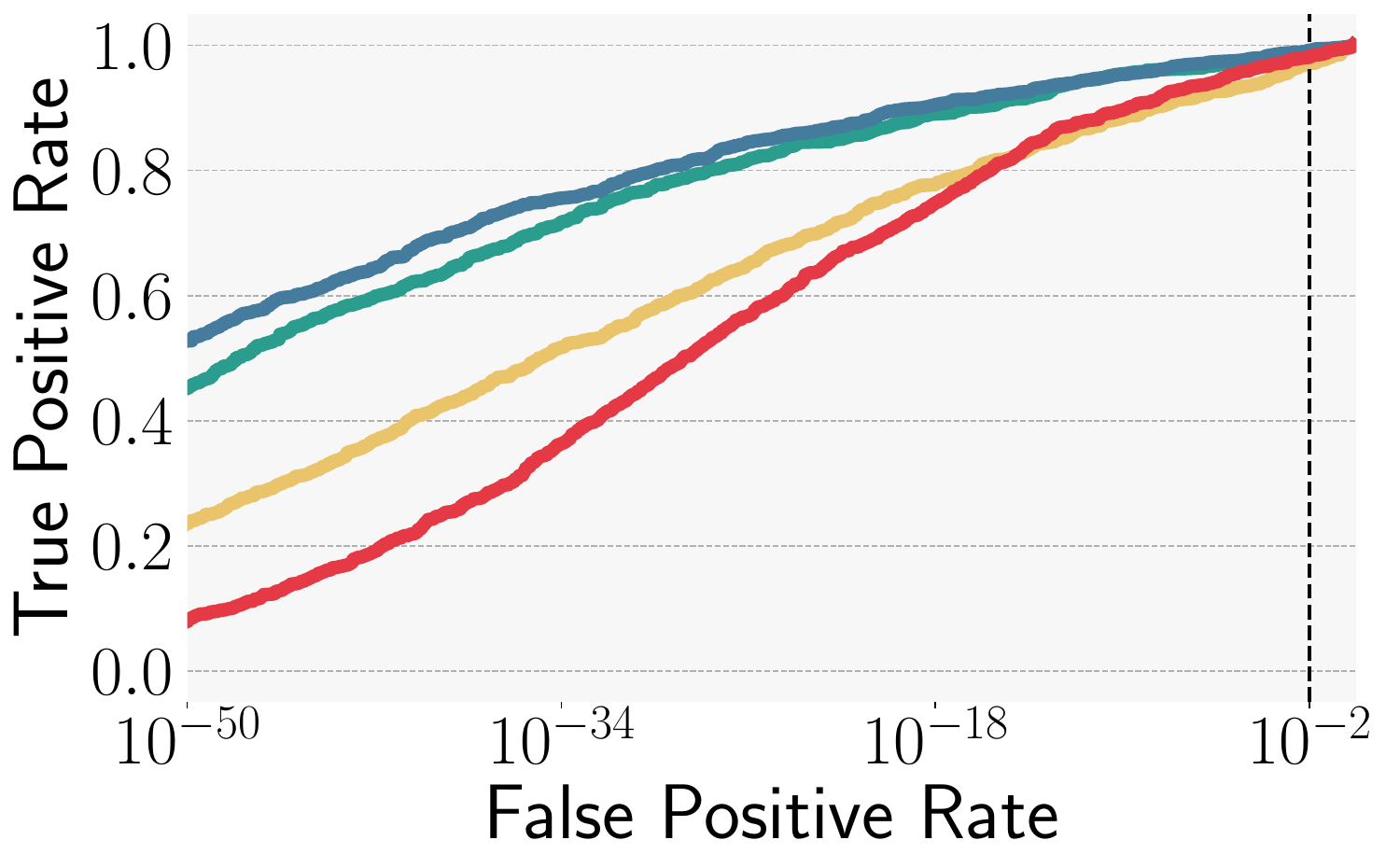}
    }
    \caption{Token match and watermark power results for \cham, analogous to~\cref{fig:power}.}
    \label{fig:cham_power}
\end{figure}  

%% file: figures/appendix/cham_robustness.tex
\begin{figure}[t]
    \centering
    \includegraphics[width=\textwidth]{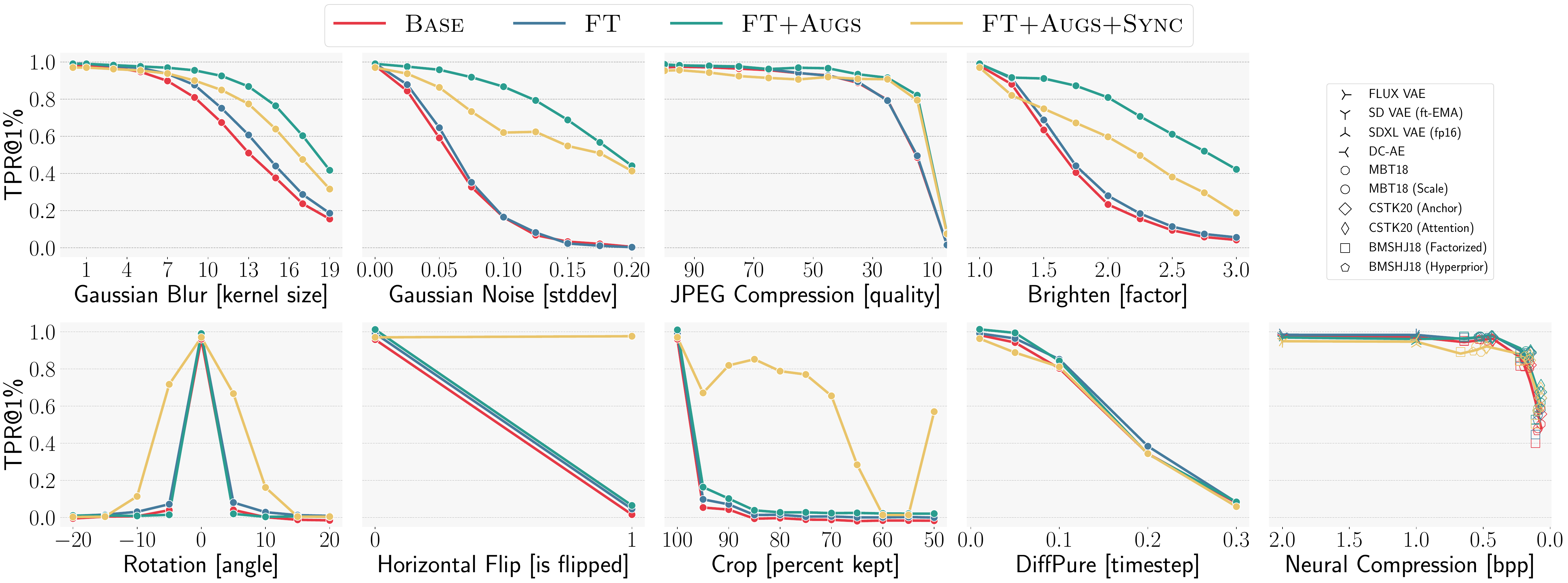}
    \caption{Robustness results for \cham, analogous to~\cref{fig:robustness}.}
    \label{fig:cham_robustness}
\end{figure}

%% file: figures/appendix/rarxl_power.tex
\begin{figure}[t]
    \centering
    \resizebox{0.75 \textwidth}{!}{
    \includegraphics[width=0.4\textwidth]{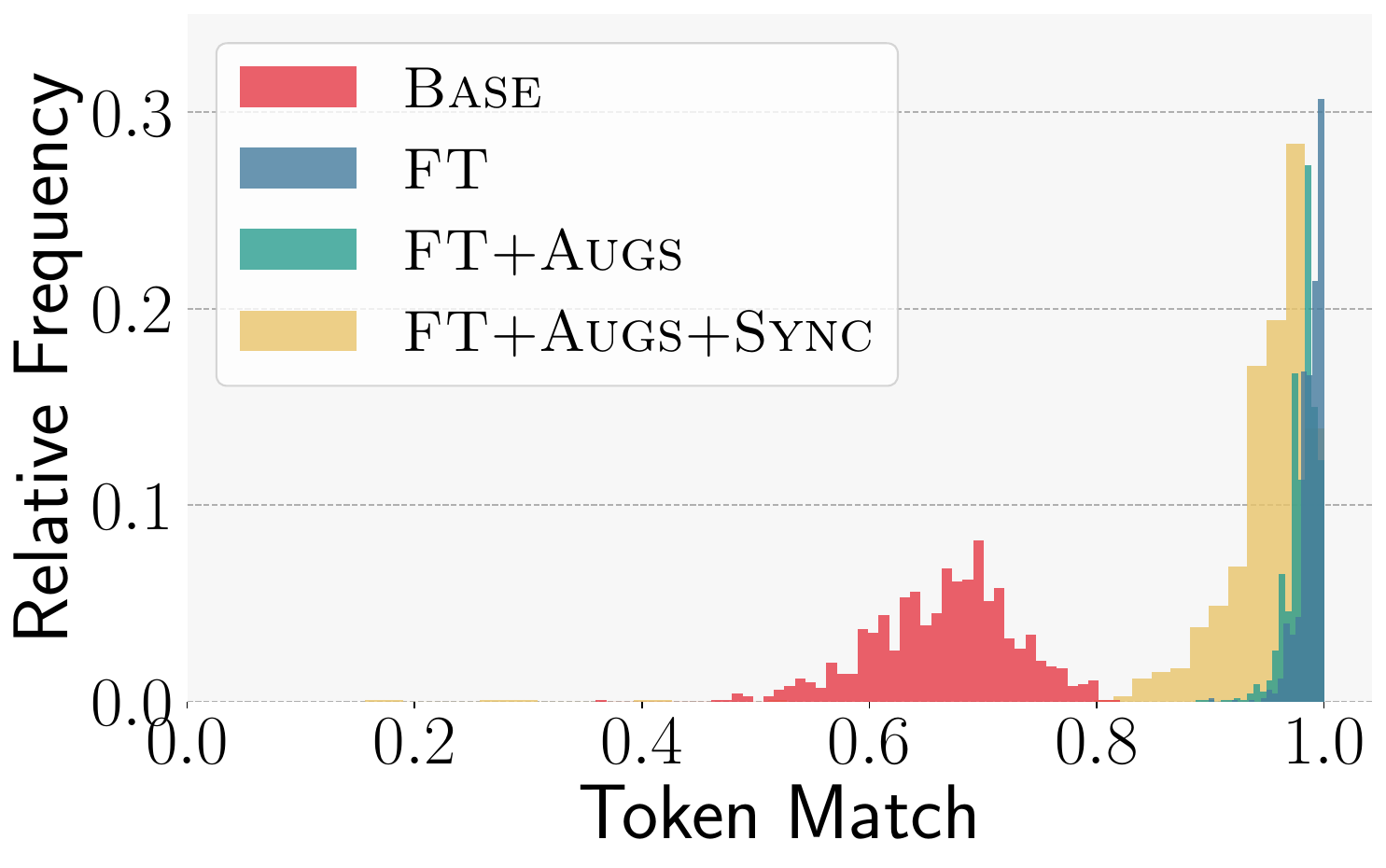}
    \quad \quad \quad \quad 
    \includegraphics[width=0.4\textwidth]{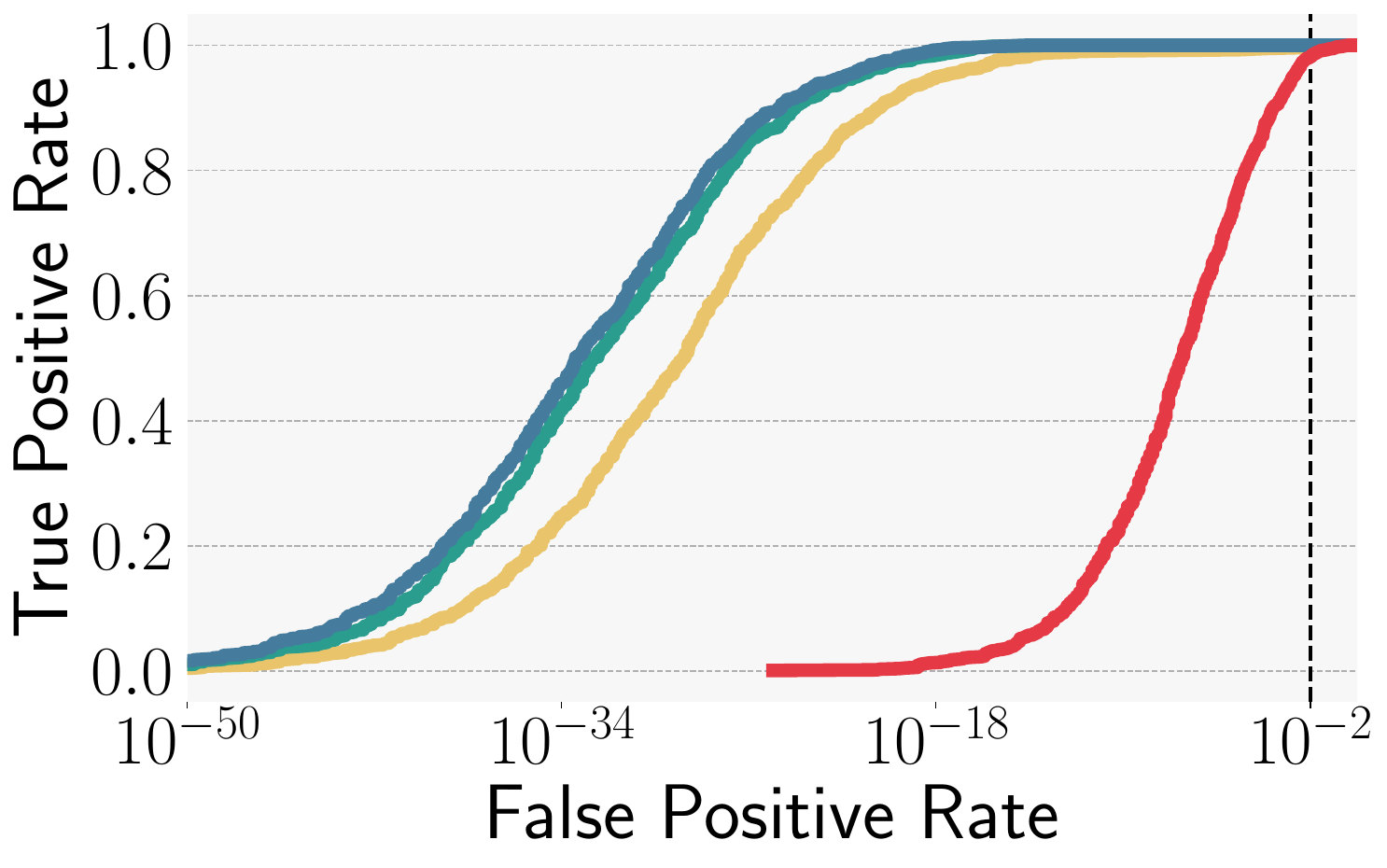}
    }
    \caption{Token match and watermark power results for \rarxl, analogous to~\cref{fig:power}.}
    \label{fig:rarxl_power}
\end{figure}   

%% file: figures/appendix/rarxl_robustness.tex
\begin{figure}[t]
    \centering
    \includegraphics[width=\textwidth]{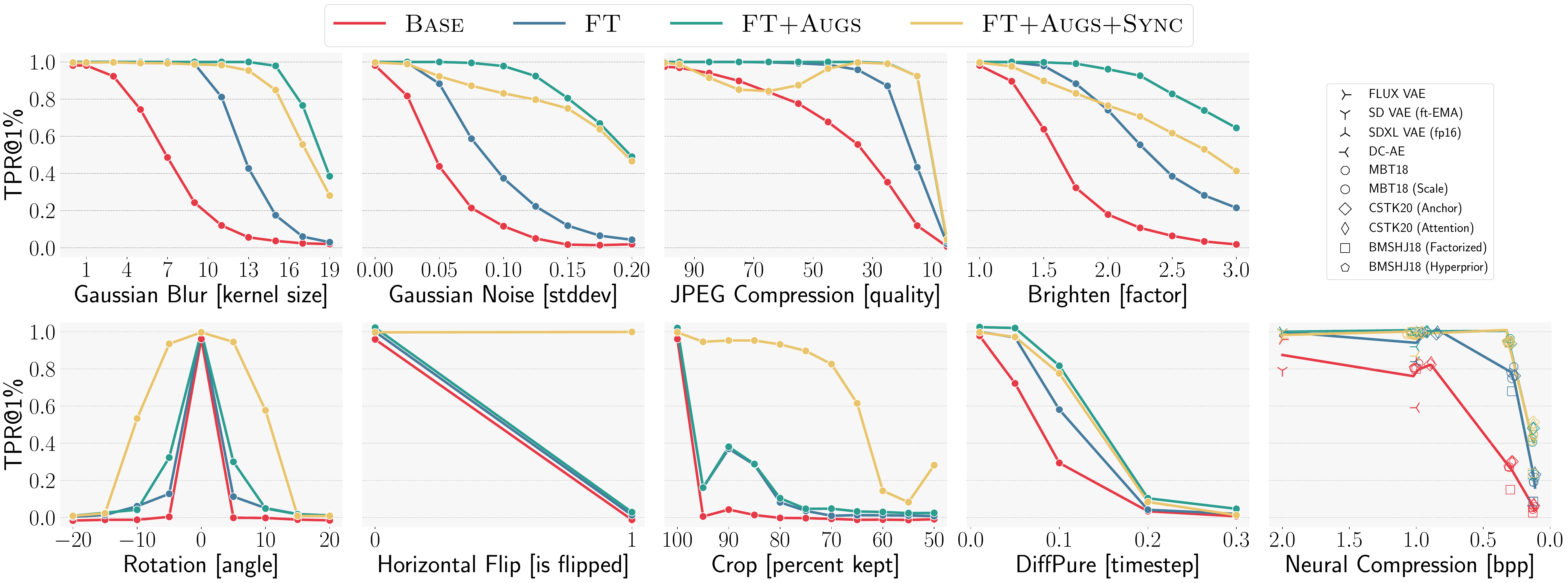}
    \caption{Robustness results for \rarxl, analogous to~\cref{fig:robustness}.}
    \label{fig:rarxl_robustness}
\end{figure} 

%% file: figures/appendix/fpr.tex
\begin{wrapfigure}[13]{r}{0.4\textwidth}
    \centering
		\includegraphics[width=\linewidth]{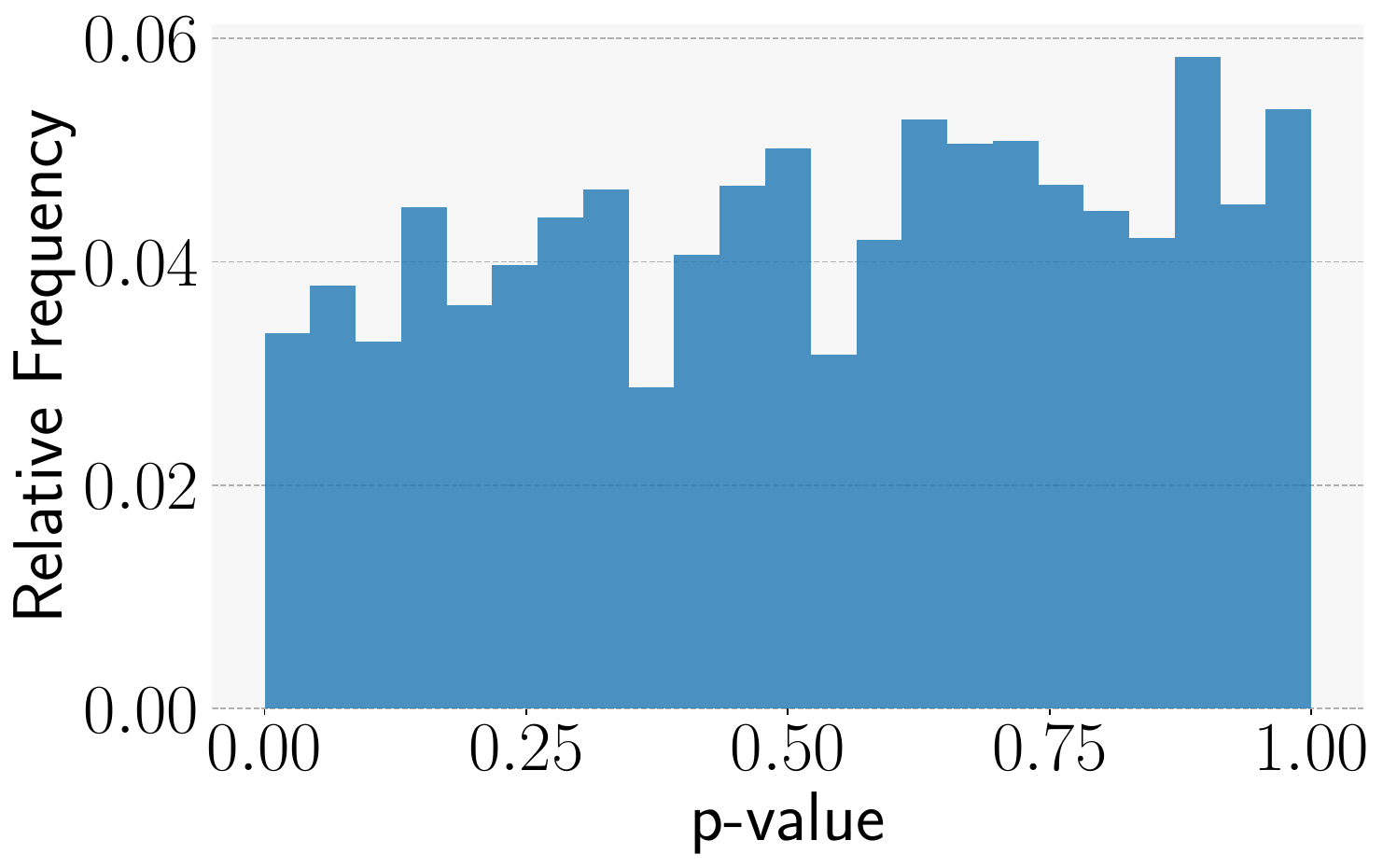}
        \caption{The distribution of p-values on unwatermarked images.}
  \label{fig:fpr} 
\end{wrapfigure}  

%% file: figures/appendix/fpr_extra.tex
\begin{figure}[t]
  \centering
  \resizebox{0.85 \textwidth}{!}{
  \includegraphics[width=0.5\textwidth]{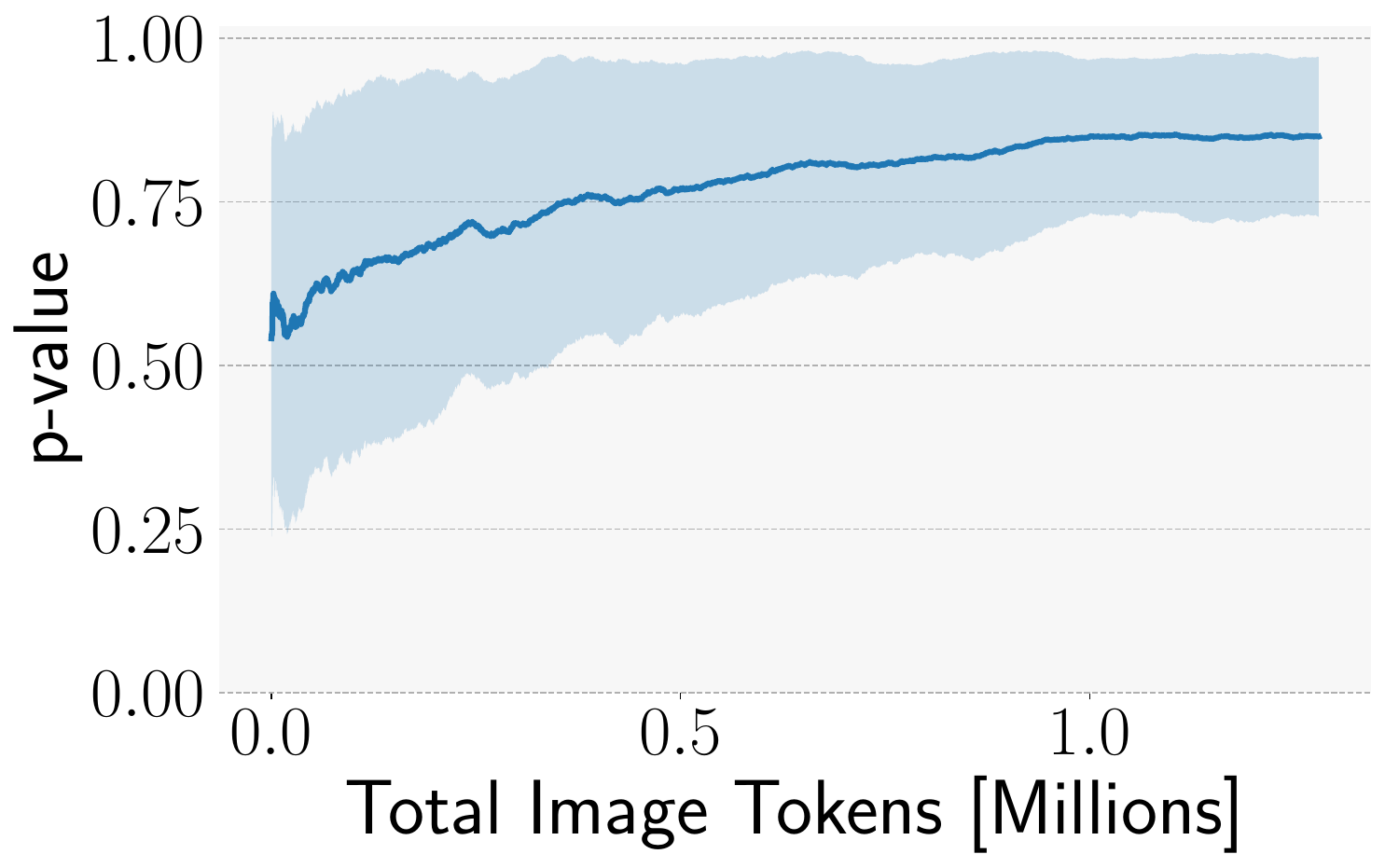}
  \quad
  \includegraphics[width=0.5\textwidth]{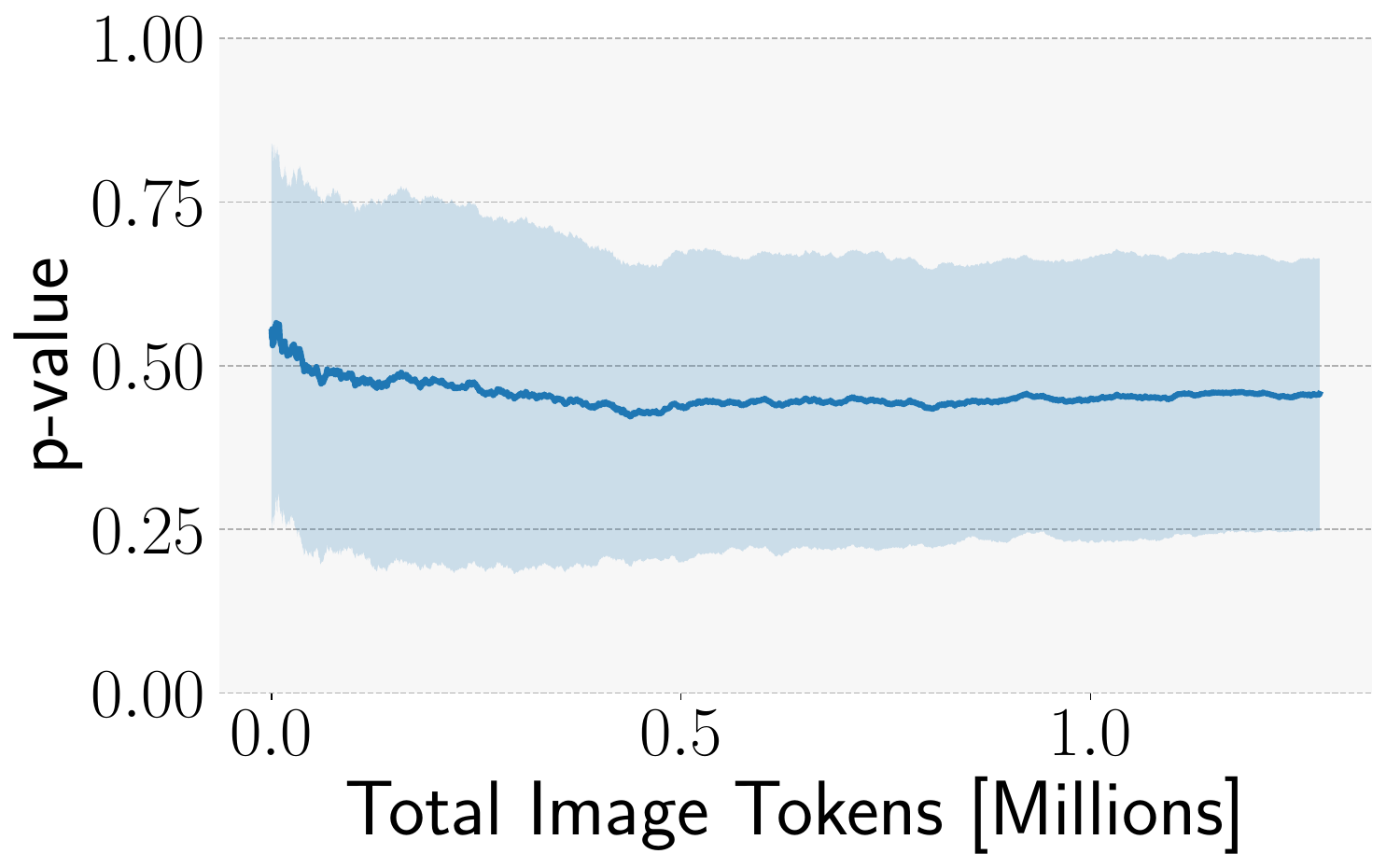}}
  \caption{
      \emph{Left}: On extremely long token sequences, the p-values on unwatermarked data increase as the real expected green ratio $\gamma'=242/971$ is slightly below $\gamma=0.25$ used in the test, making the test sound but overly conservative.
      \emph{Right}: Using $\gamma'$ in the test rectifies this, and p-values behave as expected.}
  \label{fig:fpr_extra}
  \vspace{-1em}
\end{figure}  

%% file: tables/appendix/gentime.tex
\begin{table}[t]
    \centering
    
    \caption{TPR at 1\% FPR of different generation-time baselines. As these methods are not applicable to autoregressive image generation models, we show results on the \texttt{stabilityai/stable-diffusion-2-1-base} diffusion model. The transformations and attacks are the same as in~\cref{tab:baselines}}
    \newcommand{\fivecol}[1]{\multicolumn{5}{c}{#1}}  
 
    \definecolor{badred}{rgb}{0.8,0.2,0.2}

    \resizebox{0.75 \linewidth}{!}{%
    \begingroup 
    \setlength{\tabcolsep}{5pt} %
    \begin{tabular}{l
        x{1}{2}
        x{1}{2}
        x{1}{2}
        x{1}{2}
        x{1}{2}
        }
        \toprule

        & {\textbf{None}} 
        & {\textbf{Valuemetric}} 
        & {\textbf{Geometric}} 
        & {\textbf{\shortstack[c]{Adversarial \\ Purification}}}
        & {\textbf{\shortstack[c]{Neural \\ Compression}}} \\

        \midrule
        \textsc{Stable Signature~\citep{fernandez2023stable}} & 1.00 & 0.71 & 0.71 & \textcolor{badred}{0.39} & \textcolor{badred}{0.54} \\
        \textsc{Tree-Ring~\citep{wen2023tree}} & 1.00 & 0.89 & \textcolor{badred}{0.36} & 0.81 & 0.85 \\
        \textsc{Gaussian Shading~\citep{yang2024gaussian}} & 1.00 & 1.00 & \textcolor{badred}{0.01} & 1.00 & 1.00 \\

        \bottomrule
    \end{tabular}
    \endgroup}
    \label{tab:gentime}
\end{table}

%% file: figures/appendix/ft_rcc_moshi_mimi.tex
\begin{figure}[t]
    \centering
    \includegraphics[width=0.9\textwidth]{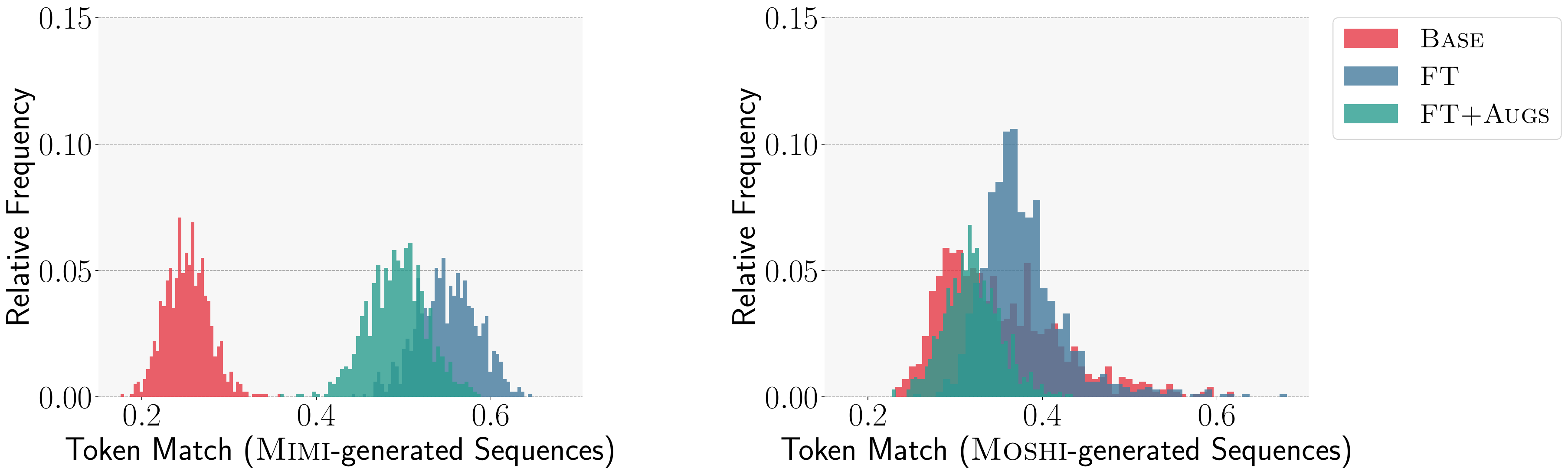}
    \caption{Token match histograms for sequences generated with \mimi (left) and \moshi (right).
    We observe substantial consistency gains for \mimi-generated sequences and minimal change for \moshi-generated ones.}
    \label{fig:ft_rcc_moshi_mimi}
    \vspace{-1em}
\end{figure}  

%% file: figures/appendix/audio_big_boy.tex
\begin{table}[t!]
    \centering
    \caption{TPR at 1\% FPR for the different tokenizer models, watermarking at different values for $\delta$ and different augmentation strengths. Audios are generated with \moshi using the prompts described in \aref{app:expdetails:audio}.}
    \label{tab:full-audio-robustness}
    \resizebox{0.98\textwidth}{!}{
        \input{tables/appendix/audio_full_robustness.tex}
    }
\end{table}

%% file: tables/appendix/audio_full_robustness.tex
\begin{tabular}{l *{3}{x{1}{2}} *{3}{x{1}{2}} *{3}{x{1}{2}}}
\toprule
& \multicolumn{3}{c}{$\delta=0.5$} & \multicolumn{3}{c}{$\delta=2.0$} & \multicolumn{3}{c}{$\delta=4.0$} \\
\cmidrule(lr){2-4} \cmidrule(lr){5-7} \cmidrule(lr){8-10}
Transformation & \base & \ft & \shortstack[c]{\textsc{FT+}\\\textsc{Augs}} & \base & \ft & \shortstack[c]{\textsc{FT+}\\\textsc{Augs}} & \base & \ft & \shortstack[c]{\textsc{FT+}\\\textsc{Augs}} \\
\midrule
\hspace{-4pt}\textit{Identity} \\
Identity & 0.31 & 0.57 & 0.46 & 0.98 & 0.99 & 0.99 & 0.99 & 0.99 & 0.99 \\
\hspace{-4pt}\textit{Time-frequency} \\
Speed 0.75 & 0.06 & 0.04 & 0.05 & 0.09 & 0.03 & 0.08 & 0.19 & 0.07 & 0.17 \\
Speed 0.9 & 0.07 & 0.07 & 0.08 & 0.27 & 0.20 & 0.30 & 0.63 & 0.49 & 0.70 \\
Speed 1.1 & 0.04 & 0.03 & 0.04 & 0.21 & 0.06 & 0.20 & 0.59 & 0.14 & 0.55 \\
Speed 1.25 & 0.02 & 0.01 & 0.02 & 0.09 & 0.02 & 0.07 & 0.26 & 0.04 & 0.19 \\
Crop 0.5 & 0.06 & 0.04 & 0.06 & 0.24 & 0.15 & 0.23 & 0.49 & 0.28 & 0.46 \\
Crop 0.7 & 0.06 & 0.05 & 0.07 & 0.30 & 0.19 & 0.33 & 0.59 & 0.35 & 0.54 \\
Crop 0.9 & 0.07 & 0.05 & 0.08 & 0.37 & 0.21 & 0.36 & 0.69 & 0.40 & 0.63 \\
Shift 10.0 & 0.09 & 0.09 & 0.14 & 0.56 & 0.33 & 0.80 & 0.91 & 0.68 & 0.97 \\
Shift 20.0 & 0.06 & 0.04 & 0.03 & 0.26 & 0.12 & 0.11 & 0.65 & 0.32 & 0.30 \\
Shift 40.0 & 0.06 & 0.08 & 0.03 & 0.17 & 0.24 & 0.12 & 0.48 & 0.64 & 0.39 \\
\hspace{-4pt}\textit{Valuemetric} \\
Bandpass (1000, 8000) & 0.03 & 0.04 & 0.16 & 0.10 & 0.15 & 0.45 & 0.28 & 0.38 & 0.85 \\
Bandpass (300, 3000) & 0.15 & 0.15 & 0.18 & 0.66 & 0.69 & 0.95 & 0.92 & 0.92 & 0.98 \\
Bandpass (500, 5000) & 0.10 & 0.17 & 0.18 & 0.45 & 0.53 & 0.94 & 0.83 & 0.85 & 0.98 \\
Boost 50 & 0.37 & 0.27 & 0.22 & 0.98 & 0.98 & 0.97 & 0.98 & 0.99 & 0.98 \\
Boost 90 & 0.21 & 0.19 & 0.14 & 0.96 & 0.93 & 0.88 & 0.98 & 0.98 & 0.98 \\
Duck 50 & 0.14 & 0.15 & 0.12 & 0.81 & 0.84 & 0.81 & 0.98 & 0.98 & 0.97 \\
Duck 90 & 0.11 & 0.09 & 0.14 & 0.48 & 0.30 & 0.58 & 0.83 & 0.63 & 0.90 \\
Echo (0.1, 0.2) & 0.12 & 0.20 & 0.13 & 0.87 & 0.97 & 0.93 & 0.98 & 0.99 & 0.98 \\
Echo (0.3, 0.5) & 0.05 & 0.09 & 0.07 & 0.55 & 0.79 & 0.60 & 0.89 & 0.97 & 0.93 \\
Echo (0.5, 0.7) & 0.04 & 0.05 & 0.04 & 0.38 & 0.56 & 0.40 & 0.77 & 0.91 & 0.79 \\
Highpass 100 & 0.27 & 0.49 & 0.40 & 0.98 & 0.99 & 0.99 & 0.98 & 0.99 & 0.99 \\
Highpass 1000 & 0.02 & 0.07 & 0.11 & 0.09 & 0.44 & 0.42 & 0.23 & 0.80 & 0.81 \\
Highpass 500 & 0.04 & 0.16 & 0.19 & 0.32 & 0.81 & 0.95 & 0.80 & 0.98 & 0.98 \\
Lowpass 1000 & 0.00 & 0.01 & 0.01 & 0.07 & 0.07 & 0.06 & 0.27 & 0.23 & 0.22 \\
Lowpass 3000 & 0.33 & 0.34 & 0.39 & 0.97 & 0.98 & 0.98 & 0.99 & 0.98 & 0.99 \\
Lowpass 8000 & 0.31 & 0.40 & 0.46 & 0.98 & 0.99 & 0.99 & 0.99 & 0.99 & 0.99 \\
MP3 128 & 0.28 & 0.48 & 0.44 & 0.97 & 0.99 & 0.98 & 0.99 & 0.99 & 0.99 \\
MP3 16 & 0.30 & 0.32 & 0.25 & 0.97 & 0.98 & 0.98 & 0.99 & 0.99 & 0.99 \\
MP3 64 & 0.27 & 0.43 & 0.43 & 0.97 & 0.99 & 0.99 & 0.98 & 0.99 & 0.99 \\
Noise 0.001 & 0.24 & 0.04 & 0.47 & 0.97 & 0.38 & 0.99 & 0.98 & 0.72 & 0.99 \\
Noise 0.01 & 0.03 & 0.01 & 0.31 & 0.50 & 0.01 & 0.98 & 0.86 & 0.01 & 0.98 \\
Noise 0.05 & 0.05 & 0.00 & 0.04 & 0.17 & 0.00 & 0.29 & 0.39 & 0.00 & 0.63 \\
Pink 0.01 & 0.24 & 0.49 & 0.43 & 0.97 & 0.98 & 0.99 & 0.98 & 0.98 & 0.99 \\
Pink 0.05 & 0.17 & 0.18 & 0.32 & 0.97 & 0.97 & 0.98 & 0.98 & 0.98 & 0.98 \\
Pink 0.1 & 0.07 & 0.06 & 0.14 & 0.93 & 0.80 & 0.96 & 0.97 & 0.96 & 0.97 \\
Smooth 0.001 & 0.06 & 0.04 & 0.10 & 0.44 & 0.24 & 0.79 & 0.83 & 0.52 & 0.96 \\
Smooth 0.005 & 0.01 & 0.00 & 0.10 & 0.02 & 0.01 & 0.58 & 0.07 & 0.01 & 0.90 \\
Smooth 0.01 & 0.01 & 0.01 & 0.07 & 0.02 & 0.02 & 0.38 & 0.04 & 0.03 & 0.76 \\
UpDown Res. 24000 & 0.31 & 0.57 & 0.46 & 0.98 & 0.99 & 0.99 & 0.99 & 0.99 & 0.99 \\
UpDown Res. 36000 & 0.30 & 0.52 & 0.46 & 0.98 & 0.99 & 0.99 & 0.99 & 1.00 & 0.99 \\
UpDown Res. 48000 & 0.31 & 0.52 & 0.46 & 0.98 & 0.99 & 0.99 & 0.99 & 1.00 & 0.99 \\
\hspace{-4pt}\textit{Neural Compression} \\
DAC & 0.24 & 0.17 & 0.38 & 0.97 & 0.96 & 0.99 & 0.98 & 0.98 & 0.99 \\
EnCodec & 0.07 & 0.06 & 0.09 & 0.62 & 0.55 & 0.69 & 0.91 & 0.91 & 0.94 \\
\bottomrule
\end{tabular}

%% file: sections/app-qualitative.tex
\section{Qualitative Examples} \label{app:qualitative}

In~\cref{fig:qualitative} (\taming), \cref{fig:qualitative_cham} (\cham), and \cref{fig:qualitative_rar} (\rarxl) we show qualitative examples of images generated with our watermark and images post-hoc watermarked by the baselines previously evaluated in~\cref{ssec:experiments:robustness}.
We notice no observable degradation in quality, which is also confirmed by FID measurements.
Interestingly, \rarxl shows larger pixel-wise differences in decoder behavior due to finetuning (confirming PSNR results in~\aref{app:moreresults:decoderquality}), but the finetuned decoders are still able to generate images of good quality.
We remark that we use the same finetuning parameters for all models, which may not be optimal. 
Tuning the hyperparameters specifically for \rarxl may thus be able to achieve similar watermark strength and robustness with smaller decoder changes.

\input{figures/appendix/qualitative.tex}  
\clearpage
\input{figures/appendix/qualitative_cham.tex}
\clearpage 
\input{figures/appendix/qualitative_rar.tex}

%% file: figures/appendix/qualitative.tex
\newcommand{\lf}[1]{\includegraphics[width=0.21\linewidth]{figures/qualitative/#1.jpg}}
\providecommand{\twocol}[1]{\multicolumn{2}{c}{#1}} 
\providecommand{\colspace}{\hspace{0.035\linewidth}}
\providecommand{\dubspace}{\hspace{0.07\linewidth}}

\begin{figure}[h]
  \centering
  \resizebox{\linewidth}{!}{ 
    \begin{tabular}{
cc@{\colspace}c@{\dubspace}c@{\colspace}c@{\dubspace}c@{\colspace}c
    }
      \toprule 
       & \twocol{Border Collie} & \twocol{Goldfish} & \twocol{Lakeside} \\
      \midrule
      \raisebox{0.1\linewidth}{\base} 
      & \lf{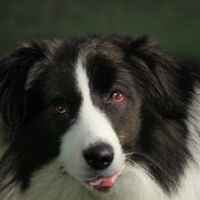} 
      & \lf{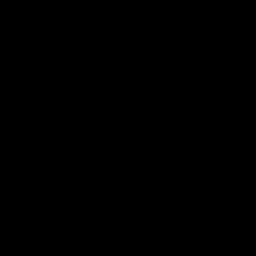} 
      & \lf{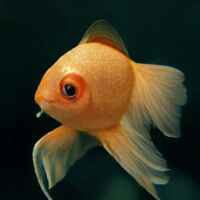} 
      & \lf{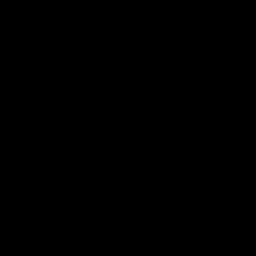} 
      & \lf{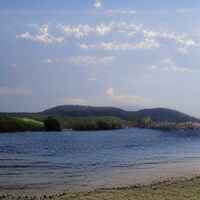} 
      & \lf{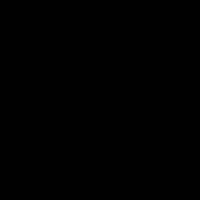} \\
      \raisebox{0.1\linewidth}{\ft} 
      & \lf{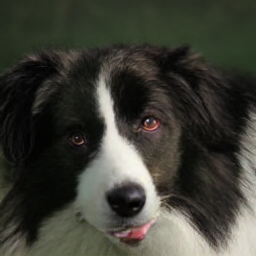} 
      & \lf{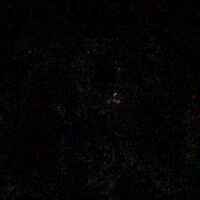} 
      & \lf{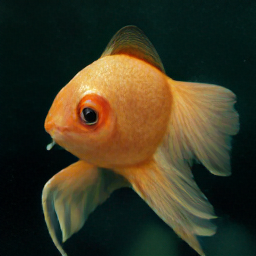} 
      & \lf{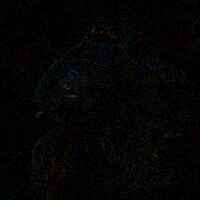} 
      & \lf{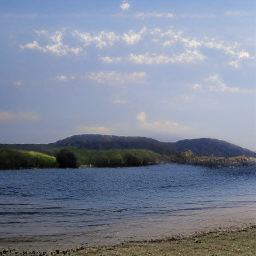} 
      & \lf{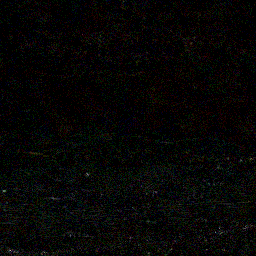} \\
      \raisebox{0.1\linewidth}{\ftaug} 
      & \lf{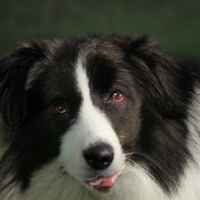} 
      & \lf{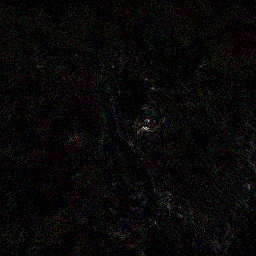} 
      & \lf{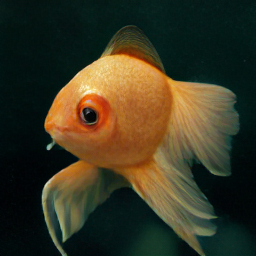} 
      & \lf{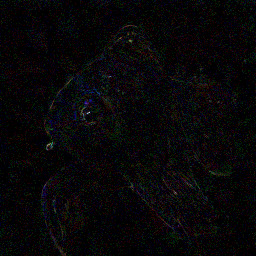} 
      & \lf{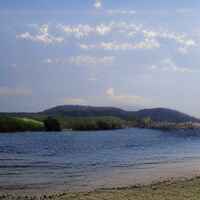} 
      & \lf{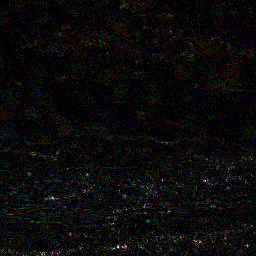} \\
      \raisebox{0.1\linewidth}{\ftaugsync} 
      & \lf{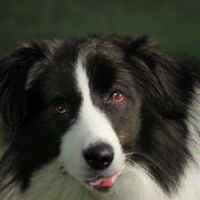} 
      & \lf{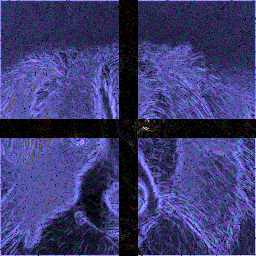} 
      & \lf{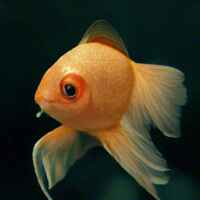} 
      & \lf{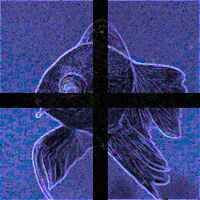} 
      & \lf{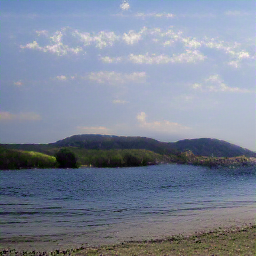} 
      & \lf{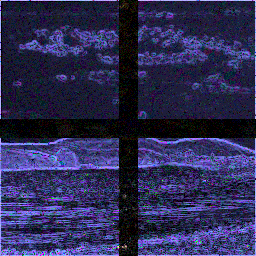} \\
      \midrule
      \raisebox{0.1\linewidth}{Original} 
      & \lf{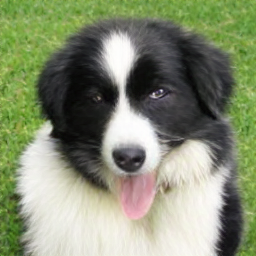} 
      & \lf{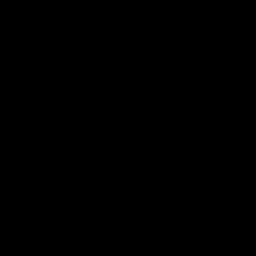} 
      & \lf{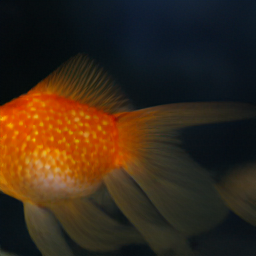} 
      & \lf{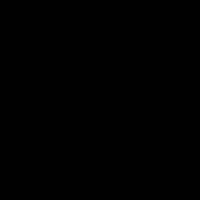} 
      & \lf{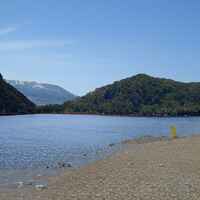} 
      & \lf{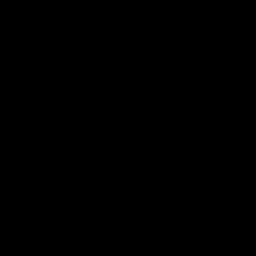} \\
      \raisebox{0.1\linewidth}{\textsc{CIN}} 
      & \lf{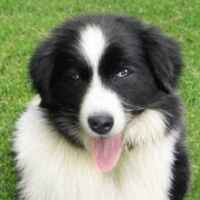} 
      & \lf{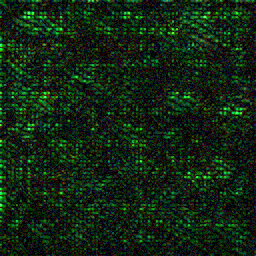} 
      & \lf{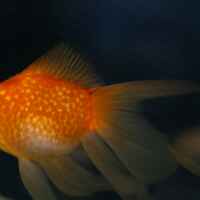} 
      & \lf{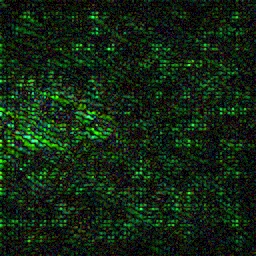} 
      & \lf{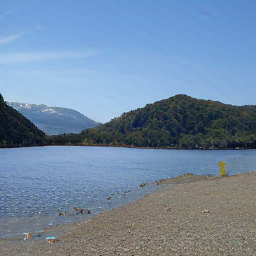} 
      & \lf{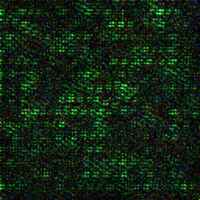} \\
      \raisebox{0.1\linewidth}{\textsc{MBRS}} 
      & \lf{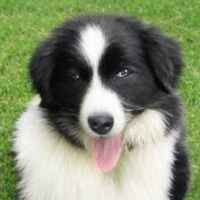} 
      & \lf{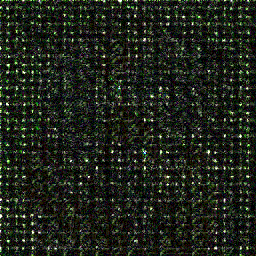} 
      & \lf{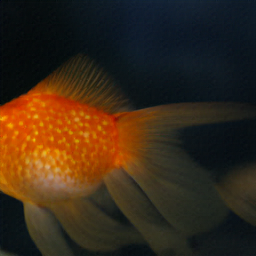} 
      & \lf{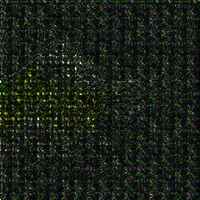} 
      & \lf{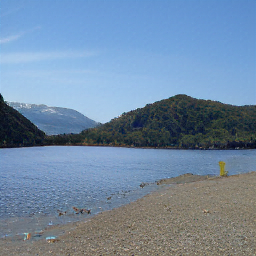} 
      & \lf{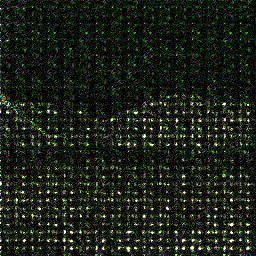} \\
      \raisebox{0.1\linewidth}{\textsc{Trustmark}} 
      & \lf{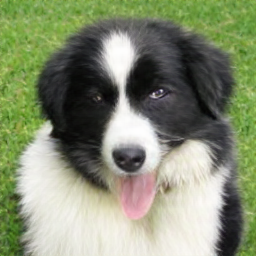} 
      & \lf{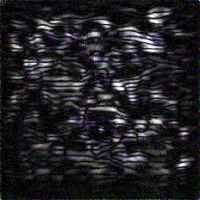} 
      & \lf{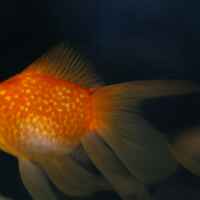} 
      & \lf{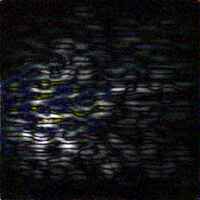} 
      & \lf{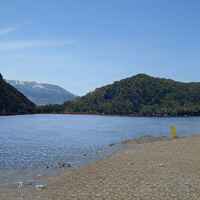} 
      & \lf{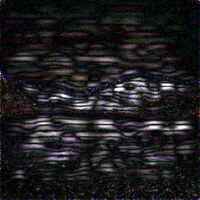} \\
      \raisebox{0.1\linewidth}{\textsc{WAM}} 
      & \lf{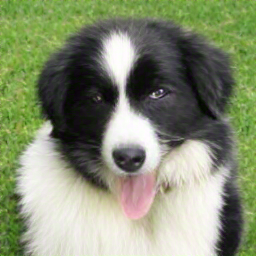} 
      & \lf{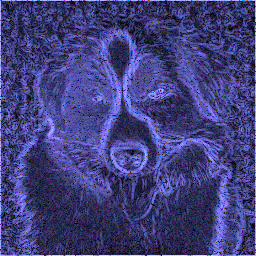} 
      & \lf{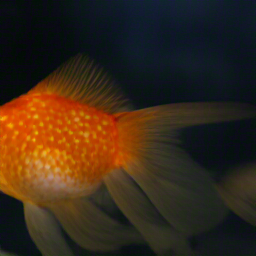} 
      & \lf{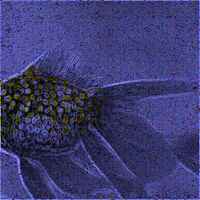} 
      & \lf{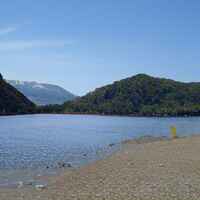} 
      & \lf{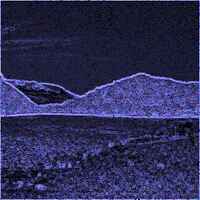} \\
    \bottomrule
    \end{tabular}
  }
  \caption{Qualitative results on \taming with samples from $3$ of the ImageNet-1k classes. The left column shows the images and the right column the diffs. For variants of our watermark (\emph{top}) the diff is computed w.r.t. the \base decoder (as there is no notion of an original unwatermarked image). For post-hoc baselines (\emph{bottom}) the diff is computed w.r.t. the original image. All diffs are displayed with the same postprocessing applied for visibility, namely $\text{clip}(|a-b| \cdot 30, 0, 255)$, where $a$ and $b$ are pixel values of the two images in range $[0,255]$.}
  \label{fig:qualitative}
\end{figure} 

%% file: figures/appendix/qualitative_cham.tex
\newcommand{\lfc}[1]{\includegraphics[width=0.21\linewidth]{figures/qualitative_cham/#1.jpg}}
\providecommand{\twocol}[1]{\multicolumn{2}{c}{#1}} 
\providecommand{\colspace}{\hspace{0.035\linewidth}}
\providecommand{\dubspace}{\hspace{0.07\linewidth}}

\begin{figure}[t]
  \centering
  \resizebox{\linewidth}{!}{%
    \begin{tabular}{
      cc@{\colspace}c@{\dubspace}c@{\colspace}c@{\dubspace}c@{\colspace}c
    } 
      \toprule
       & \twocol{\shortstack[c]{This large pizza has a lot\\ of cheese and tomato sauce}} 
       & \twocol{\shortstack[c]{A vase filled with \\yellow and white flowers}} 
       & \twocol{\shortstack[c]{The elephant is standing alone \\ outside by the trees}} \\
      \midrule
      \raisebox{0.1\linewidth}{\base} 
      & \lfc{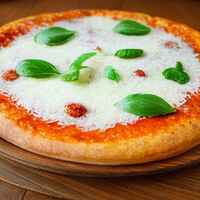} 
      & \lfc{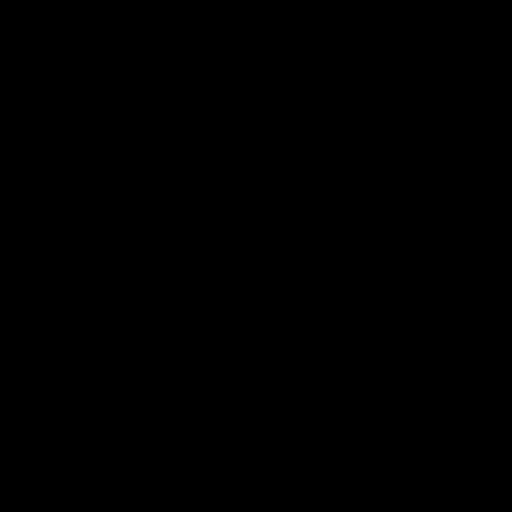} 
      & \lfc{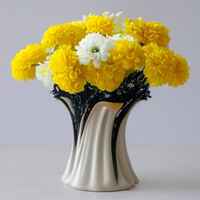} 
      & \lfc{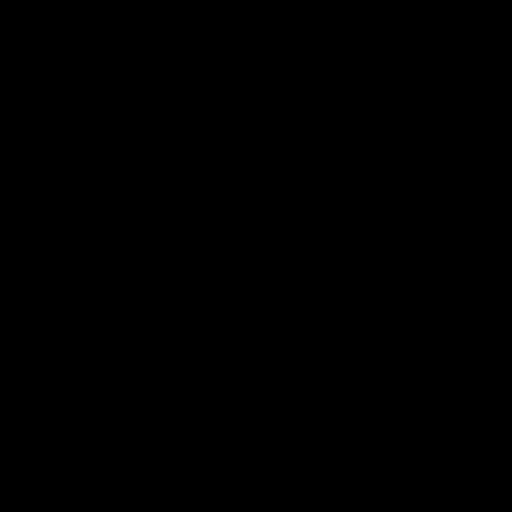} 
      & \lfc{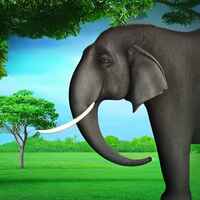} 
      & \lfc{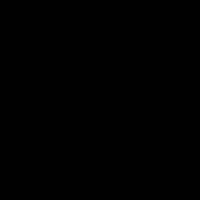} \\
      \raisebox{0.1\linewidth}{\ft} 
      & \lfc{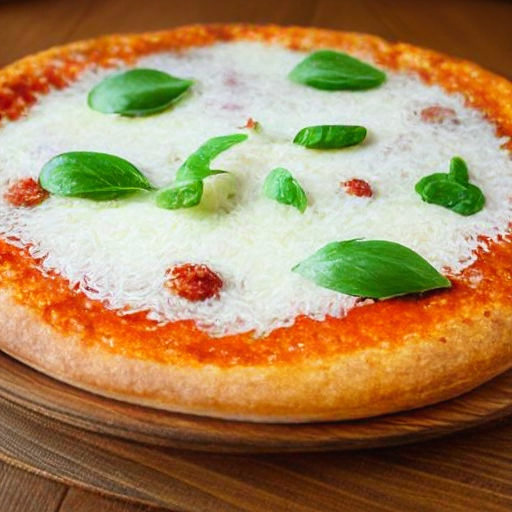} 
      & \lfc{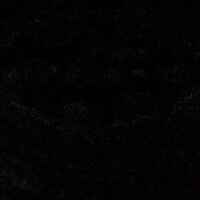} 
      & \lfc{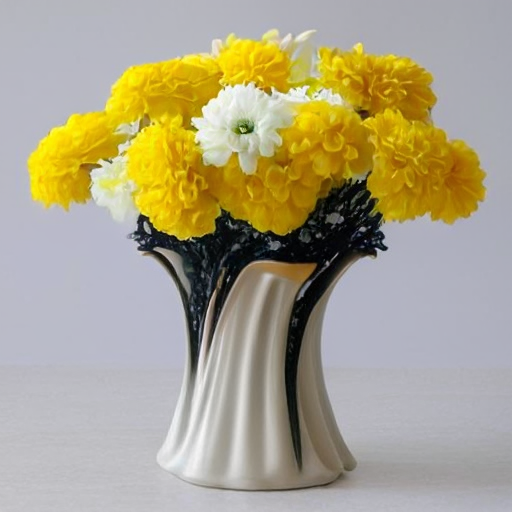} 
      & \lfc{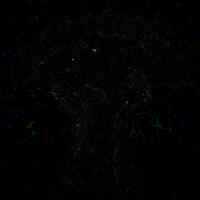} 
      & \lfc{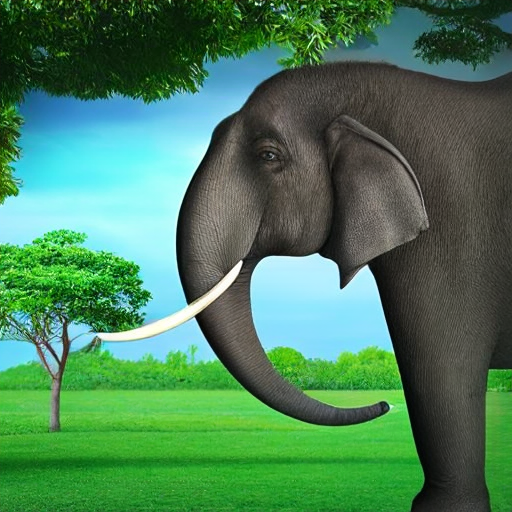} 
      & \lfc{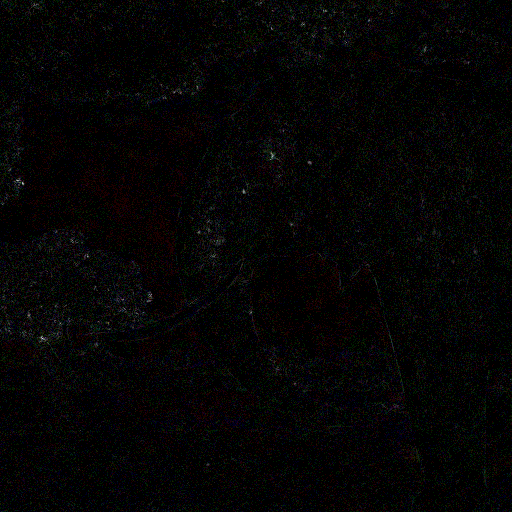} \\
      \raisebox{0.1\linewidth}{\ftaug} 
      & \lfc{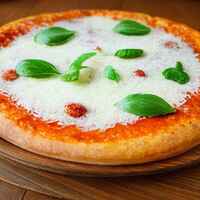} 
      & \lfc{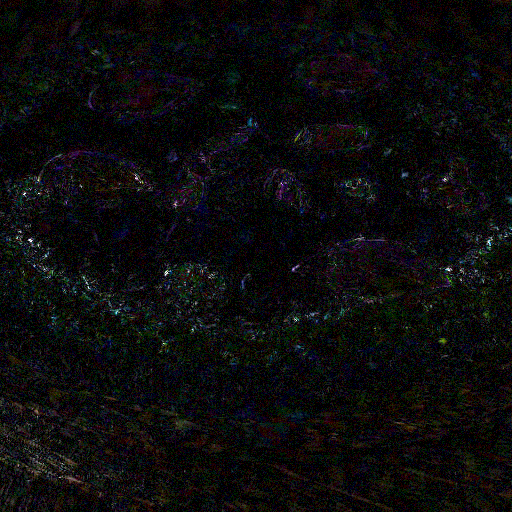} 
      & \lfc{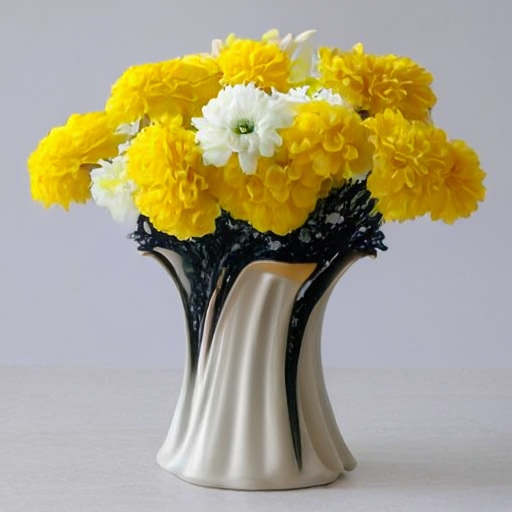} 
      & \lfc{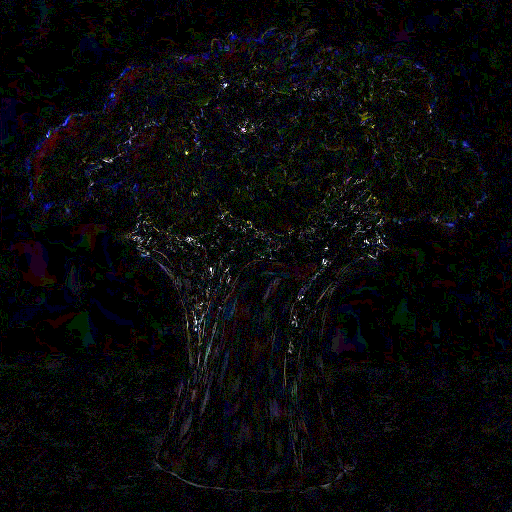} 
      & \lfc{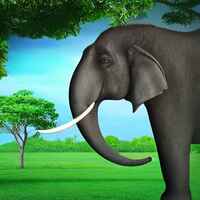} 
      & \lfc{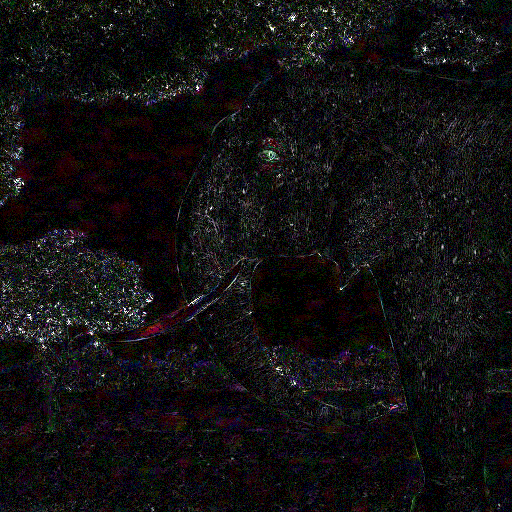} \\
      \raisebox{0.1\linewidth}{\ftaugsync} 
      & \lfc{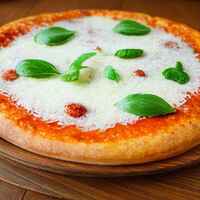} 
      & \lfc{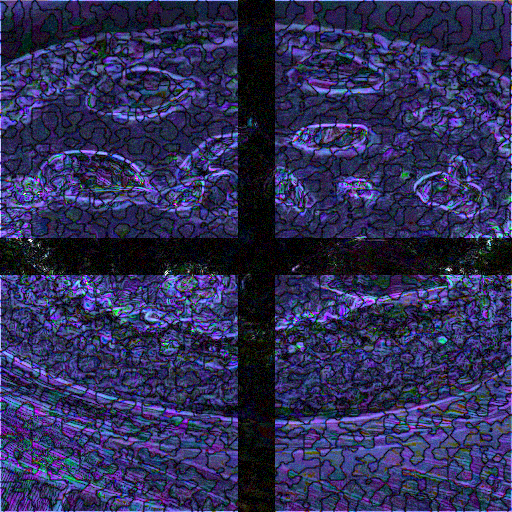} 
      & \lfc{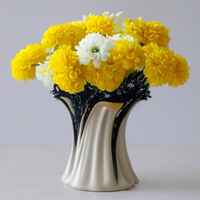} 
      & \lfc{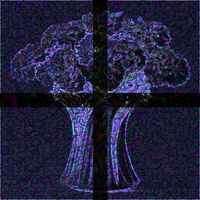} 
      & \lfc{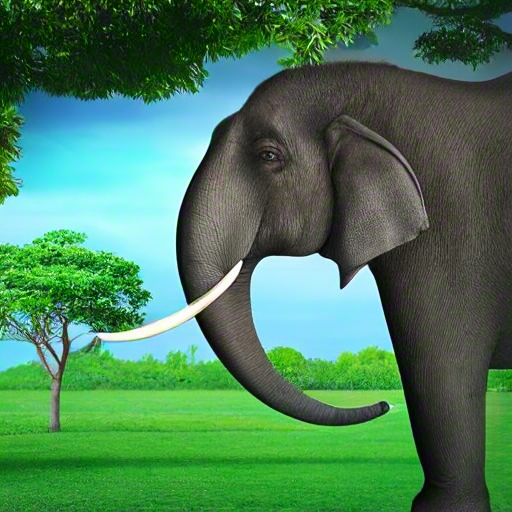} 
      & \lfc{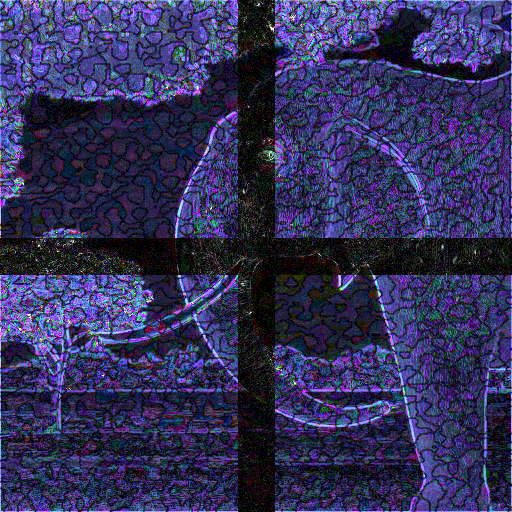} \\
      \midrule
      \raisebox{0.1\linewidth}{Original} 
      & \lfc{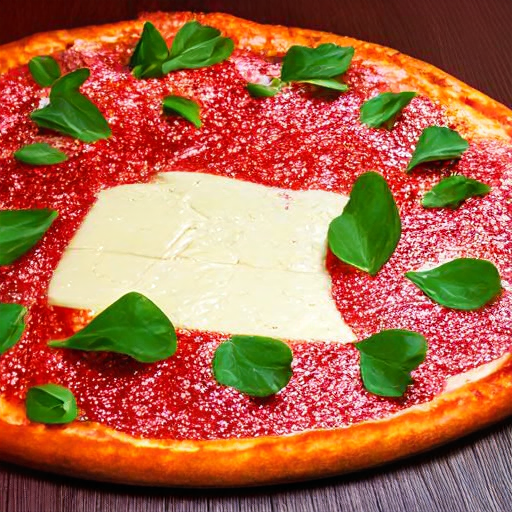} 
      & \lfc{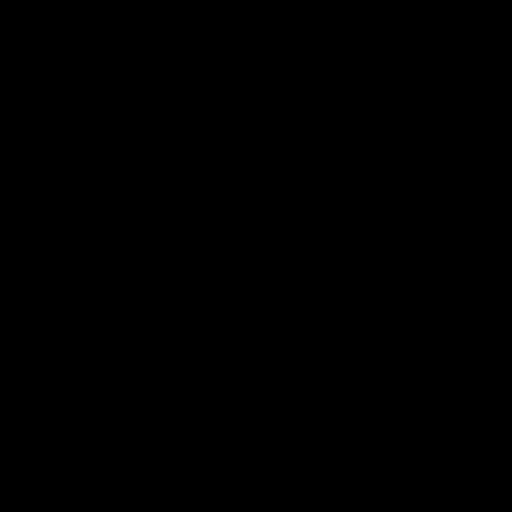} 
      & \lfc{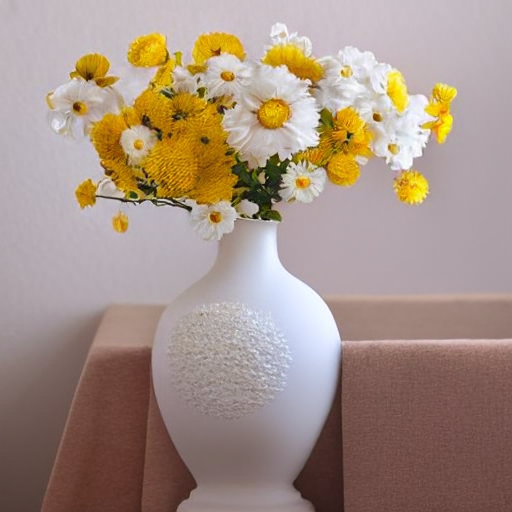} 
      & \lfc{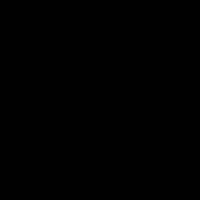} 
      & \lfc{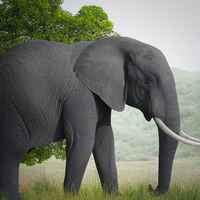} 
      & \lfc{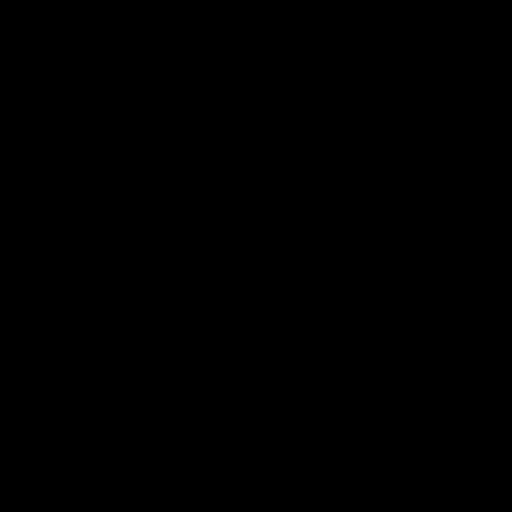} \\
      \raisebox{0.1\linewidth}{\textsc{CIN}} 
      & \lfc{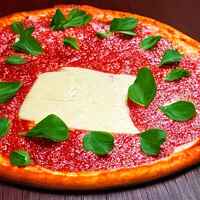} 
      & \lfc{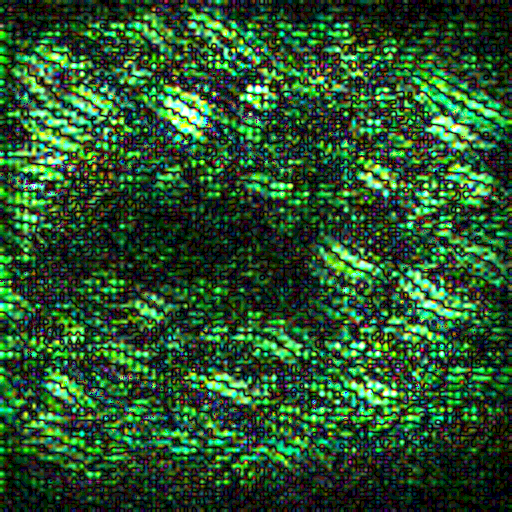} 
      & \lfc{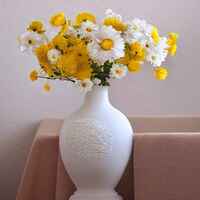} 
      & \lfc{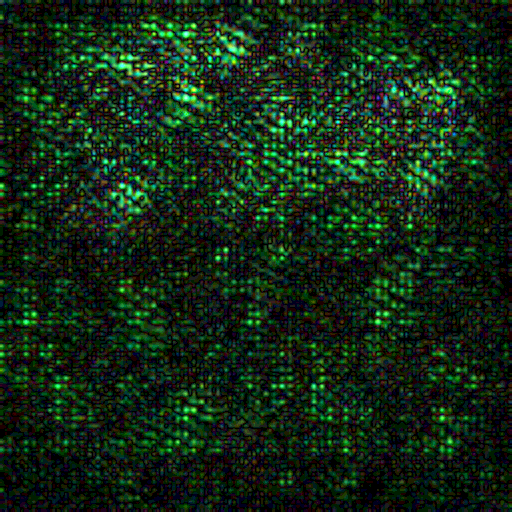} 
      & \lfc{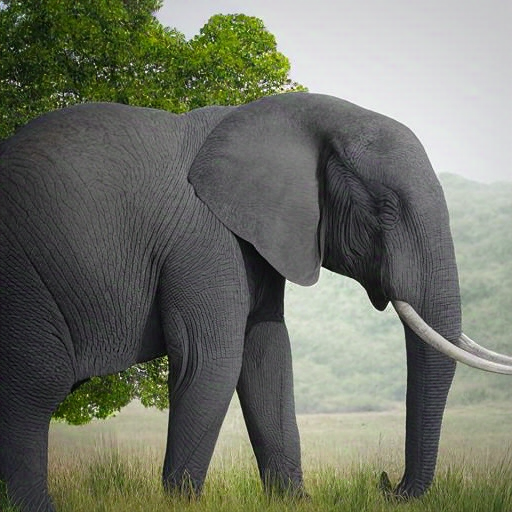} 
      & \lfc{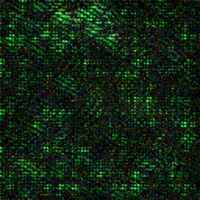} \\
      \raisebox{0.1\linewidth}{\textsc{MBRS}} 
      & \lfc{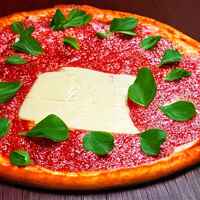} 
      & \lfc{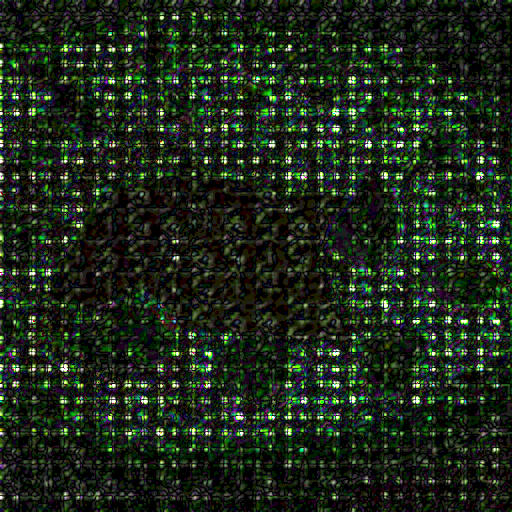} 
      & \lfc{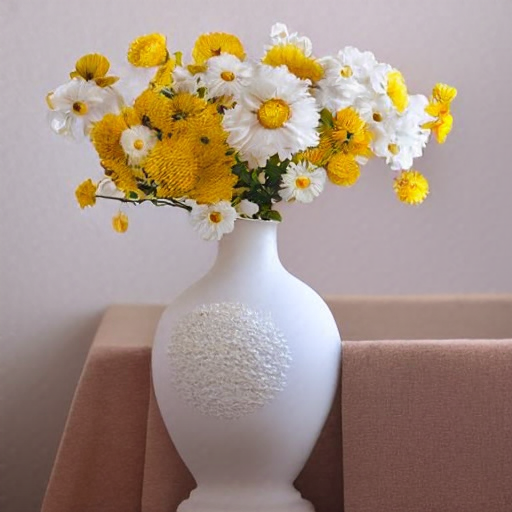} 
      & \lfc{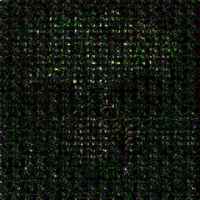} 
      & \lfc{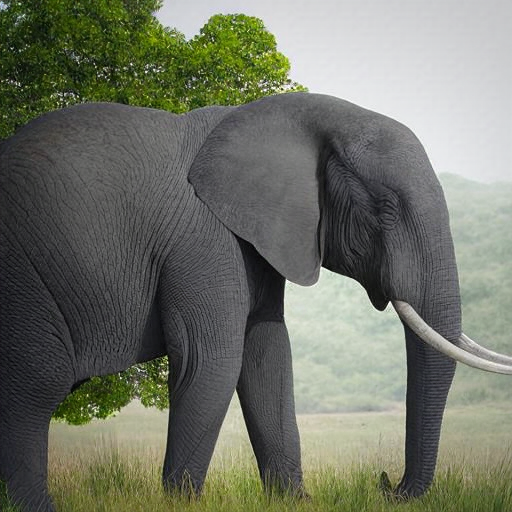} 
      & \lfc{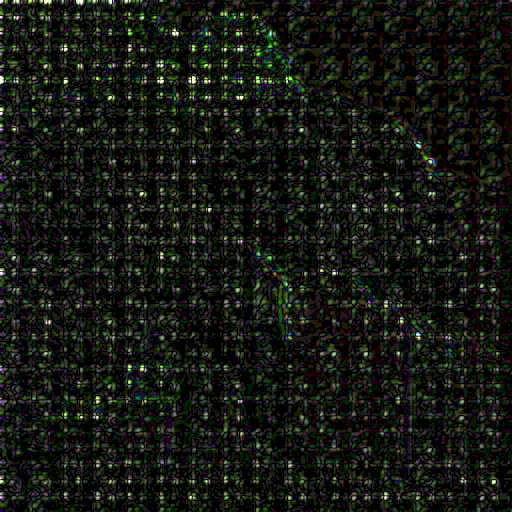} \\
      \raisebox{0.1\linewidth}{\textsc{Trustmark}} 
      & \lfc{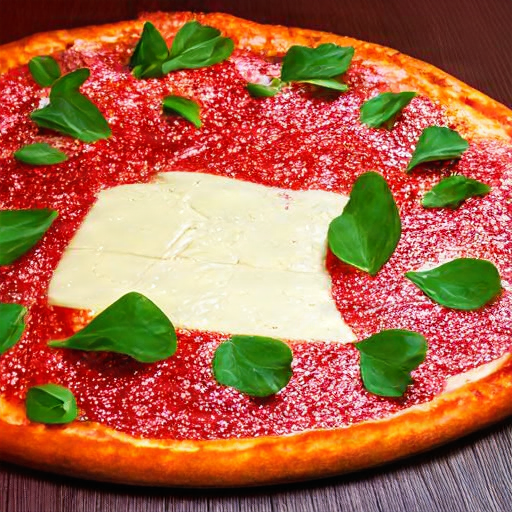} 
      & \lfc{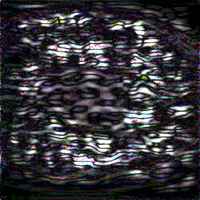} 
      & \lfc{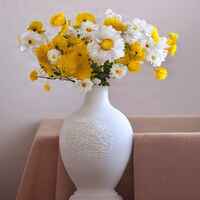} 
      & \lfc{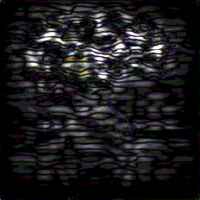} 
      & \lfc{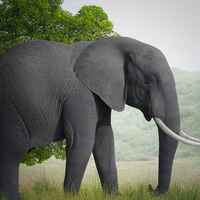} 
      & \lfc{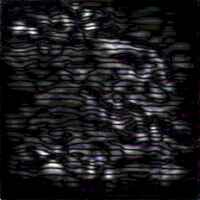} \\
      \raisebox{0.1\linewidth}{\textsc{WAM}} 
      & \lfc{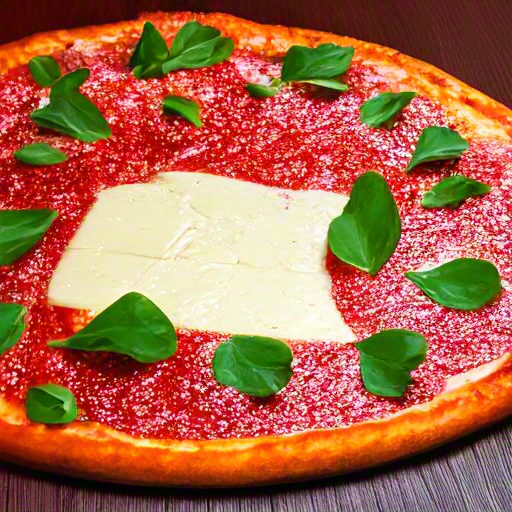} 
      & \lfc{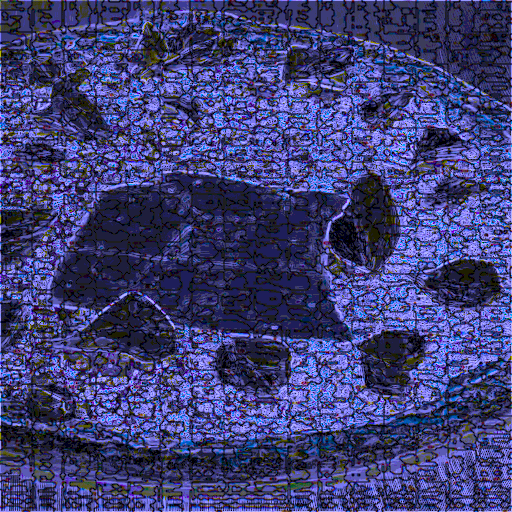} 
      & \lfc{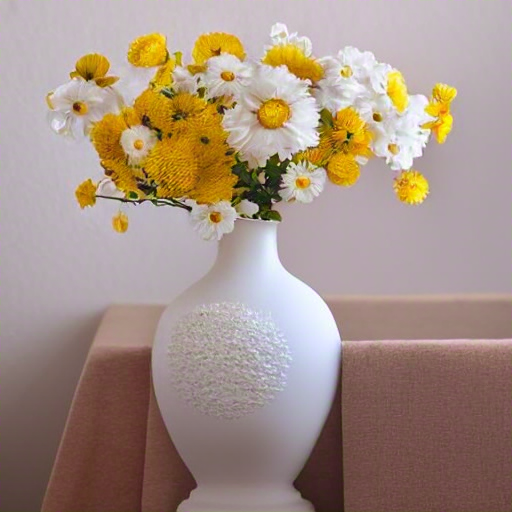} 
      & \lfc{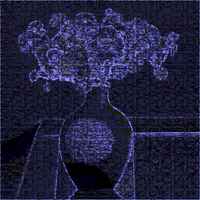} 
      & \lfc{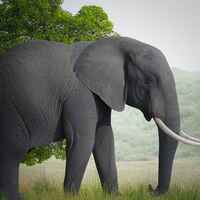} 
      & \lfc{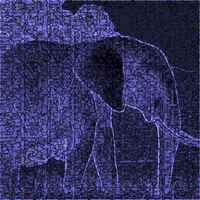} \\
      \bottomrule
    \end{tabular} 
    }
  \caption{Qualitative results on \cham with $3$ COCO validation prompts. The left column shows the images and the right column the diffs. For variants of our watermark (\emph{top}) the diff is computed w.r.t. the \base decoder (as there is no notion of an original unwatermarked image). For post-hoc baselines (\emph{bottom}) the diff is computed w.r.t. the original image. All diffs are displayed with the same postprocessing applied for visibility, namely $\text{clip}(|a-b| \cdot 30, 0, 255)$, where $a$ and $b$ are pixel values of the two images in range $[0,255]$.}
  \label{fig:qualitative_cham}
\end{figure} 

%% file: figures/appendix/qualitative_rar.tex
\newcommand{\lfrar}[1]{\includegraphics[width=0.21\linewidth]{figures/qualitative_rar/#1.jpg}}
\providecommand{\twocol}[1]{\multicolumn{2}{c}{#1}} 
\providecommand{\colspace}{\hspace{0.035\linewidth}}
\providecommand{\dubspace}{\hspace{0.07\linewidth}}

\begin{figure}[h]
  \centering
  \resizebox{0.8\linewidth}{!}{ 
    \begin{tabular}{
cc@{\colspace}c@{\dubspace}c@{\colspace}c@{\dubspace}c@{\colspace}c
    }
      \toprule 
       & \twocol{Broccolli} & \twocol{Ostrich} & \twocol{Lakeside} \\
      \midrule
      \raisebox{0.1\linewidth}{\base} 
      & \lfrar{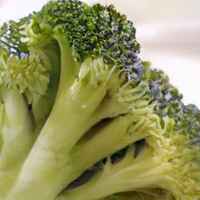} 
      & \lfrar{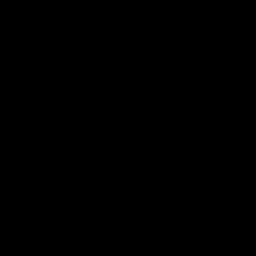} 
      & \lfrar{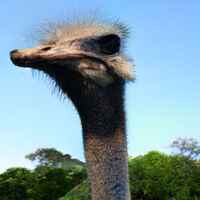} 
      & \lfrar{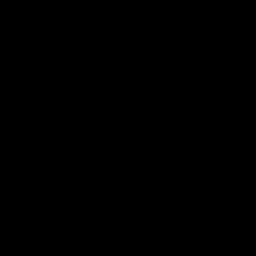} 
      & \lfrar{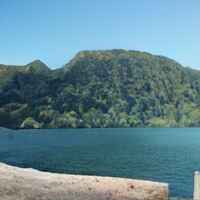} 
      & \lfrar{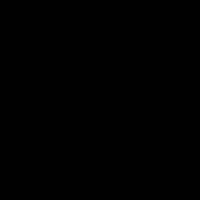} \\
      \raisebox{0.1\linewidth}{\ft} 
      & \lfrar{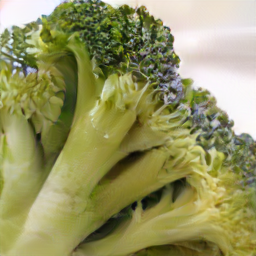} 
      & \lfrar{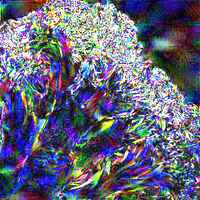} 
      & \lfrar{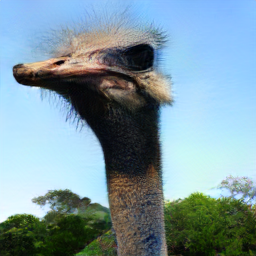} 
      & \lfrar{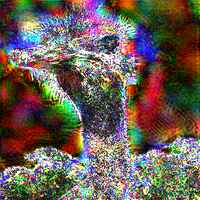} 
      & \lfrar{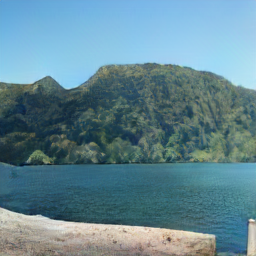} 
      & \lfrar{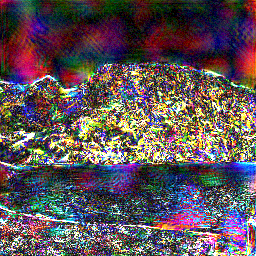} \\
      \raisebox{0.1\linewidth}{\ftaug} 
      & \lfrar{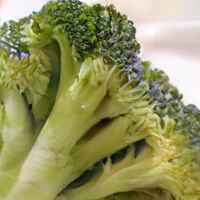} 
      & \lfrar{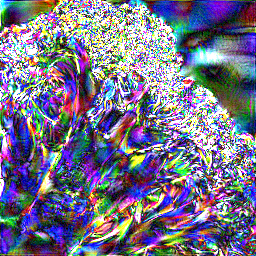} 
      & \lfrar{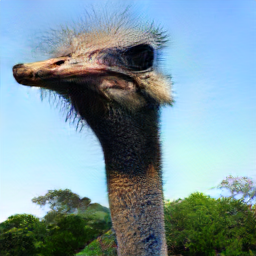} 
      & \lfrar{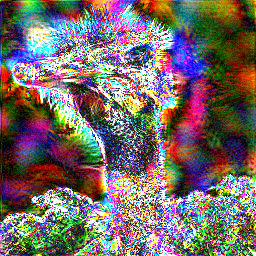} 
      & \lfrar{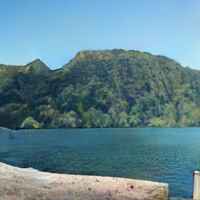} 
      & \lfrar{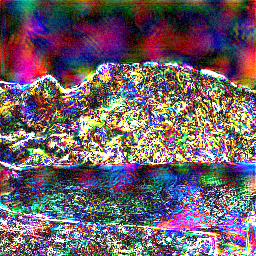} \\
      \raisebox{0.1\linewidth}{\ftaugsync} 
      & \lfrar{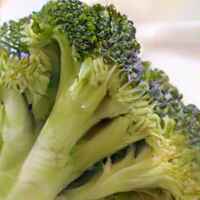} 
      & \lfrar{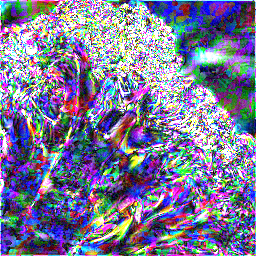} 
      & \lfrar{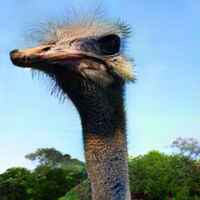} 
      & \lfrar{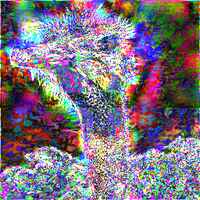} 
      & \lfrar{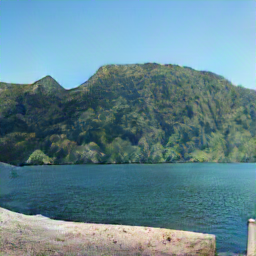} 
      & \lfrar{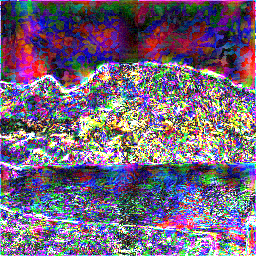} \\
      \midrule
      \raisebox{0.1\linewidth}{Original} 
      & \lfrar{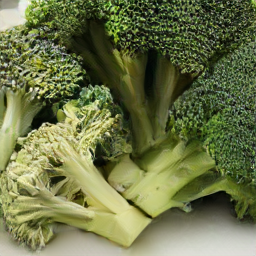} 
      & \lfrar{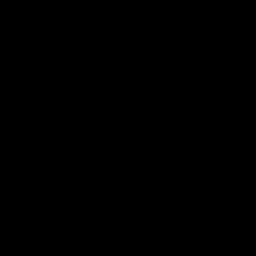} 
      & \lfrar{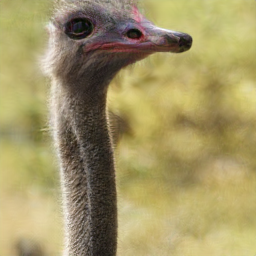} 
      & \lfrar{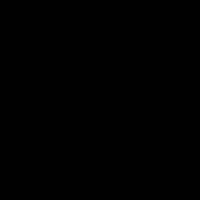} 
      & \lfrar{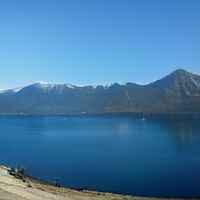} 
      & \lfrar{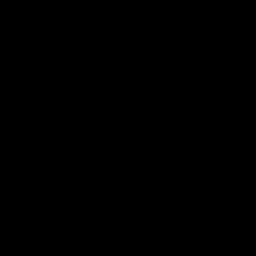} \\
      \raisebox{0.1\linewidth}{\textsc{CIN}} 
      & \lfrar{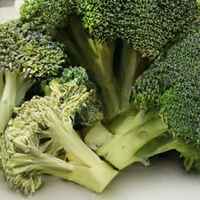} 
      & \lfrar{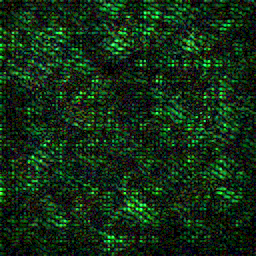} 
      & \lfrar{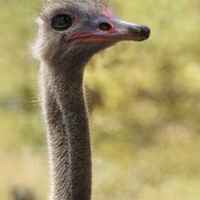} 
      & \lfrar{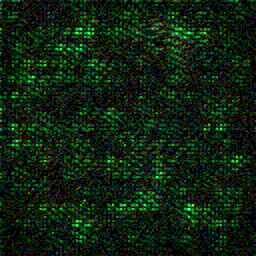} 
      & \lfrar{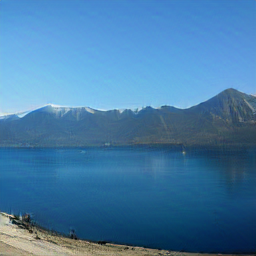} 
      & \lfrar{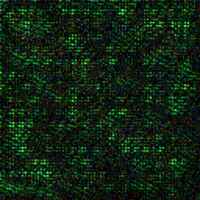} \\
      \raisebox{0.1\linewidth}{\textsc{MBRS}} 
      & \lfrar{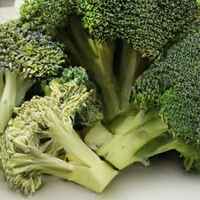} 
      & \lfrar{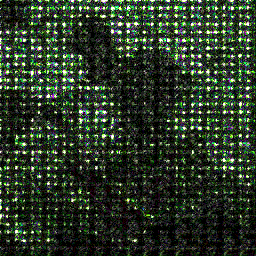} 
      & \lfrar{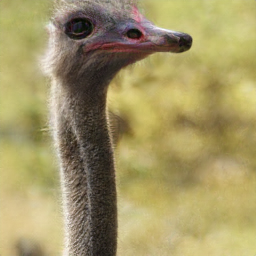} 
      & \lfrar{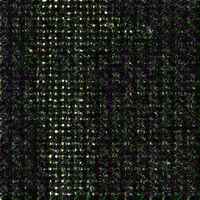} 
      & \lfrar{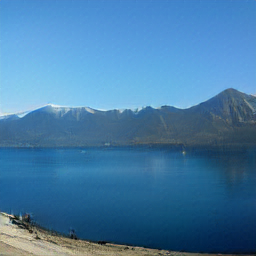} 
      & \lfrar{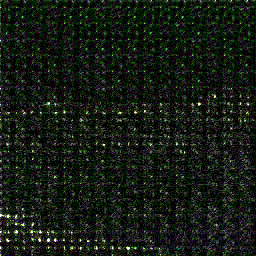} \\
      \raisebox{0.1\linewidth}{\textsc{Trustmark}} 
      & \lfrar{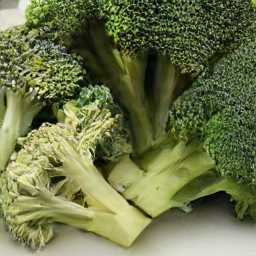} 
      & \lfrar{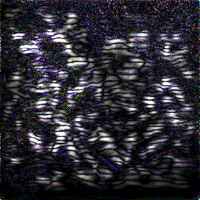} 
      & \lfrar{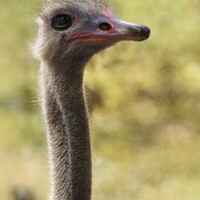} 
      & \lfrar{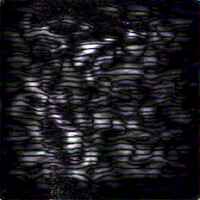} 
      & \lfrar{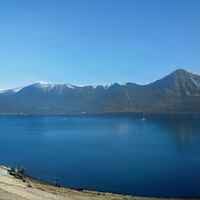} 
      & \lfrar{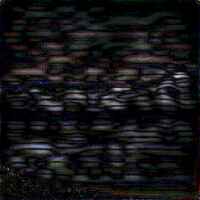} \\
      \raisebox{0.1\linewidth}{\textsc{WAM}} 
      & \lfrar{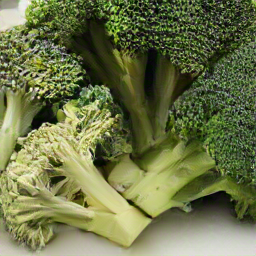} 
      & \lfrar{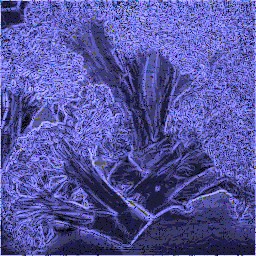} 
      & \lfrar{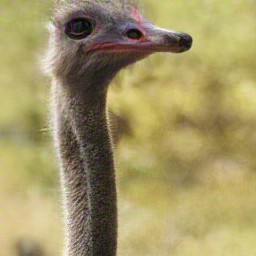} 
      & \lfrar{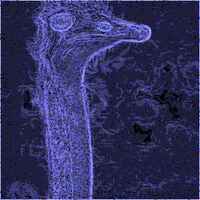} 
      & \lfrar{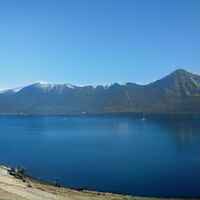} 
      & \lfrar{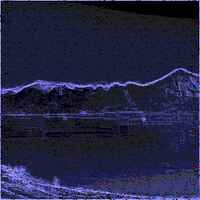} \\
      \midrule
    \bottomrule
    \end{tabular}
  }
  \caption{Qualitative results on \rarxl with samples from $3$ of the ImageNet-1k classes, as in~\cref{fig:qualitative} and \cref{fig:qualitative_cham}.  We observe bigger pixel-wise differences of finetuned decoders for our watermark compared to other models, but no observable degradation in quality, confirmed by FID. More elaborate tuning of the finetuning hyperparameters (we use the same ones for all models) could achieve same results with smaller decoder changes.
  }
  \vspace{-1em}
  \label{fig:qualitative_rar}
\end{figure}

%% file: sections/app-interleaved.tex
\section{On Joint Watermarking of Interleaved Modalities} \label{app:interleaved}
In this section, we extend our discussion on joint watermarking of interleaved modalities from~\cref{ssec:experiments:interleaved}, provide omitted experimental details, and present extended experimental results.

\paragraph{Experimental details.}
We query \cham with $1000$ prompts, each generated from one of ImageNet-1K classes, asking the model to teach the user about the notion represented by the class label and illustrate it.
We use $\gamma=0.25$ and set $h=0$ for images and $h=1$ for text. 
Two example interactions are shown in~\aref{app:interleaved:examples}.
For text corruption we explore percentages in $[0, 60]$.
For Gaussian noise corruption of images, we use $\sigma \in \{0, 0.05, 0.1, 0.15, 0.2, 0.25, 0.3\}$.

\paragraph{Full experimental results.}
Extended results are shown in~\cref{fig:interleaved_full}.
We observe that for high-quality images where the watermark signal is preserved, joint detection is always beneficial, while for highly corrupted images, it almost never is. 
Between these two, joint detection becomes beneficial starting from some text corruption level.
The intuitive understanding is that joint detection that integrates a \emph{better quality} signal is always beneficial.
Perhaps unexpectedly, it can be also beneficial when signal is of slightly lower quality if it sufficiently increases the number of tokens.
This is important as a single image consists of a large number of tokens, in particular $1024$ for \cham, while the average text length in our experiment is ${\approx}227$.

\input{figures/appendix/interleaved_full.tex}
\paragraph{When is joint detection beneficial?}
First, to elaborate on our claim from the main paper that \emph{scoring more equally watermarked tokens improves power}, we plot the p-value of the watermark detection as a function of token length for a \emph{fixed green ratio $S/(T-h)$} (\cref{fig:interleaved_extra}, left).
For each line, increasing the number of tokens can rapidly improve the detection p-value. 
To illustrate the point made above that increasing the number of tokens can be beneficial even if the green ratio drops, note the two orange crosses.
Increasing the number of tokens fivefold ($500 \to 2500$) while reducing the green ratio from $28\%$ to $27\%$ has still dropped the p-value from around $0.06$ to almost $0.01$ which for FPR of $1\%$ is the passing threshold. 

\input{figures/appendix/interleaved_extra.tex}
Rounding the average text length in our experiment to $256$, the fivefold increase corresponds exactly to the average case we observed ($256 \to 1024+256$).
We simulate this case (a combination of $256$ text and $1024$ image tokens) while fixing the green ratio for both modalities.
Our results in~\cref{fig:interleaved_extra} (right) show a similar trend as in our above experiment on real data. 
For example, merging text with $28\%$ green tokens with an image with only $26\%$ green tokens still improves the p-value.

These results illustrate the benefits of watermarking modalities jointly.
In contrast, if text and images were watermarked separately, to reject the null hypothesis of \emph{this online article was not generated by my model $\model$} in a statistically sound way, we would need to take the maximum of the p-values of the two individual tests, \ie the improvements seen above would not be possible.

\paragraph{Outlook.}
We hope this investigation opens the door to more studies of cross-modal watermarking.
Other interesting questions include the impact of transformations on the watermark \emph{across modalities}, and the relationship between using more tokens from the same or a different modality.
Intuitively, as vocabularies are disjoint, for long sequences the former may add many duplicate tokens, which do not affect the p-value.
On the other hand, the probability of such duplicates may be higher for some modalities: in the first example in~\aref{app:interleaved:examples}, the image mostly contains white tokens. 

\clearpage 
\subsection{Example Interactions} \label{app:interleaved:examples}
We show two real interactions with \cham from our joint watermark experiment (\cref{ssec:experiments:interleaved}), along with the corresponding watermark detection results.
In the detector view, for both modalities, \boxyGreen{} denotes tokens scored green, \boxyRed{} denotes tokens scored red, and \boxyWhite{} those ignored due to duplication (\eg plain white visual tokens in the first example, we use gray for visibility).
\input{prompts/guitar.tex}
\clearpage
\input{prompts/thatch.tex}

%% file: figures/appendix/interleaved_full.tex
\begin{wrapfigure}[15]{r}{0.38\textwidth}
    \centering
    \vspace{-2em}
		\includegraphics[width=\linewidth]{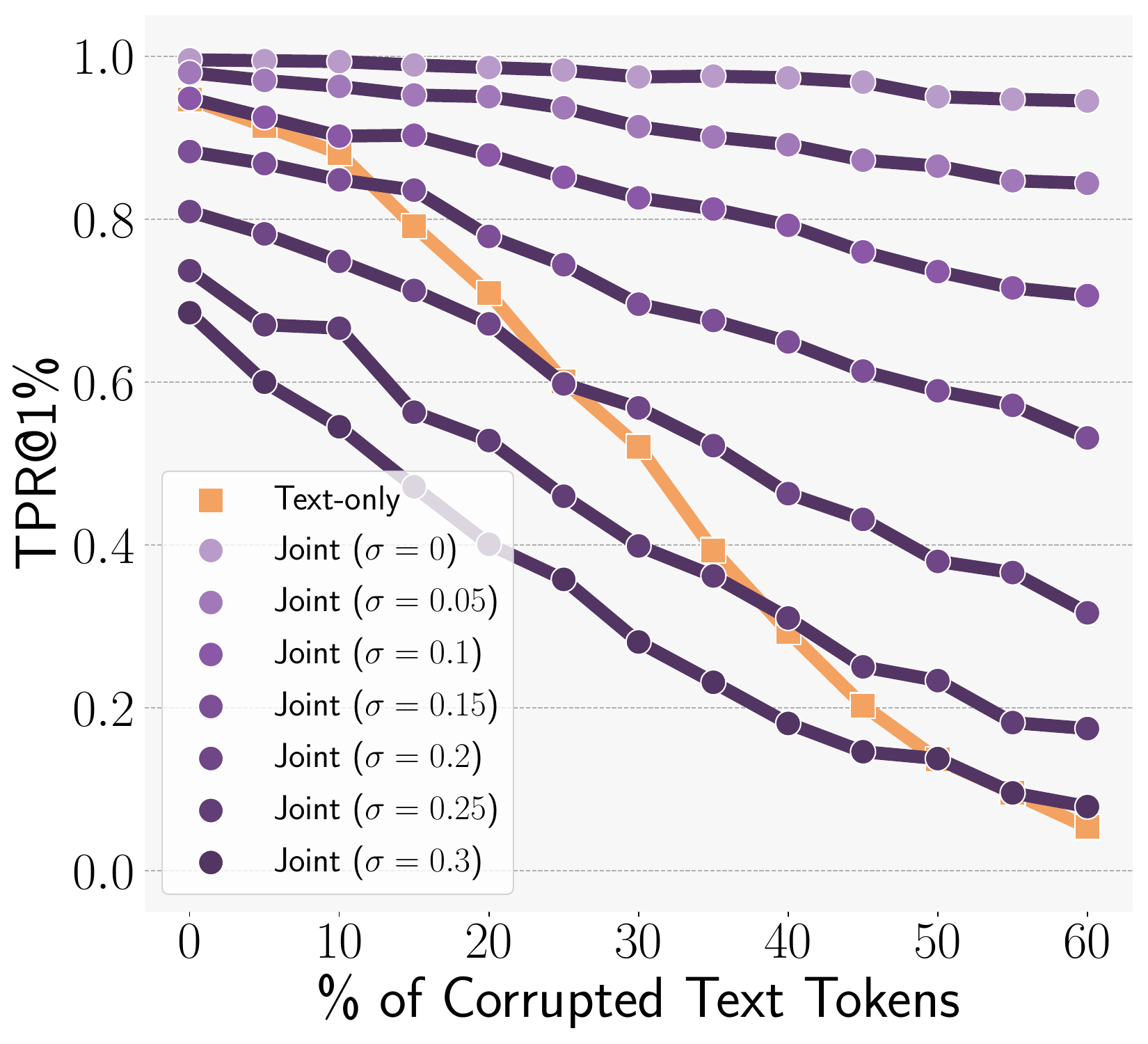}
        \caption{Full results on joint watermark detection, extending~\cref{fig:interleaved}.
        }
  \label{fig:interleaved_full} 
\end{wrapfigure}
    

%% file: figures/appendix/interleaved_extra.tex
\begin{figure}[t]
  \centering
  \resizebox{0.87\textwidth}{!}{
  \includegraphics[width=0.5\textwidth]{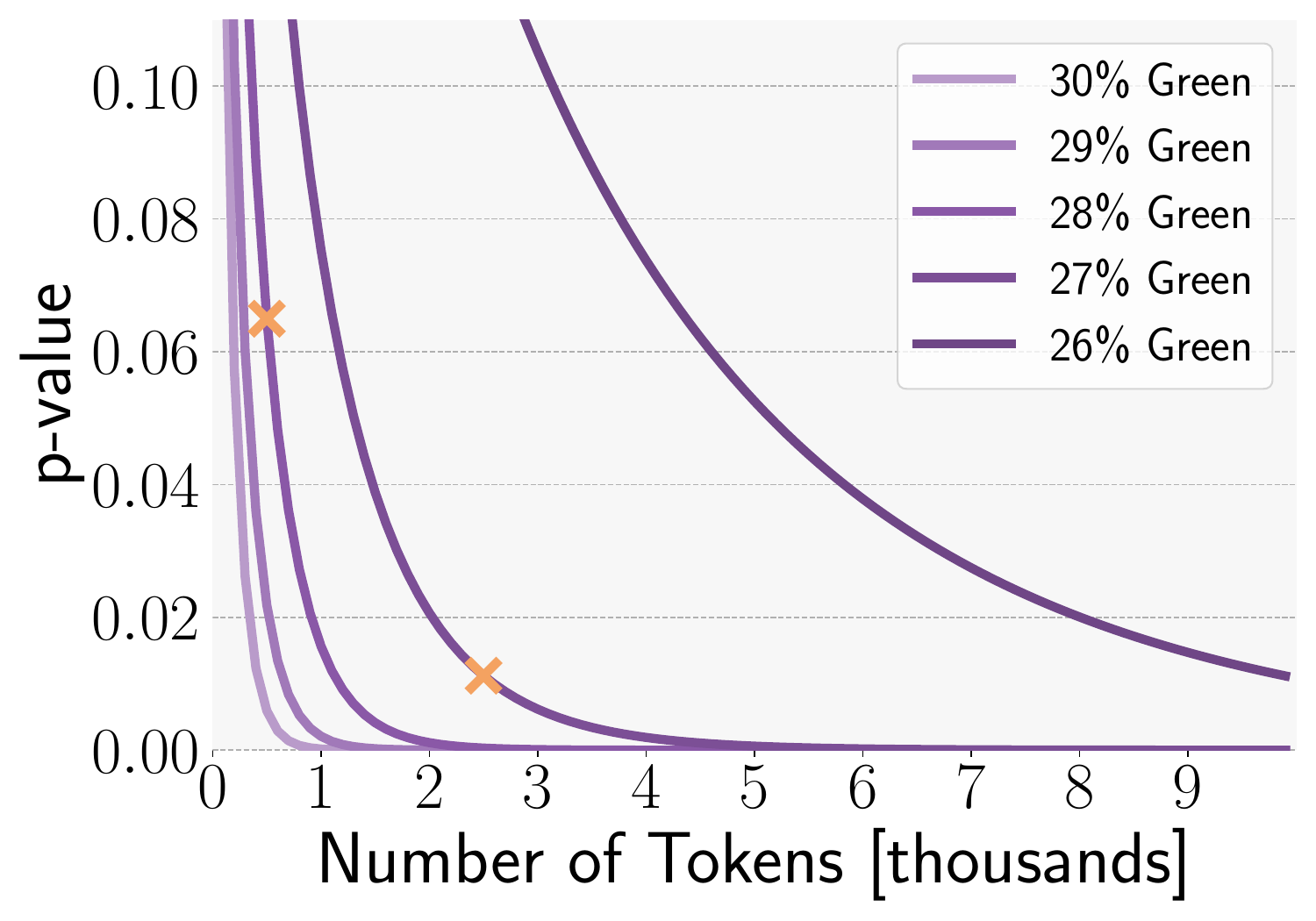}
  \quad
  \includegraphics[width=0.4\textwidth]{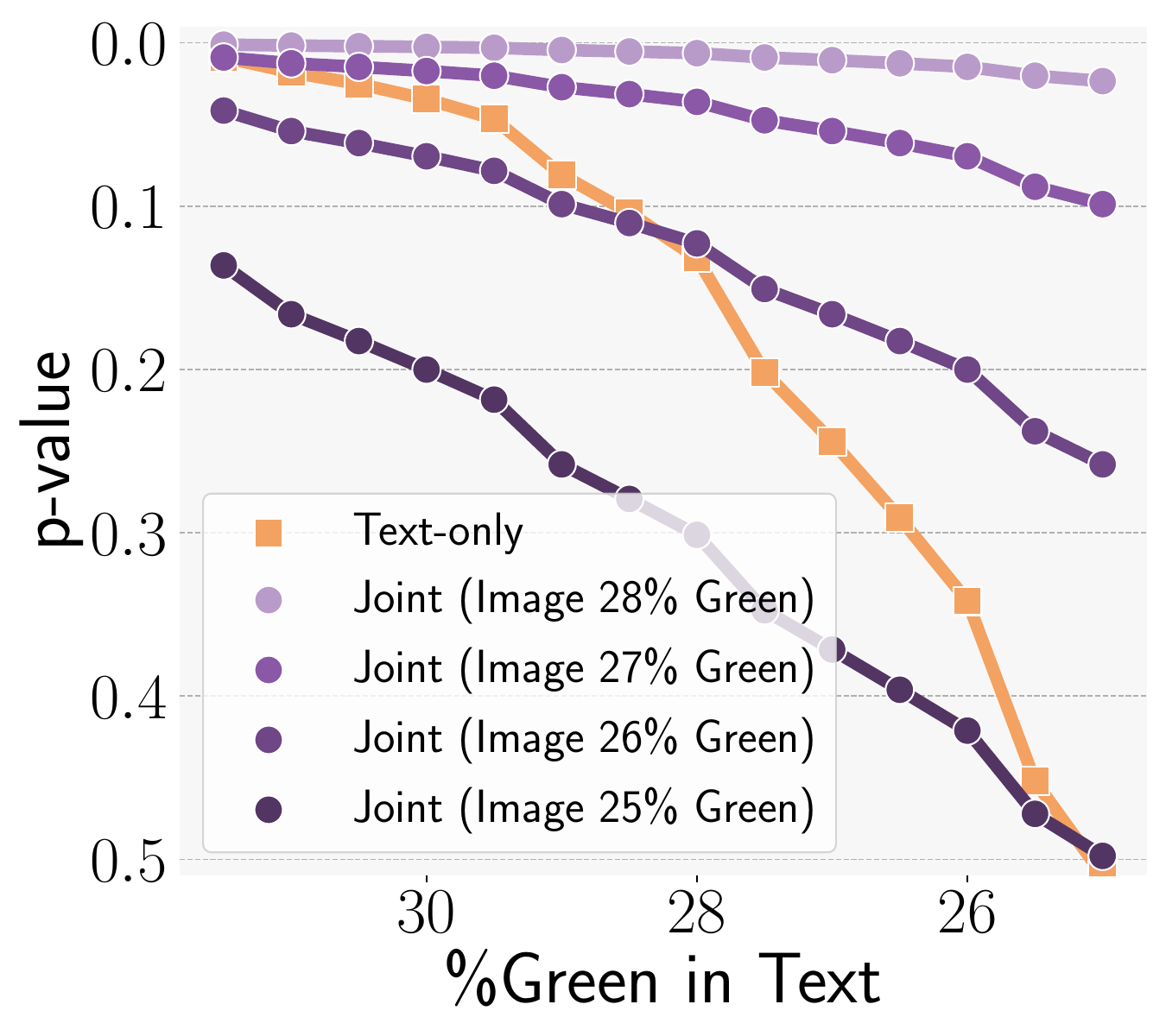}}
  \caption{
      \emph{Left}: Using more tokens at the same green ratio reduces the p-value. As orange crosses show, using $5\times$ more tokens can improve the p-value even if the green ratio drops ($28\% \to 27\%$).
      \emph{Right}: Simulated results similar to~\cref{fig:interleaved}. Both axes are flipped to match~\cref{fig:interleaved}: x-axis reduces the green\% (\emph{weaker signal}) and y-axis reduces the p-value (\emph{stronger detection}). As images carry many tokens, merging \eg text with $28\%$ green tokens with an image with $26\%$ green tokens still improves detection.}
  \label{fig:interleaved_extra}
  \vspace{-1em}
\end{figure}  

%% file: prompts/guitar.tex
\begin{prompt}
    I am wondering about the following: acoustic guitar. Can you teach me something about it in a way that is easy to understand and also generate a picture of it?
\end{prompt}
\begingroup\raggedleft
\begin{chameleon}{guitar}
Certainly, I'd be happy to help!

The acoustic guitar, as shown in the image below, is a popular instrument for playing music. It is made of a wooden body, a neck, and strings that are plucked or strummed to produce sound. The sound of an acoustic guitar is created by the vibrations of the strings, which are then amplified by the body of the guitar.
\end{chameleon}
\endgroup
\begingroup\raggedleft 
\begin{detector}{guitar_annotated}{$p=1.7 \cdot 10^{-16}$}
Certainly@,@@ I@^'^@d@^ be^^ happy^@ to@^ help^^!^^ 
^^
^^The^^ acoustic^^ guitar^@,@@ as@@ shown@@ in@@ the@^ image^@ below@^,^@ is@^ a^@ popular@@ instrument@@ for@@ playing@^ music^@.@@ It@^ is^@ made@@ of@^ a^@ wooden@@ body@^,^^ a^@ neck@@,@^ and^^ strings^^ that^^ are^@ pl@^ucked^^ or^^ str^^um^^med^^ to^^ produce^^ sound^^.^^ The^@ sound@@ of@@ an@^ acoustic^ guitar^ is^@ created@@ by@^ the^@ vibrations@^ of^^ the^^ strings^^,^^ which^^ are^@ then@^ amplified^@ by@ the^ body^^ of^ the^ guitar^^.^
\end{detector}
\endgroup

%% file: prompts/thatch.tex
\begin{prompt}
    I am wondering about the following: thatch. Can you teach me something about it in a way that is easy to understand and also generate a picture of it?
\end{prompt}
\vspace{1em}
\begingroup\raggedleft
\begin{chameleon}{thatch}
Sure! Thatch refers to the plant material used to cover roofs and walls in traditional or indigenous architecture. It is made from various types of grasses, leaves, or other vegetation. Here is an image of an example of thatch, used as roofing material in the construction of this house.Thatch is often made from locally available resources, such as reeds, rushes, or grasses, and is often woven or tied together to create a durable and effective roofing material. It requires regular maintenance to keep it in good condition and can last for many years with proper care. In some areas, thatch is still used today as a traditional building material, while in others, it has been largely replaced by modern roofing systems.
\end{chameleon}
\endgroup
\vspace{1em}
\begingroup\raggedleft 
\begin{detector}{thatch_annotated}{$p=1.2 \cdot 10^{-34}$}
Sure^!^^ Th^^atch^^ refers^^ to^@ the@@ plant@^ material^@ used@^ to^@ cover@^ roofs^@ and@@ walls@@ in@^ traditional^@ or@@ indigenous@^ architecture^^.^@ It@^ is^@ made@^ from^@ various@@ types@^ of^^ grass^^es^^,^^ leaves^^,^@ or@@ other@^ vegetation^@.@@ Here@@ is@^ an^@ image@^ of^@ an@^ example^^ of^^ that^^ch^@,@@ used@^ as^@ roof@@ing@^ material^@ in@@ the@@ construction@^ of^@ this@@ house@@.@@That@^ch^^ is^@ often@@ made@ from@ locally@^ available^@ resources@^,^@ such@^ as^@ re@@eds@@,@@ r@@ushes@^,^ or@ grass@es,^ and^^ is^ often^ woven^^ or^@ tied@@ together@@ to@^ create^^ a^^ durable^^ and^@ effective@^ roof^ing material^.^ It@ requires@^ regular^^ maintenance^@ to@^ keep^^ it^^ in^^ good^^ condition^@ and@@ can@^ last^^ for^^ many^^ years^^ with^@ proper@^ care^^.^@ In@^ some^@ areas@^,^^ that^ch is^ still^^ used^@ today@^ as^^ a^^ traditional^@ building@@ material@^,^^ while^^ in^^ others^@,@^ it^@ has@@ been@@ largely@^ replaced^^ by^^ modern^^ roof^ing@ systems@@.@

\end{detector}
\endgroup